\documentclass[nohyperref]{article}

\usepackage{microtype}
\usepackage{graphicx}
\usepackage{booktabs} % for professional tables

\usepackage{hyperref}

\usepackage[accepted]{icml2023}

\usepackage{amsmath}
\usepackage{amssymb}
\usepackage{mathtools}
\usepackage{amsthm}
\usepackage{subfig}
\usepackage{adjustbox}
\usepackage{algorithmic}
\usepackage{multirow}

\usepackage[capitalize,noabbrev]{cleveref}

\theoremstyle{plain}

\theoremstyle{definition}

\theoremstyle{remark}

\usepackage[textsize=tiny]{todonotes}

\icmltitlerunning{Variational Inference for Longitudinal Data Using Normalizing Flows}

\begin{document}

\onecolumn
\icmltitle{Variational Inference for Longitudinal Data Using Normalizing Flows}

% It is OKAY to include author information, even for blind
% submissions: the style file will automatically remove it for you
% unless you've provided the [accepted] option to the icml2022
% package.

% List of affiliations: The first argument should be a (short)
% identifier you will use later to specify author affiliations
% Academic affiliations should list Department, University, City, Region, Country
% Industry affiliations should list Company, City, Region, Country

% You can specify symbols, otherwise they are numbered in order.
% Ideally, you should not use this facility. Affiliations will be numbered
% in order of appearance and this is the preferred way.
%\icmlsetsymbol{equal}{*}

\begin{icmlauthorlist}
\icmlauthor{Clément Chadebec}{sch}
\icmlauthor{Stéphanie Allassonnière}{sch}
%\icmlauthor{Firstname4 Lastname4}{sch}
%\icmlauthor{Firstname5 Lastname5}{yyy}
%\icmlauthor{Firstname6 Lastname6}{sch,yyy,comp}
%\icmlauthor{Firstname7 Lastname7}{comp}
%\icmlauthor{}{sch}
%\icmlauthor{Firstname8 Lastname8}{sch}
%\icmlauthor{Firstname8 Lastname8}{yyy,comp}
%\icmlauthor{}{sch}
%\icmlauthor{}{sch}
\end{icmlauthorlist}

%\icmlaffiliation{yyy}{Department of XXX, University of YYY, Location, Country}
%icmlaffiliation{comp}{Company Name, Location, Country}
\icmlaffiliation{sch}{Université Paris Cité, INRIA, Inserm, SU,  Centre de Recherche des Cordeliers }

\icmlcorrespondingauthor{Clément Chadebec}{clement.chadebec@inria.fr}
%\icmlcorrespondingauthor{Firstname2 Lastname2}{first2.last2@www.uk}

% You may provide any keywords that you
% find helpful for describing your paper; these are used to populate
% the "keywords" metadata in the PDF but will not be shown in the document
%\icmlkeywords{Machine Learning, ICML}

\vskip 0.3in

% this must go after the closing bracket ] following \twocolumn[ ...

% This command actually creates the footnote in the first column
% listing the affiliations and the copyright notice.
% The command takes one argument, which is text to display at the start of the footnote.
% The \icmlEqualContribution command is standard text for equal contribution.
% Remove it (just {}) if you do not need this facility.

%\printAffiliationsAndNotice{}  % leave blank if no need to mention equal contribution
\printAffiliationsAndNotice{} % otherwise use the standard text.

\begin{abstract}
This paper introduces a new latent variable generative model able to handle high dimensional longitudinal data and relying on  variational inference. The time dependency between the observations of an input sequence is modelled using normalizing flows over the associated latent variables. The proposed method can be used to generate either fully synthetic longitudinal sequences or trajectories that are conditioned on several data in a sequence and demonstrates good robustness properties to missing data. We test the model on 6 datasets of different complexity and show that it can achieve better likelihood estimates than some competitors as well as more reliable missing data imputation.\footnote{A code is made available at \url{https://github.com/clementchadebec/variational_inference_for_longitudinal_data}}

\end{abstract}

\section{Introduction}
\label{submission}

Longitudinal data are more than common in many application fields such a medicine \emph{e.g.} for disease progression modelling ~\cite{aghili2018predictive,zhao2021longitudinal} or monitoring treatment response~\cite{blackledge2014assessment}. They consist in the observation of a given entity's or individual's evolution though time but contrary to \textit{time-series}, the number of observations of a single entity may be pretty small. Moreover, such data can be of high dimension (\emph{e.g.} images) and we may only have access to a reduce number of different entities (\emph{e.g.} rare diseases follow-ups) leading to small databases and missing values (\emph{e.g.} a missing observation at a given time or loss in follow-up of a given entity). All of these aspects make these data challenging to model. 

Generative models such as Variational Autoencoders (VAEs) introduced in \citep{kingma_auto-encoding_2014, rezende_stochastic_2014} appeared to be powerful models to model distributions and would be an interesting choice to consider for longitudinal data. Unfortunately, while they appear to be able to perform some disentanglement of the input data in their latent space \citep{higgins_beta-vae_2017,burgess2018vae, kim2018factorvae,chen2019vae}, they struggle to capture more complex correlations such as time evolution for longitudinal data \cite{ramchandran_longitudinal_2020}. To address this limitation and improve the latent representations of the input data, methods trying to account for the correlations of the data in the latent space of VAEs \cite{sohn2015learning}, proposing new prior distributions \citep{nalisnick_approximate_2016,sonderby_ladder_2016,dilokthanakul_deep_2017,tomczak_vae_2018,razavi_generating_2019,pang_learning_2020} or seeking to enhance the expressiveness of the approximate posterior distribution \citep{salimans_markov_2015, rezende_variational_2015} were proposed. With a specific focus on temporal coherence, works introducing priors using Gaussian Processes were also introduced \citep{casale2018gaussian,fortuin2020gp,ramchandran_longitudinal_2020}. Nonetheless, those models were mainly designed to perform missing data imputation or for conditional settings and so are not well suited for unconditional sequence generation.
Approaches relying on neural ODE (NODE) \cite{chen2018neural,xu2021learning,massaroli2020dissecting,dupont2019augmented}, deep state models \cite{rangapuram2018deep,klushyn2021latent}, latent RNN models \cite{chung2015recurrent,serban2017hierarchical} or latent SDE \cite{tzen2019neural,li2020scalable} have also demonstrated promising results to model time dependent data and to perform time series forecasting, interpolation or for classification purposes. Nonetheless, their applicability to the context of unconditional high dimensional data generation remains poorly explored.

Focusing more specifically on medical applications, several works have analysed longitudinal data through the prism of progression models using in particular mixed-effects models \cite{schiratti2015learning,bone2018learning}. In these approaches, patients are assumed to follow a given trajectory that deviates from a reference curve that may, for example, represent the average progression of a given disease. These approaches were then combined with dimensionality reduction using autoencoders \citep{louis2019riemannian} or VAEs \citep{sauty2022progression}. However, these methods remain limited to the context of disease progression because they assume the existence of an intrinsic average trajectory from which each subject deviates, which may no longer be a valid assumption for heterogeneous datasets. 

In this paper, we take quite a different approach and propose the following contributions:
\begin{itemize}
    \item We propose a new generative latent variable model imposing time dependency of the observations in an input sequence using normalizing flows on the associated latent variables. A training procedure relying on variational inference is also derived. 
    \item We show that the model is capable of handling high dimensional longitudinal data and able to generate fully synthetic sequences or trajectories conditioned on several input data.
    \item We discuss the modularity of the proposed model and show that it can benefit pretty easily from improvements available in the variational inference literature.   
    \item We show that the method achieves better likelihood estimates that competitors on benchmark datasets and can outperform them for missing data imputation.
\end{itemize}

\section{Background}
In this section, we first recall some elements on variational inference and normalizing flows needed in the proposed method.
\subsection{Variational Inference}
Given observations $x \in \mathbb{R}^D$ and associated latent variables $z \in \mathbb{R}^d$ with joint distribution $p(x, z)$, variational inference \cite{jordan_introduction_1999} is a method that aims at approximating an untractable conditional distribution $p(z|x)$ of the latent variables given the observations using a family of parametrized distributions $q_{\phi}(z|x)$  \cite{blei2017variational}. The idea is to find the set of parameters $\phi$ that minimises the Kullback-Leibler (KL) divergence between the approximate posterior and the true one \emph{i.e.} $\min \limits _{\phi} \mathrm{KL}(q_{\phi}(z|x)||p(z|x))$. However, this objective is most of the time untractable since $p(z|x)$ is unknown and so a surrogate objective is optimised instead and obtained using Jensen's inequality \cite{jordan_introduction_1999}:
\begin{equation}
    \begin{aligned}
        \log p(x) = \log\int_{\mathbb{R}^d} p(x,z)dz &= \log \mathbb{E}_{q_{\phi}} \Bigg [\frac{p(x,z)}{q_{\phi}(z|x)}\Bigg]
        \geq \mathbb{E}_{q_{\phi}} \log \Bigg [\frac{p(x,z)}{q_{\phi}(z|x)}\Bigg]\,.
    \end{aligned}
\end{equation}
The right hand side of the equation is called the Evidence Lower BOund (ELBO) and one may notice that the difference between the left hand side of the equation and the ELBO gives $\mathrm{KL}(q_{\phi}(z|x)||p(z|x))$. Hence, maximising the ELBO amounts to minimising the KL and so the ELBO is used as objective for the variational approximation.

\subsection{Normalizing Flows}

Normalizing flows are flexible models that can be used to transform simple probability densities into much complex ones by re-coursing to sequences of invertible smooth mappings. They have, for instance, been proposed to enhance the expressiveness of the approximate posterior distribution used in the context of variational inference in \cite{rezende_variational_2015}. These models rely on the rule of change of variables such that if $z \in \mathbb{R}^d$ is a random variable that follows the distribution $q(z)$ and $f:\mathbb{R}^d \to \mathbb{R}^d$ is an invertible smooth function, then the random variable $z' = f(z)$ has a distribution given by 
\begin{equation}\label{eq: change of variable}
    q(z') = q(z)\bigg |\det \frac{\partial f^{-1}}{\partial z'} \bigg| = q(z) \bigg|\det \frac{\partial f}{\partial z} \bigg|^{-1}\,.
\end{equation}
In this setting, $f$ is called a \emph{normalizing flow} and so several flows can be composed to form a new flow $g = f_K \circ f_{K-1}\circ\dots \circ f_1$ allowing to model richer distributions. In the context of variational inference, these flows can be parameterised as well and so can be used to have access to enhanced approximate posterior distributions $q_{\phi}(z|x)$ provided that the computation of the det-Jacobian of the flows is tractable. Amongst the most widely known flows we can cite NICE \cite{dinh2014nice}, linear and planar flows \citep{rezende_variational_2015}, RealNVP \cite{dinh2016density}, Masked Autoregressive Flows (MAF) \cite{papamakarios2017masked}
or Inverse Autoregressive Flows (IAF) \cite{kingma2016improved}.

\section{The Proposed Model}
In this section, we propose a new generative latent variable model suited for longitudinal data. 
\subsection{Problem Setting}
Let us define $P$ as the number of entities observed through time. For each entity $i \in \{1, \dots, P\}$, we are given a sequence of $t_i+1$ observations $x^i = (x^i_0, \dots, x^i_{t_i})$ such that $x^i_j \in \mathcal{X} =\mathbb{R}^D,~\forall j \in \{0, \dots, t_i\}$. Assuming that the sequence $x^i$ is generated from an unknown distribution $p$, our goal is to infer $p$ with a parametric model $\{p_{\theta}, \theta \in \Theta\}$.

\subsection{The Probabilistic Model}
Given an entity $i \in \{1, \dots, P\}$ and a sequence of observation $(x^i_0, \dots, x^i_{t_i})$, we assume that for each $x^i_j$ where $j \in \{0, \dots, t_i\}$, there exists an associated latent variable $z^i_j \in \mathcal{Z} =\mathbb{R}^d$ involved in the generative process of the observation $x_j^i$ such that $x_j^i\sim p_{\theta}(x_j^i|z_j^i)$. One may further write the joint distribution $p_{\theta}$ as follows:
\begin{equation}\label{eq: simple obj}
    p_{\theta}(x_0^i, \cdots, x_{t_i}^i) = \int_{\mathcal{Z}} p_{\theta}(x_0^i, \cdots, x_{t_i}^i|z_j^i)p(z_j^i)dz_j^i\,,
\end{equation}
where $p(z_j^i)$ is a prior distribution over $z_j^i$. An important point in this setting is that the observations $x_j^i$ are no longer independent and so the joint likelihood is no longer factorisable. In this paper, we propose to model the time dependency of the observations in an input sequence using the latent variables and normalizing flows as follows
\begin{equation}\label{eq: time dep}
    \begin{aligned}
        z_0^i  \sim p(z_0^i),~
        z_1^i = f_1(z_0^i),~
        %z_2^i = f_2(z_1^i)~
        \dots~,
        z_{t_i}^i = f_{t_i}(z_{t_{i-1}}^i)\,,
    \end{aligned}
\end{equation}
\begin{figure}[t]
\centering
    \captionsetup[subfigure]{position=below, labelformat = empty}
    \subfloat[Inference model]{\includegraphics[width=0.5\linewidth]{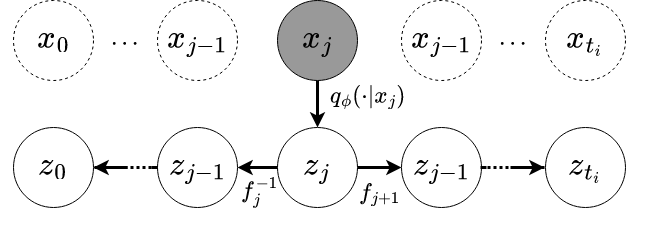}}\hspace{3em}
    \subfloat[Generative model]{\includegraphics[width=0.3\linewidth]{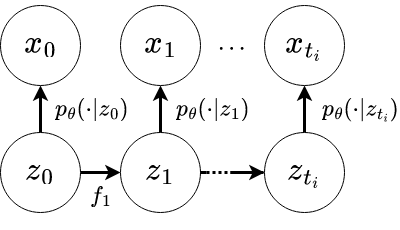}}
    \caption{Proposed inference and generative models.}
    \label{fig:graph models}
\end{figure}
where $p$ is a simple prior distribution over $z_0^i$ (\emph{e.g.} standard Gaussian) and $f_j$ are normalizing flows for any $j \in \{1, \dots, t_i\}$. The main idea is to assume that it is the distribution of the latent variables that evolves through time and we propose to model this evolution using the flows. As such, the time dependency is imposed on the latent variables and not directly on the observations. Note that the initial distribution can be chosen as complex as desired and that for any $j \in \{1,\dots, t_i\}$ we have access to a tractable density for $p(z_j^i)$ using Eq.~\eqref{eq: change of variable}:
\begin{equation}\label{eq: priors}
    p(z_j^i) = p(z_0^i)\prod\limits_{l=1}^j \left | \det \frac{\partial f_l}{\partial z_{l-1}^i}\right|^{-1}\,.
\end{equation}
In addition, the relation between two latent variables $z_j^i$ and $z_k^i$ with $j, k \in \{0, \dots, t_i\}$ such that $j < k$ is explicit and completely deterministic since we have:
\begin{equation}\label{eq: deterministic}
    z_k^i = \bigcirc_{l={j+1}}^{k} f_l(z_j^i)~\text{and}~z_j^i = \bigcirc_{l={k}}^{j+1} (f_l)^{-1}(z_k^i)\,.
\end{equation}
Hence, we can see that given a latent vector $z_j^i$ we can now retrieve the complete sequence $(z_0 ^i, \dots, z_{t_i}^i)$ using Eq.~\eqref{eq: deterministic}. Assuming $(x_j^i)_{j \in \{1,\dots, t_i\}}$ are independent knowing $(z_0 ^i, \dots, z_{t_i}^i)$, the conditional distribution in Eq.~\eqref{eq: simple obj} writes 
\begin{equation}\label{eq: conditional split}
    \begin{aligned}
        p_{\theta}(x_0^i, \cdots, x_{t_i}^i|z_j^i) = \prod\limits_{l=0}^{t_i} p_{\theta}(x_l^i|z_j^i) = \prod\limits_{l=0}^{t_i} p_{\theta}(x_l^i|z_l^i)\,.
    \end{aligned}
\end{equation}
Using Eq.~\eqref{eq: priors} and Eq.~\eqref{eq: conditional split} allows to derive another expression of the joint distribution of the observations:
\begin{equation}\label{eq: joint lik}
    p_{\theta}(x_0^i, \cdots, x_{t_i}^i) = \int_\mathcal{Z} \prod\limits_{l=0}^{t_i} p_{\theta}(x_l^i|z_l^i)p(z_j^i)dz_j^i\,.
\end{equation}
Since this integral is most of the time intractable, we propose to rely on variational inference \cite{jordan_introduction_1999}. We indeed introduce a parametrized variational distribution $q_{\phi}(z_j^i|x_j^i)$ such that we can obtain an unbiased estimate of the joint likelihood:
\begin{equation}\begin{aligned}
   \mathbb{E}_{q_{\phi}}\Bigg[ \frac{\prod\limits_{l=0}^{t_i} p_{\theta}(x_l^i|z_l^i)p(z_j^i)}{q_{\phi}(z_j^i|x_j^i)}\Bigg] &= p_{\theta}(x_0^i, \cdots, x_{t_i}^i)\,.
\end{aligned}
\end{equation}
Using Jensen's inequality allows to derive a lower bound (ELBO) on the true objective \emph{i.e.} the log joint likelihood :
\begin{equation}\begin{aligned}\label{eq: ELBO}
    \log~p_{\theta}(x_0^i, \cdots, x_{t_i}^i) = \log \mathbb{E}_{q_{\phi}}\Bigg[ \frac{\prod\limits_{l=0}^{t_i} p_{\theta}(x_l^i|z_l^i)p(z_j^i)}{q_{\phi}(z_j^i|x_j^i)}\Bigg]
    &\geq  \mathbb{E}_{q_{\phi}} \log \Bigg[ \frac{\prod\limits_{l=0}^{t_i} p_{\theta}(x_l^i|z_l^i)p(z_j^i)}{q_{\phi}(z_j^i|x_j^i)}\Bigg] \,,\\
    &\geq \mathbb{E}_{q_{\phi}} \log \prod\limits_{l=0}^{t_i}p_{\theta}(x_l^i|z_l^i) - \mathrm{KL}(q_{\phi}(z_j^i|x_j^i)|p(z_j^i))\,.
\end{aligned}
\end{equation}

\begin{algorithm}[t]
   \caption{Training Procedure}
   \label{alg:training}
\begin{algorithmic}
   \STATE{\bfseries Input:} Observations $(x_0^i, \cdots, x_{t_i}^i)$
   \WHILE{not converged}
   \STATE Pick $j \in \{0, \dots, t_i\}$ randomly\;
   \STATE $z_j^i \sim q_{\phi}(\cdot| x_j^i)$\; 
   \FOR{$l=j+1$ {\bfseries to} $t_i$}
   \STATE $z_l^i = f_l(z_{l-1})$  \COMMENT{propagate in future}\;
   \ENDFOR
   \FOR{$l=j-1$ {\bfseries to} $0$}
   \STATE $z_l^i = (f_{l+1})^{-1}(z_{l+1})$ \;  \COMMENT{propagate in past}
   \ENDFOR
   \STATE $\mathcal{L} = -\frac{1}{t_i+1}\sum\limits_{l=0}^{t_i} \log p_{\theta}(x_l^i|z_l^i) + \log q_{\phi}(z_j^i|x_j^i) - \log p(z_0^i) - \sum\limits_{l=1}^j \log \Big| \det \frac{\partial (f_l)^{-1}}{\partial z_l^i}\Big|$
   \ENDWHILE
\end{algorithmic}
\end{algorithm}
The graphical models for the proposed method can be found in Fig.~\ref{fig:graph models}. In practice and inspired from the VAE framework, the variational distribution is chosen as a multivariate Gaussian distribution $q_{\phi}(z_j^i|x_j^i)=\mathcal{N}(z_j^i; \mu_{\phi}(x_j^i), \Sigma_{\phi}(x_j^i))$ for $j\in\{0, \dots, t_i\}$ and where $\mu_{\phi}$ and $\Sigma_{\phi}$ are given by neural networks and $\Sigma_{\phi}$ is chosen as a diagonal matrix. The conditional distributions $p_{\theta}(x_j^i|z_j^i)$ are chosen depending on the input data (\emph{e.g.} multivariate Gaussians for RGB images) and $p(z_j^i)$ is given by Eq.~\eqref{eq: priors}. To mitigate the impact of the sequence length on the ELBO in Eq.~\eqref{eq: ELBO}, we average the left hand side term over the sequence length. This impedes the reconstruction term to over-weight the KL for long sequences. As for the normalizing flows, in this paper, we use Inverse Autoregressive Flows (IAF) \cite{kingma2016improved} with MADE \cite{germain2015made} for the autoregressive networks since we need a tractable inverse. It should be noted that for such flows the computation of the inverse is however sequential and its time proportional to the dimensionality of the latent variables due to the autoregressive property of the flows. Nonetheless, in practice the dimension of the latent variables is often much smaller than the dimensionality of the input data making this choice reasonable. We choose IAF over MAF \cite{papamakarios2017masked} so that the generation of a synthetic sequence from the prior $z_0\sim p(z_0)$ is fast since it does not require inverting the flows. Finally, a pseudo code of the training algorithm is provided in Alg.~\ref{alg:training} and an implementation using PyTorch \cite{paszke_automatic_2017} and based on \cite{chadebec2022pythae} is made available in the supplementary materials.

\subsection{Dealing with Missing Data in the Sequence}\label{sec: missing data method}
In \textit{real-life} applications, it is not rare to find sequences with missing observations (\emph{e.g.} in medicine a loss of patient follow-up or a patient not coming to a specific visit induces missing observations for the patient's evolution). As explained above and shown in Alg.~\ref{alg:training}, during training we perform variational inference using only one element in the sequence. Thus, the training can be modified pretty easily to handle such missing data in the input sequences and consists in only using the observed data.   

Nonetheless, this can be seen as a weakness of the method at inference time. Let us indeed imagine that we are given a sequence of 5 measure times $(x_0^i, x_1^i, x_2^i, x_3^i, x_4^i)$ where only 3 are actually observed, say $x_1^i, x_2^i$ and $x_4^i$. In its current shape, during inference, the method will choose an observation time $j \in \{1, 2, 4\}$, say $j=2$, sample a latent variable associated to observation $x_2^i$ using the approximate posterior $q_{\phi}(\cdot|x_2^i)$ and then generate a sequence $(z_l^i)_{l \in \{0, \dots, 4\}}$ using the learned flows. This sequence is then used to sample a reconstructed sequence in the observations space using $p_{\theta}(x|z)$. This is actually sub-optimal since this would be equivalent to only have access to observation $x_2^i$ without benefiting from the information provided by $x_1^i$ and $x_4^i$. In order to address this potential limitation of the model, we propose to generate a sequence (actually we can generate an arbitrary number of sequences) for each index corresponding to an observed input data in the sequence  (\emph{i.e.} $\{1, 2, 4\}$ in the example) and keep the generated sequence achieving the highest likelihood on the observed data. In other words, if we denote $\mathcal{O}_i$ the set of observed indices in the input sequence, we define the \textit{optimal} index that should be used to complete the sequence as follows:
\begin{equation}\label{eq: missing optimal}
    \begin{aligned}
        j_{\mathrm{opt}} = \arg \max_{j\in \mathcal{O}_i} \sum\limits_{l\in\mathcal{O}_i} \log p_{\theta}(x_{l,j}^i|z_{l, j}^i)\,,
    \end{aligned}
\end{equation}
where $z_{l, j}^i$ is the latent variable and $x_{l, j}^i$ the data generated at time $l$ using index $j$. Alg.~\ref{alg: infer with missing} shows the inference procedure.

\begin{algorithm}[t]
   \caption{Inference Procedure for Missing Observations}
   \label{alg: infer with missing}
\begin{algorithmic}
   \STATE{\bfseries Input:} A sequence $(x^i_{j})_{j\in \mathcal{O}_i}$ with missing observations 
   \FOR{$j \in \mathcal{O}_i$} 
   \STATE $z_{j,j}^i \sim q_{\phi}(\cdot| x_j^i)$\;
    \STATE $\hat{x}_{j, j}^i \sim p_{\theta}(\cdot|z_{j, j}^i)$\;
   \FOR{$l=j+1$ {\bfseries to} $t_i$}
   \STATE $z_{l, j}^i = f_l(z_{l-1, j})$ \;
   \STATE $\hat{x}_{l, j}^i \sim p_{\theta}(\cdot|z_{l, j}^i)$\;
   \ENDFOR
   \FOR{$l=j-1$ {\bfseries to} $0$}
   \STATE $z_{l, j}^i = (f_{l+1})^{-1}(z_{l+1, j})$ \;
   \STATE $\hat{x}_{l, j}^i \sim p_{\theta}(\cdot|z_{l, j}^i)$\;
   \ENDFOR
   \ENDFOR
   \STATE $j_{\mathrm{opt}} = \arg \max \limits _{j\in \mathcal{O}_i}  \sum\limits_{l\in\mathcal{O}_i} \log p_{\theta}(x_{l, j}^i|z_{l, j}^i)$
   \STATE \textbf{return} $(\hat{x}_{0, j_{\mathrm{opt}}}^i, \cdots, \hat{x}_{t_i, j_{\mathrm{opt}}}^i)$ \COMMENT{obtained with $j_{\mathrm{opt}}$}
\end{algorithmic}
\end{algorithm}

\subsection{Enhancing the Model}
One interesting aspect of the model is that one may use improvements that have been proposed and proved useful in the literature related to variational inference and VAEs to enhance several part of the model independently. 
\paragraph{Improving the prior} Even-though, a simple distribution such as a standard Gaussian appeared to work well in practice, a smarter choice in the prior distribution may result in an enhanced data generation or better likelihood estimates~\cite{hoffman_elbo_2016}. As such, richer priors \cite{nalisnick_approximate_2016,dilokthanakul_deep_2017} or priors that are learned \cite{chen_variational_2016,razavi_generating_2019,pang_learning_2020,aneja_ncp-vae_2020} can be easily plugged into our model. We show in the experiments section how changing the prior from a standard Gaussian to a VAMP prior \cite{tomczak_vae_2018} can influence the results. 
\paragraph{Improving the variational bound} Following \cite{rezende_variational_2015} insights, another way to improve the expressiveness of the model and ideally achieve a tighter ELBO consists in enriching the potentially too simplistic parameterised variational distribution $q_{\phi}(z|x)$ using flows. This improvement can be easily integrated within our framework as well. In the experiments section, we also propose a variant model where the posterior distributions are enriched using IAF flows as proposed in \cite{kingma2016improved}. Methods proposing to use MCMC sampling steps with learned Markov kernels \cite{salimans_markov_2015} or relying on Hamiltonian dynamics \cite{caterini_hamiltonian_2018,chadebec_data_2021} could also be envisioned but are not tested in combination with our model due to the strong computation burden they imply.

\section{Related Works}

Variational Autoencoders (VAEs) \cite{kingma_auto-encoding_2014,rezende_stochastic_2014} share some aspects with our method. First, they try to maximise the likelihood of a set of data using a variational approach. Second, they try to take advantage of the flexibility a latent space provides by mapping potentially high dimensional input data into a lower dimensional  space. However, they assume that the input data are independent and so are the latent variables. This impedes the model to capture the potentially complex time dependency that exists with longitudinal data.

There exist however some works on VAEs that are worth citing since they stress the flexibility offered by considering a latent space. In particular, they motivated our idea to impose the time dependency over the latent variables and not directly on the observations. First, several papers argued that the latent space of the VAE can reveal representative and interpretable features through its ability to perform disentanglement \citep{higgins_beta-vae_2017,burgess2018vae,korkinof_high-resolution_2018,chen2019vae}. Studying the latent space with a geometric point of view, other works showed that this latent space can actually be modelled with a specific geometry (\emph{e.g.} hyper-sphere, Poincaré Disk, Riemannian manifold) \cite{davidson_hyperspherical_2018,falorsi_explorations_2018,mathieu_continuous_2019,kalatzis_variational_2020,chadebec_data_2021} or that a Riemannian geometry can naturally arise in the latent space \cite{arvanitidis_latent_2018,chen_metrics_2018,shao_riemannian_2018,chadebec2022geometric}. Finally, another way to enhance the representation capability of the model that was proposed in the literature consists in increasing the expressiveness of the variational posterior distribution using MCMC sampling \citep{salimans_markov_2015} or normalizing flows \citep{rezende_variational_2015}.

Arguing that VAEs still fail to capture complex correlations, there were some proposals in the literature trying to constraint the model to account for these correlations in the latent space. For instance, the conditional VAE \cite{sohn2015learning} feeds auxiliary variables directly to the encoder and decoder networks but fails to model the evolution of a given subject through time. Gaussian processes that are a powerful tool for time series \cite{seeger2004gaussian,roberts2013gaussian} were also proposed as prior for the VAE \cite{casale2018gaussian,fortuin2020gp} to account for the temporal structure across the samples. \cite{ramchandran_longitudinal_2020} enriched these models with the inclusion of covariates different from time using a multi-output additive Gaussian process prior. 

Also closely related to our method are approaches involving neural ordinary differential equations (NODE) that see the forward pass of a residual network as solving an ODE \cite{chen2018neural}, an approach that was for instance extended in \cite{dupont2019augmented,rubanova2019latent,massaroli2020dissecting}. In particular, the latent neural ODE model proposed in \cite{chen2018neural} defines a generative model by assuming that the initial state latent variable follows a given prior distribution and a latent trajectory is then obtained by solving an ODE. The model also relies on variational inference but considers an approximate posterior conditioned on the entire input sequence leading to very different latent representations when compared to our method. The idea was further enriched in \cite{rubanova2019latent} to handle irregularly-sampled time series and extended to SDE in \cite{chung2015recurrent,serban2017hierarchical}. However, these models were rarely used on high dimensional sequences such as images while the method we propose appears well suited for such type of data. We can nonetheless cite \cite{kanaa2021simple,park2021vid} but that only validated their method on simple databases and \cite{yildiz2019ode2vae} that proposed to optimize a complex loss function and to rely on adversarial training making the training procedure tricky. Moreover, these works were mainly designed for conditional generation (prediction of future time points or interpolation), while the usability of these methods for unconditional generation remains to be proven.

Modelling longitudinal data and trying to understand the underlying evolution dynamic is something that has also been quite studied under the prism of disease progression modelling \cite{jedynak2012computational,fonteijn2012event}. In such literature, \textit{mixed-effect} models \cite{laird1982random} that parameterise a patient's evolution as a deviation from a reference trajectory have become more and more popular \cite{diggle2002analysis,singer2003applied}. First applied on Euclidean data \cite{bernal2013statistical}, they were then extended with a Riemannian geometry viewpoint \cite{schiratti2015learning,singh2016hierarchical,koval2017statistical,bone2018learning} or combined with dimensionality reduction \cite{louis2019riemannian,sauty2022progression}. Despite being adapted to model disease progression, it is unclear how these models would apply to datasets where there is no clear \emph{average} evolution. 

Finally, deep learning based methods relying on recurrent neural networks are also worth citing as they revealed useful for time varying data \cite{pearlmutter1989learning}. To cite a few, GRUI-GAN \cite{luo2018multivariate} and BRITS \cite{cao2018brits} were proposed with the aim of handling missing data but with the drawback of relying on adversarial training for the first one and not being generative for the second. \cite{chung2015recurrent} proposed a combination of VAE and RNN for structured sequential data but there exists no clear way how the model would handle missing data.

\section{Experiments}

In this section, we validate the proposed method through series of experiments. We place ourselves in the context of high-dimensional data (images) and so set $d \ll D$ (\emph{i.e.} the latent variables live in a much lower-dimensional latent space when compared to the input images size). In line with the VAE framework, the inference network providing the parameters of the variational distribution $q_{\phi}(z|x)$ can then be interpreted as an \emph{encoder} and the generative model $p_{\theta}(x|z)$ as a \emph{decoder}. Note that neither the \textit{encoder} nor the \textit{decoder} depend on time. First, we show that the proposed model is able to achieve better joint likelihood estimates than several models proposed in the literature on 5 datasets. Then, we show that the method is also able to impute missing data (and features) and compare its performances in term of reconstruction with benchmark models. Finally, we evaluate the ability of the proposed model to generate relevant conditioned and fully synthetic sequences. We also conduct in Appendix \ref{app: ablation study} an ablation study stressing the influence of the flows, latent space dimension and prior complexity and discuss the relevance of Eq.~\eqref{eq: missing optimal} for missing data imputation in Appendix~\ref{app:influence eq 11}.

\begin{table*}[t]
\caption{Negative log joint likelihood divided by the sequence length computed on an independent test set with 5 independent runs and 100 importance samples.}
\label{tab: nll}
\vskip 0.15in
\begin{center}
\begin{small}
\begin{sc}
\begin{tabular}{lccccc}
\toprule
Model   & Starmen & RotMNIST &ColorMNIST & 3d chairs & Sprites\\
\midrule
VAE                             & 3781.82 $\pm$ 0.01 & 741.03 $\pm$ 0.00 & 2179.88 $\pm$ 0.00   & \textbf{ 11359.39 $\boldsymbol{\pm}$ 0.02} & 11313.38 $\pm$ 0.00 \\
VAMP                            & 3780.99 $\pm$ 0.01 & 740.82 $\pm$ 0.00 & 2179.60 $\pm$ 0.00   & 11361.01 $\pm$ 0.02                        & 11313.43 $\pm$ 0.00 \\
TVAE (0)                           & 3806.26 $\pm$ 0.07 & 748.18 $\pm$ 0.00 & 2185.40 $\pm$ 0.00   & 11419.32 $\pm$ 0.11                        & 11332.09 $\pm$ 0.01 \\
TVAE (short)                    & 3782.39 $\pm$ 0.05 & 744.07 $\pm$ 0.00 & 2175.54 $\pm$ 0.00   & 11373.76 $\pm$ 0.07                        & 11318.96 $\pm$ 0.01 \\
TVAE (part)                      & 3780.75 $\pm$ 0.05 & 739.53 $\pm$ 0.00 & 2174.19 $\pm$ 0.00   & 11364.21 $\pm$ 0.18                        & 11308.40 $\pm$ 0.01 \\
TVAE (half)                     & 3777.57 $\pm$ 0.07 & 745.55 $\pm$ 0.00 & 2173.58 $\pm$ 0.00   & 11363.27 $\pm$ 0.12                        & 11305.62 $\pm$ 0.01 \\
BubbleVAE                       & 3780.46 $\pm$ 0.07 & 742.66 $\pm$ 0.00 & 2174.74 $\pm$ 0.00   & 11369.59 $\pm$ 0.19                        & 11310.69 $\pm$ 0.01 \\
GPVAE (Cauchy)                  & 3780.36 $\pm$ 0.03 & 740.05 $\pm$ 0.00 & 2177.21 $\pm$ 0.00   & 11367.43 $\pm$ 0.02                        & 11309.11 $\pm$ 0.01 \\
GPVAE (rbf)                     & 3787.80 $\pm$ 0.03 & 745.58 $\pm$ 0.00 & 2187.15 $\pm$ 0.00   & 11390.33 $\pm$ 0.04                        & 11315.68 $\pm$ 0.01 \\
GPVAE (diffusion)               & 3780.96 $\pm$ 0.01 & 740.26 $\pm$ 0.00 & 2178.63 $\pm$ 0.00   & \textbf{11359.21 $\boldsymbol{\pm}$ 0.00}  & 11312.50 $\pm$ 0.00 \\ 
GPVAE - matern                  & 3779.29 $\pm$ 0.02 & 739.68 $\pm$ 0.00 & 2176.63 $\pm$ 0.00   & 11360.36 $\pm$ 0.03                        & 11309.90 $\pm$ 0.00 \\
\midrule
Ours ($\mathcal{N}$)     & 3773.23 $\pm$ 0.17                        & 735.71 $\pm$ 0.00 & 2173.16 $\pm$ 0.05 & 11362.00 $\pm$ 0.62                                & 11301.51 $\pm$ 0.04                                                         \\
Ours (VAMP)              & \textbf{ 3772.91 $\boldsymbol{\pm}$ 0.16} & 736.15 $\pm$ 0.00 & 2173.00 $\pm$ 0.05 & 11364.73 $\pm$ 0.51                                & \textbf{11301.30 $\boldsymbol{\pm}$ 0.02 }                                  \\
Ours (IAF)               & \textbf{3773.01} $\boldsymbol{\pm}$ \textbf{0.17}                        & \textbf{735.27} $\boldsymbol{\pm}$ \textbf{0.01} &  \textbf{2172.85} $\boldsymbol{\pm}$ \textbf{0.05}                 & \textbf{11359.48} $\boldsymbol{\pm}$ \textbf{0.67} & 11301.97 $\pm$ 0.02                          \\
\bottomrule
\end{tabular}
\end{sc}
\end{small}
\end{center}
\vskip -0.1in
\end{table*}

\subsection{Data}
For these experiments, we consider 5 different databases that mimic longitudinal datasets. The first one is a synthetic longitudinal dataset composed of 1,000, 64x64 images of \textit{starmen} raising their left arm and generated according to the diffeomorphic model of \cite{bone2018learning}. The second one consists of 8 evenly separated rotations applied to the MNIST database \cite{lecun_mnist_1998} from 0 to 360 degrees, we call it \emph{rotMNIST}. In addition to these two \textit{toy} datasets, we also consider more challenging ones. The third one called \textit{colorMNIST} is created using the approach of \cite{keller2021topographic}. It consists of sequences of colored MNIST digits that can undergo three distinct types of transformations: color change (from turquoise to yellow), scale change or rotations. It is important to note that for this database, the time dynamic cannot be fully recovered from a single image since it can correspond to different transformations. For instance, a starting turquoise \textit{6} can either change of color or undergo a change in scale. The fourth database is created using the \textit{3d chairs} dataset \cite{aubry2014seeing} consisting of 3D CAD chair models and considering as input sequences, 11 evenly separated rotations of a chair (from 0 to 360$^{\circ}$). Finally, we use the \textit{sprites} dataset \cite{li2018disentangle} consisting of 64x64 RGB images of characters performing actions such as dancing or walking. Find more details about the datasets, potential pre-processing steps and some examples of the training sequences in Appendix.~\ref{app: the data}.

\subsection{Likelihood Estimation}

First, we compare the proposed model and two of its variants (using either a VAMP prior or IAF flows to enrich the posterior approximation) to a vanilla VAE \cite{kingma_adam_2014}, a VAE with a VAMP prior \cite{tomczak_vae_2018} and models incorporating temporal coherence such as the BubbleVAE \cite{hyvarinen2004unifying} or the Topographic VAE (TVAE) \cite{keller2021topographic}. For the latter, we consider several models with different temporal coherence length $L$: TVAE (0) \emph{i.e.} no temporal coherence, TVAE (short) \emph{i.e.} $L \approx \frac{1}{8}$ of the input sequence length $S$, TVAE (part) where $L\approx \frac{1}{4}S$ and TVAE (half) with $L= \frac{1}{2}S$, \emph{i.e.} the model takes into account the full sequence. We also compare our model to a VAE using a Gaussian Process as prior (GPVAE) proposed in \cite{fortuin2020gp}. For this model, we consider several GP kernels (RBF, Cauchy, diffusion and matern). We use the 5 datasets presented in the previous section and train all the models on a train set, keep the best model on a validation set and compute the negative log joint-likelihood on an independent test set using 100 importance samples in a similar fashion as \cite{burda_importance_2016}. We train the models for 200 epochs for \emph{sprites} and \emph{rotMNIST}, 250 for \textit{colorMNIST} and 400 for \emph{starmen} and \emph{chairs} with a latent dimension set to 16 for all datasets but for the \emph{3d chairs} dataset where it is set to 32. Any other relevant piece of information about training configurations is provided in Appendix \ref{app: experiement details}. 
We show in Table~\ref{tab: nll}, the mean and standard deviation of the negative log joint likelihood obtained with 5 independent runs. For all datasets, the model is able to either compete or outperform the competitors. Moreover, as expected, using a richer prior (VAMP) or enriching the expressiveness of the variational posterior with flows (IAF) leads most of the time to a better likelihood estimation. This is an encouraging aspect since it shows that the model can be improved pretty easily using independent bricks available in the variational inference literature.

\begin{figure}[t]
    \captionsetup[subfigure]{position=below, labelformat = empty}
    \centering
    \subfloat{\includegraphics[width=0.67\linewidth]{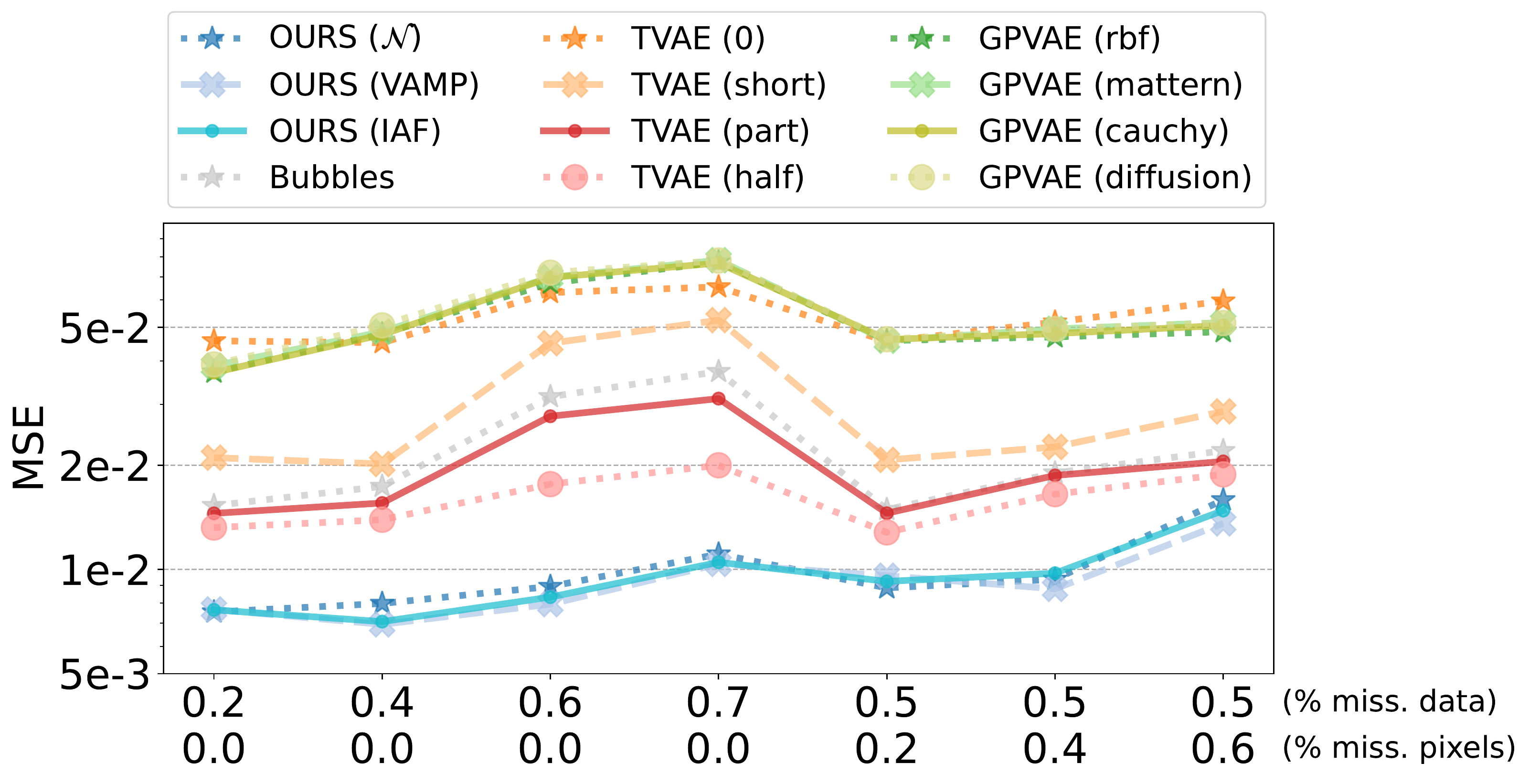}}\\\vspace{-1em}
    \subfloat{\includegraphics[width=0.67\linewidth]{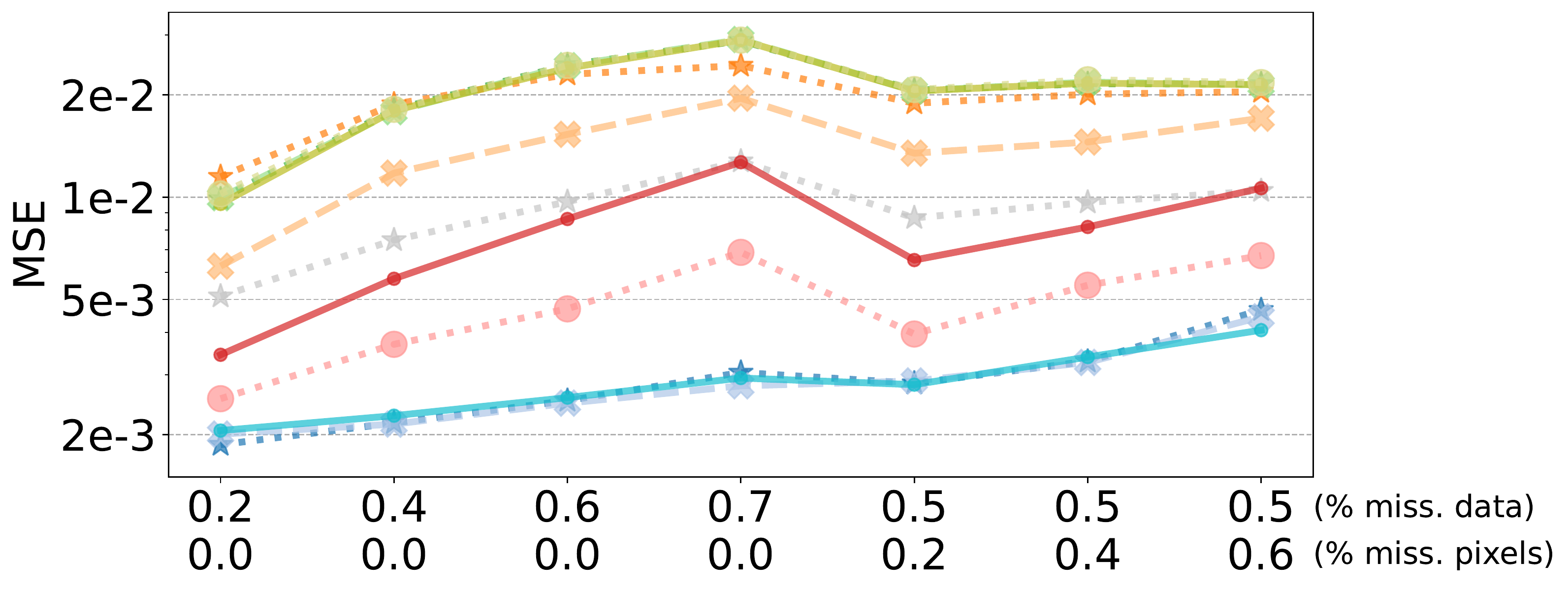}}
    \caption{Mean Square Error (MSE) on the test data for different proportions of missing observations (0.2 to 0.7) and missing pixels (0.2 to 0.6) in the input train, validation and test sequences for the \emph{starmen} (top) and \emph{sprites} (bottom) datasets. The proposed model appears very robust to incomplete sequences thanks to the flows-based structure.}
    \label{fig:missing data}
\end{figure}

\subsection{Missing Data Imputation}

The second experiment that we conduct consists in assessing the robustness of the model when it faces missing data and test its ability to impute missing values. To do so we consider 2 databases: \emph{starmen} and \emph{sprites}; and randomly remove observations in input sequences with probability 0.2, 0.4, 0.6 and 0.7. To challenge the model in the context of missing features, we also create sequences with missing observations (randomly removed with probability 0.5) and missing pixels in the observed images (randomly removed with probability 0.2, 0.4 and 0.6). All the models are trained with the same masks and are optimised using an objective computed only on the seen pixels. The charts in Fig.~\ref{fig:missing data} show the Mean Square Error (MSE) obtained on an independent test set. In all scenarios, the proposed model outperforms the TVAEs, BubbleVAEs and GPVAEs and appears as expected quite robust to missing observations in the input sequences. This is made possible thanks to the training structure that uses only one seen observation to perform variational inference.

\begin{figure}[t]
    \captionsetup[subfigure]{position=below, labelformat = empty}
    \subfloat{\includegraphics[width=1.6in]{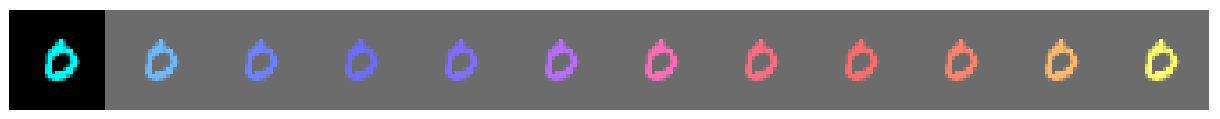}}
    \subfloat{\includegraphics[width=1.6in]{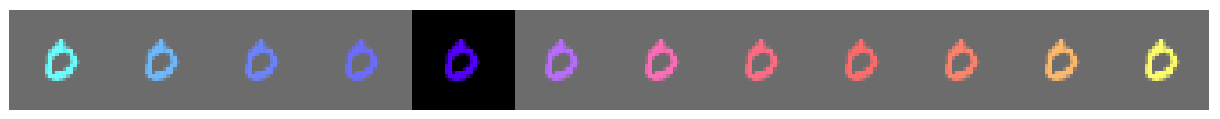}}\hfill
    \subfloat{\includegraphics[width=1.6in]{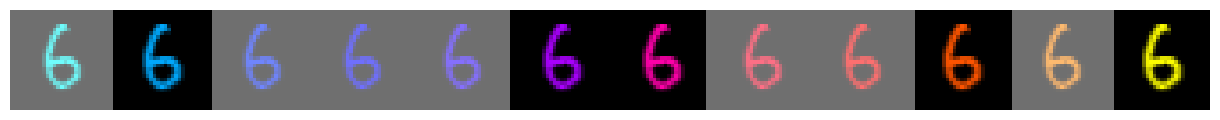}}
    \subfloat{\includegraphics[width=1.6in]{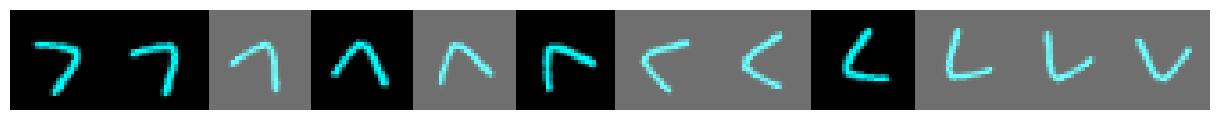}}\vspace{-1em}
    \subfloat[(a)]{\includegraphics[width=1.6in]{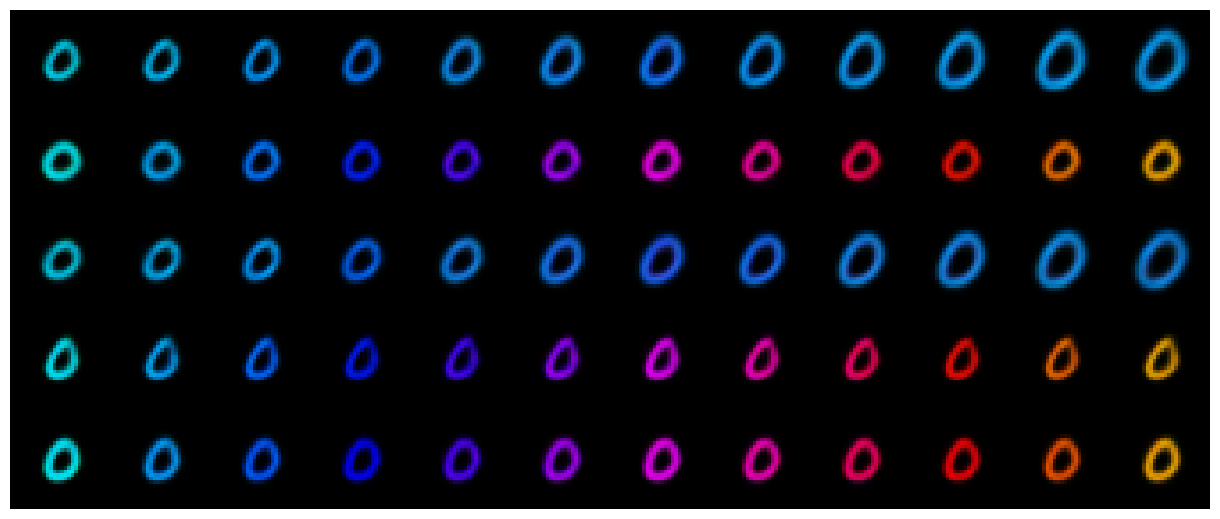}}
    \subfloat[(b)]{\includegraphics[width=1.6in]{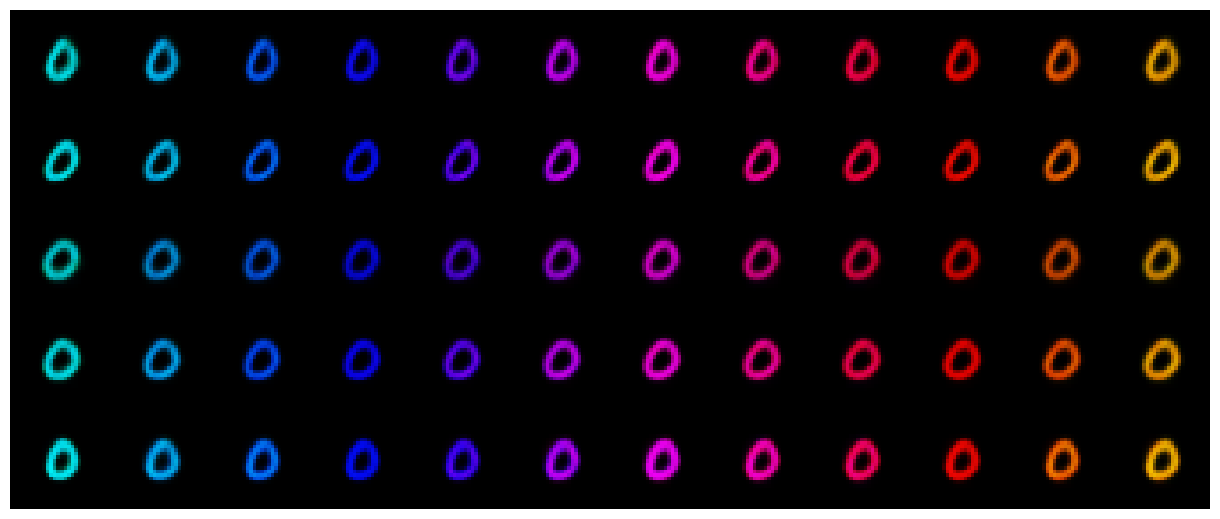}}\hfill
    \subfloat[(a)]{\includegraphics[width=1.6in]{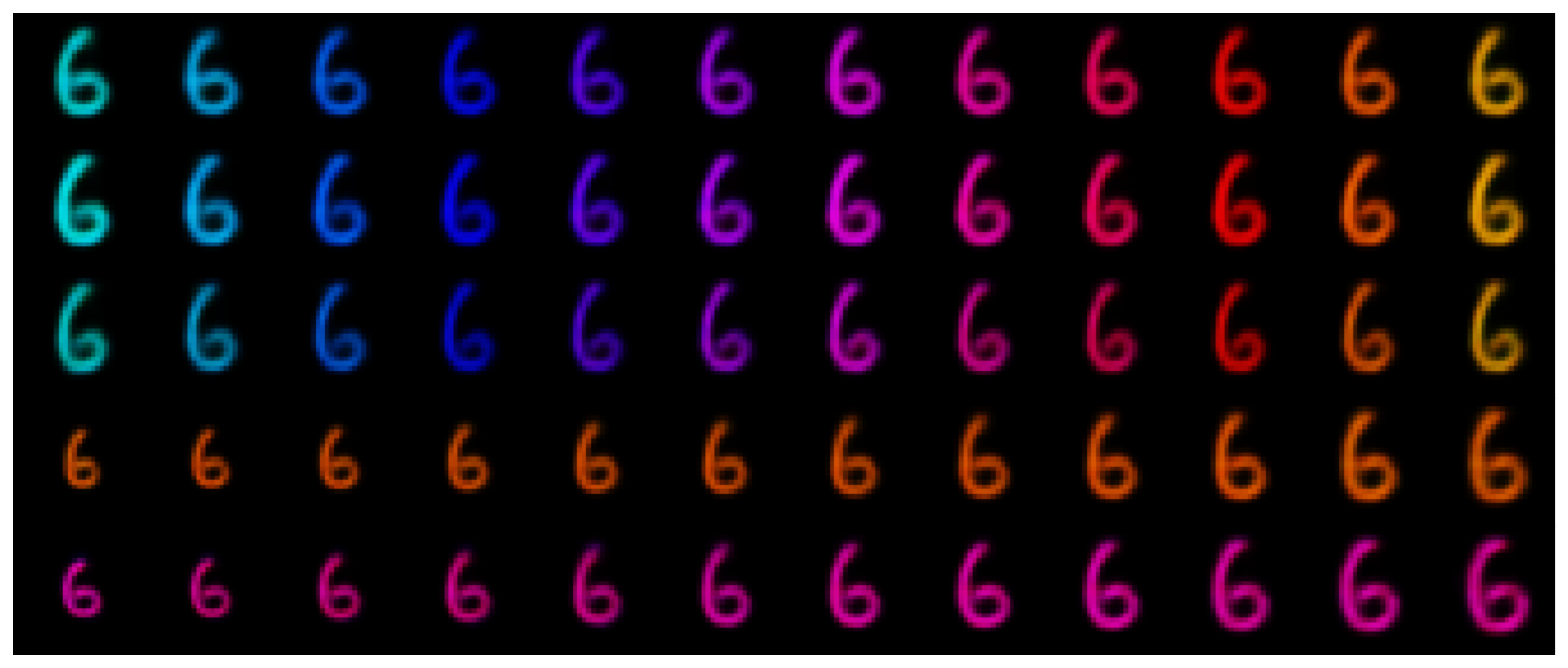}}
    \subfloat[(b)]{\includegraphics[width=1.6in]{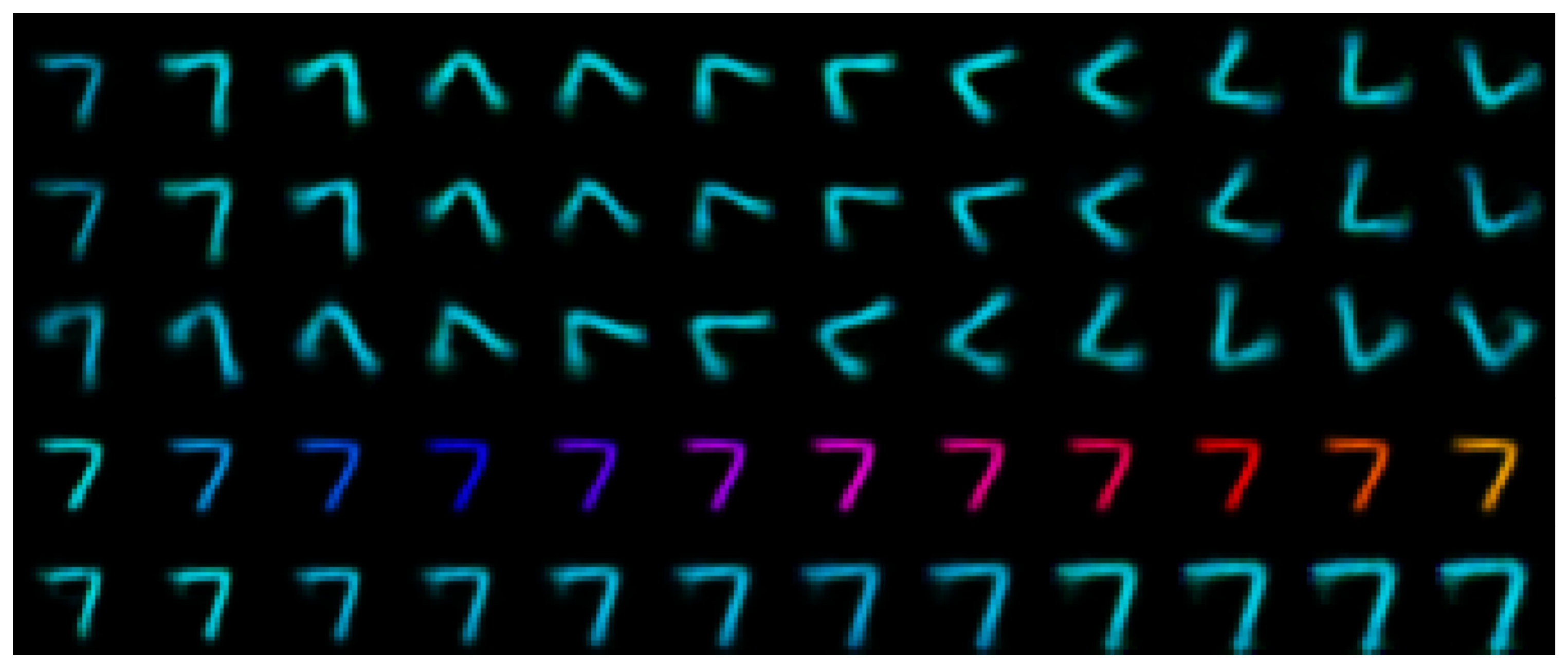}}
    \caption{Conditionally generated trajectories (greyed are unseen data). \emph{Left:} 5 generated sequences using the same input image. For each trajectory, 5 latent variables are drawn from the posterior distribution $q_{\phi}(z|x)$, passed trough the flows and decoded using $p_{\theta}(x|x)$. In a), the model is able to produce possible evolutions (changes of color or scale) for the dataset considered. \emph{Right:} Generated sequences using each seen data in the input sequence. The generated sequences are ranked as they maximise the likelihood on the seen data according to Eq.~\eqref{eq: missing optimal} (best at the top).}
    \label{fig:conditional generatoin}
\end{figure}

In Fig.~\ref{fig:conditional generatoin}, we also show some conditional generations obtained with the proposed model on the \emph{colorMNIST} dataset. At the top, we show 5 generated trajectories using 2 different images. In each case, we draw 5 random latent variables from the corresponding variational posterior $q_{\phi}(z|x)$. They are then passed through the flows according to Eq.~\eqref{eq: time dep} leading to 5 sequences and finally decoded using $p_{\theta}(x|z)$. In a), the model is able to produce a range of possible evolutions (changes of color or scale) that are plausible given the dataset considered. This is a very important property of the model since thanks to the variational posterior distribution it can generate an infinite number of possible trajectories from a single observation. Moreover, we see that the model is clearly able to keep the shape coherence all along the trajectory. At the bottom, we show the sequences obtained by using each image available in the sequence (not greyed). We rank the generated trajectories as they maximise the likelihood on the seen data (\emph{i.e.} according to Eq.~\eqref{eq: missing optimal}). This experiment shows how the model can benefit from the information available in the sequence despite only using one image to generate. In practice, one may generate as many trajectories as desired for each image available in the sequence (and not just one as in this example) and chose the one that maximises Eq.~\eqref{eq: missing optimal}. As a conclusion, these experiments show that even-though the relation between the latent variables of a sequence is deterministic, the stochasticity in the conditionally generated sequences arises from the sampling from the variational posterior that is able to capture the modularity of the data.

\begin{figure*}[t]
    \captionsetup[subfigure]{position=below, labelformat = empty}
    \begin{minipage}[c]{.71\linewidth}
    \subfloat{\includegraphics[width=.44\linewidth]{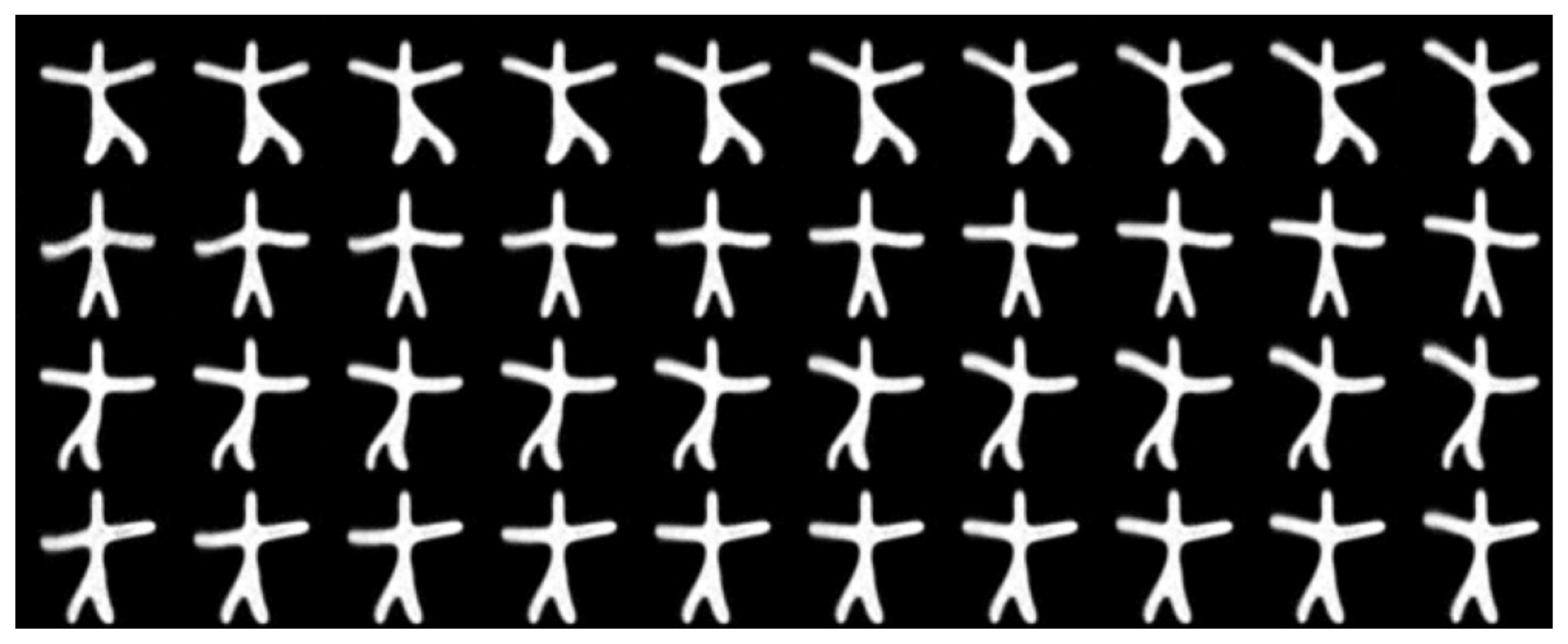}}
    \subfloat{\includegraphics[width=.54\linewidth]{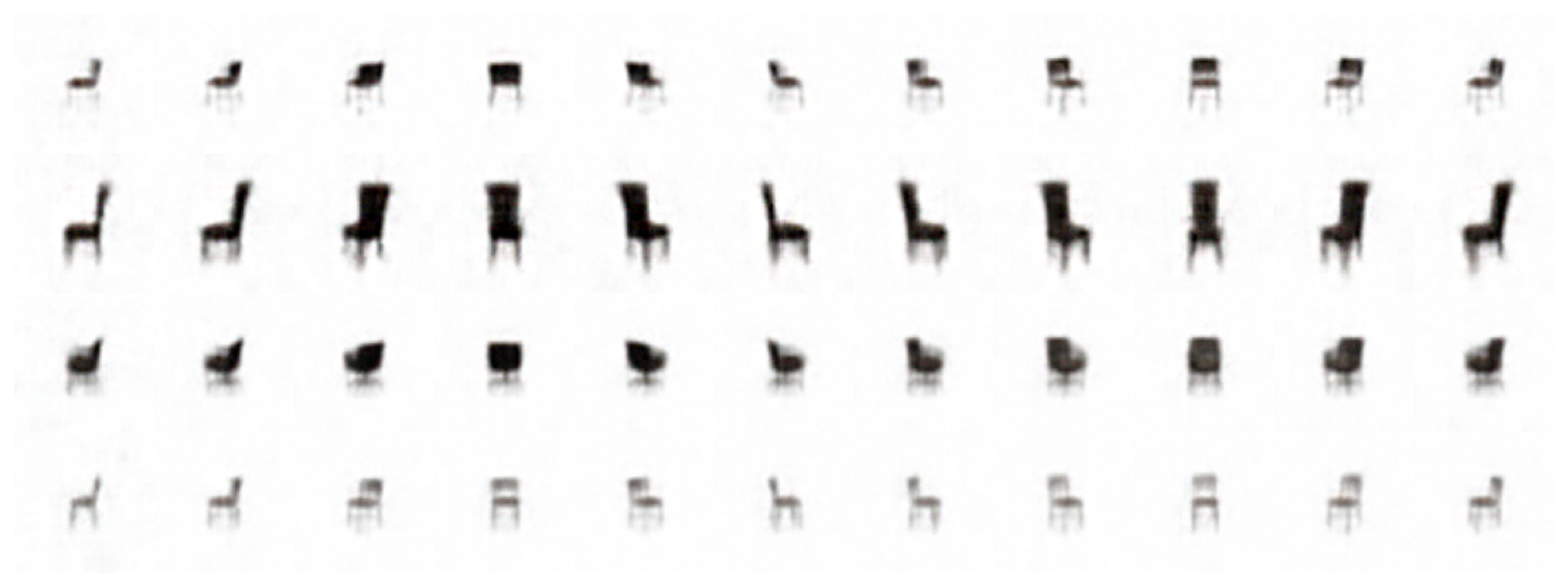}}\vspace{-1.2em}
    \subfloat{\includegraphics[width=0.44\linewidth]{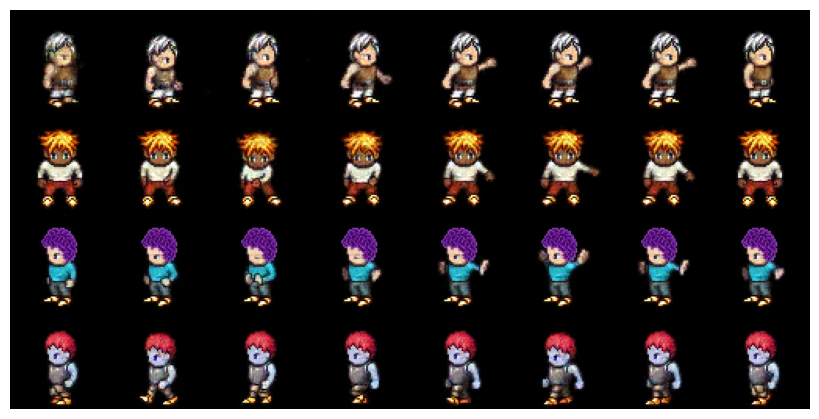}}
    \subfloat{\includegraphics[width=0.54\linewidth]{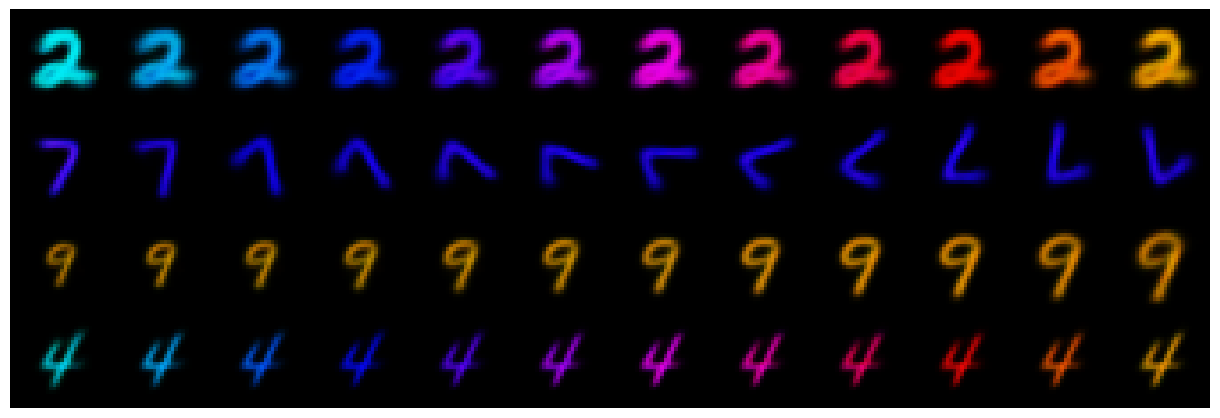}}
    \end{minipage}
    \hspace{-1.5em}
    \begin{minipage}[c]{.28\linewidth}\vspace{1.2em}
    \subfloat{\includegraphics[scale=0.49]{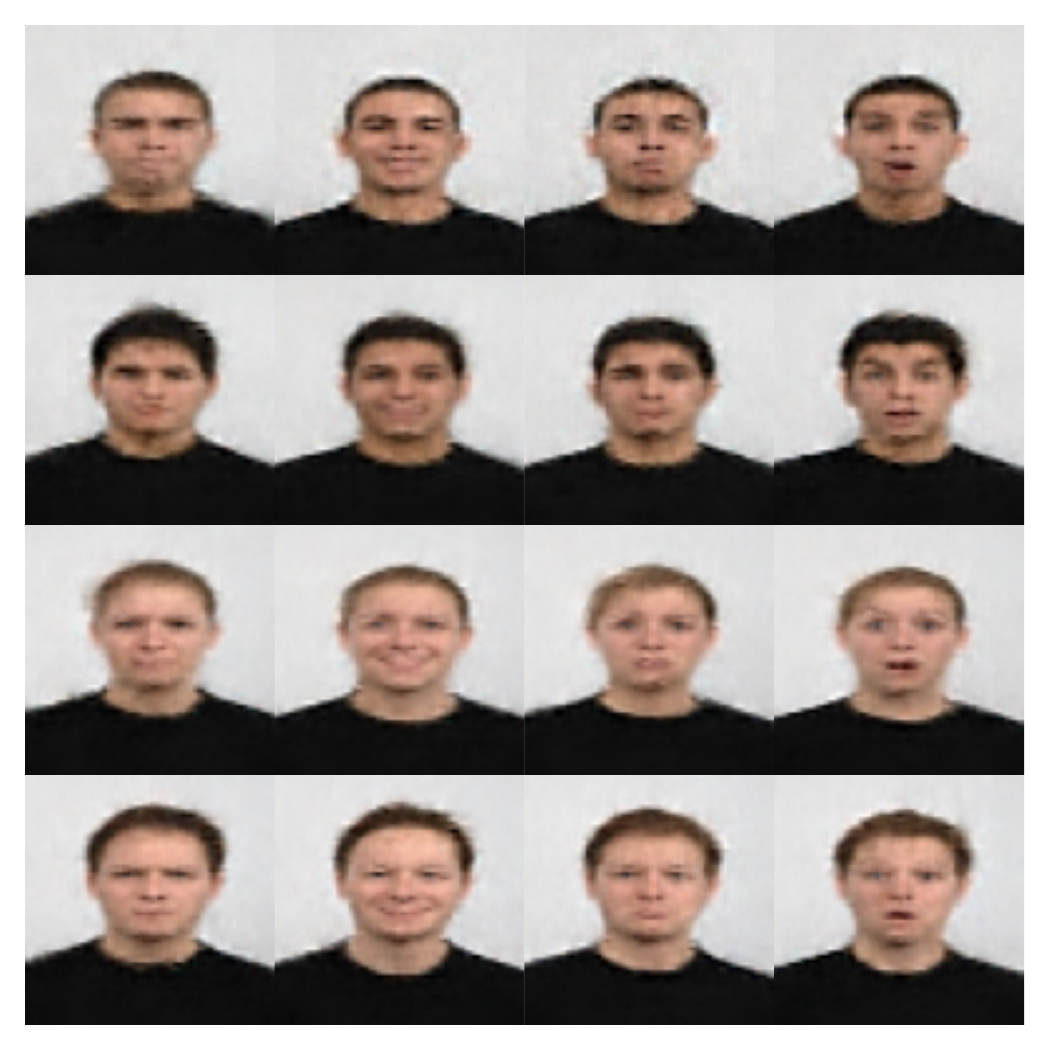}}
    \end{minipage}
    \caption{Generated sequences using the proposed model. Latent variables are sampled from the prior distribution (taken as a standard Gaussian in this example) and propagated through the flows according to Eq.~\eqref{eq: time dep}. The obtained latent sequences are then decoded using the conditional distribution $p_{\theta}(x|z)$ to create the image sequences.}
    \label{fig:generated sequences}
\end{figure*}

\subsection{Unconditional Sequence Generation}
\begin{figure}[t]
    \centering
    \adjustbox{minipage=1.7em,raise=\dimexpr 1.3\height}{\small Gen.\vspace{1.2em}
    Train}
   \subfloat{\includegraphics[width=0.405\linewidth]{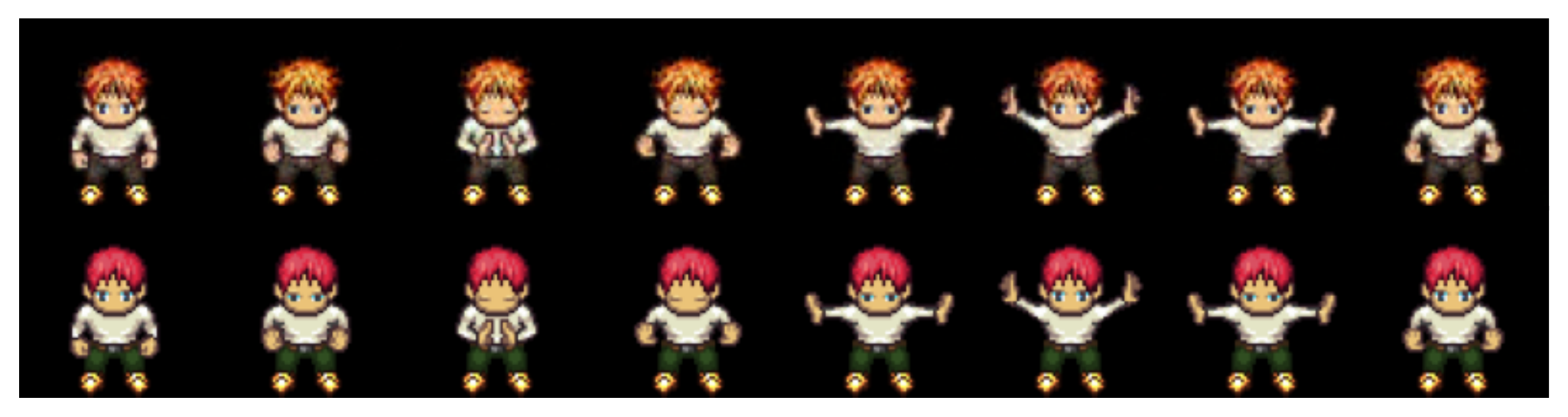}}
   \subfloat{\includegraphics[width=0.51\linewidth]{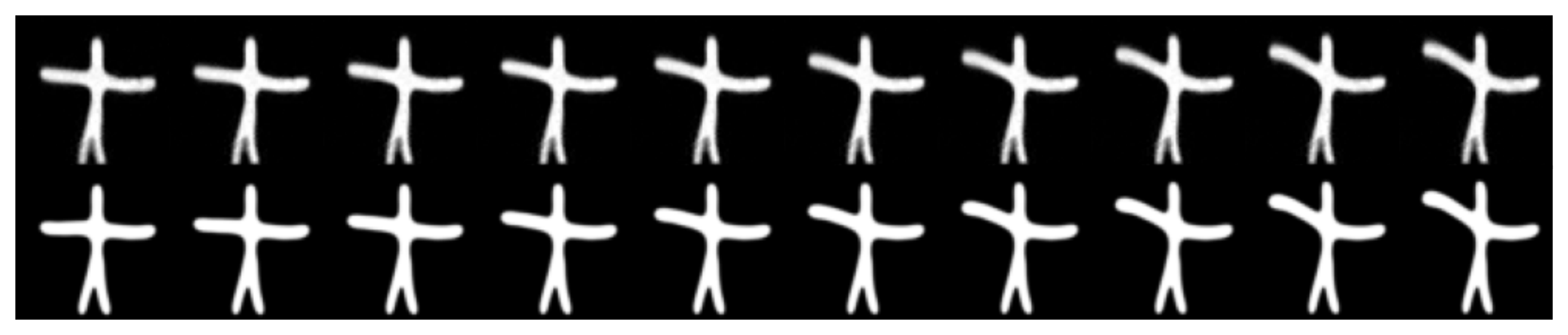}}
   \\\vspace{-1em}
   \adjustbox{minipage=1.7em,raise=\dimexpr 1.5\height}{\small Gen.\vspace{5em}
    Train}
    \subfloat{\includegraphics[width=0.46075\linewidth]{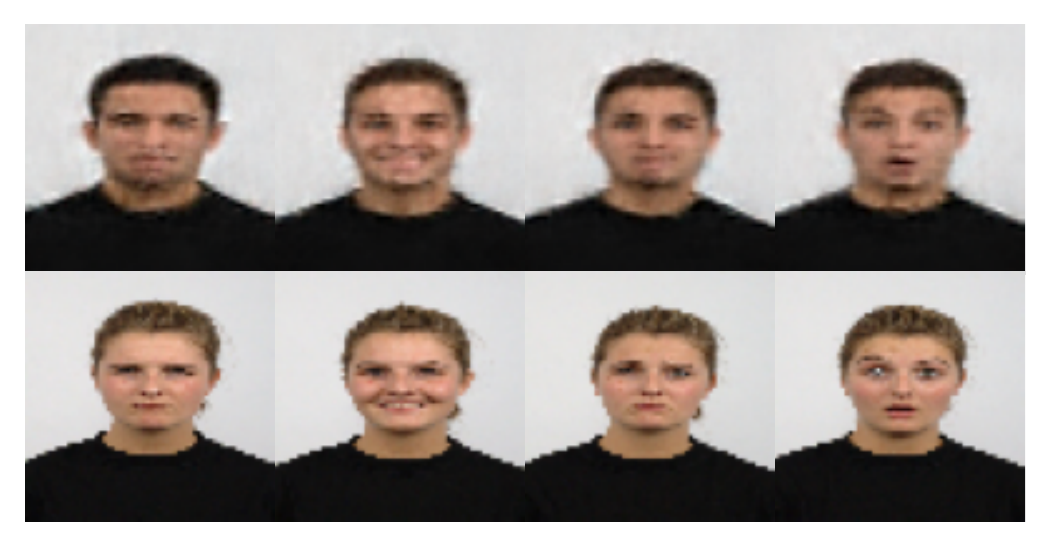}}
    \subfloat{\includegraphics[width=0.46075\linewidth]{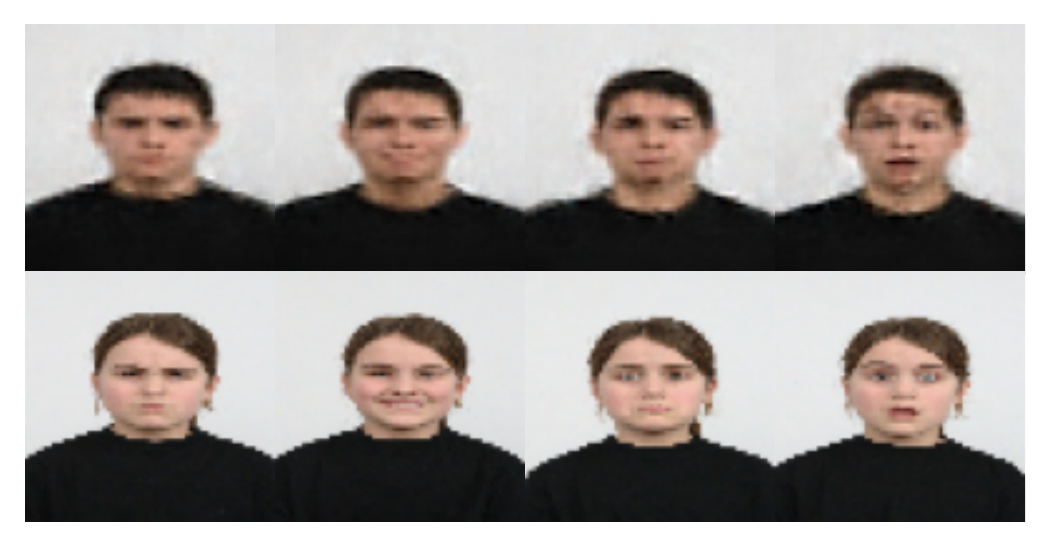}}
    \caption{Closest train sequences (train) to the generated ones (gen.). See more examples in Appendix~\ref{app: overfitting}.}
    \label{fig:closest samples}
\end{figure}

In this section, we evaluate the ability of the proposed model to generate relevant fully synthetic trajectories. For this experiment, we first compute the Frechet Inception Distance (FID) \citep{heusel_gans_2017} on the \emph{colorMNIST} and \emph{sprites} datasets. The FID is computed by generating the same number of images as available in an independent test set: 21,312 for \emph{sprites} (2,664 sequences of 8 time steps) and 120,000 for \emph{colorMNIST}. Note that in this setting the FID does not account for the temporal coherence between the generated samples within a sequence. As shown in Table~\ref{tab:fid}, the proposed model achieves the lowest FID (lower is better). The fact that it is able to outperform a VAE or a VAMP-VAE shows that the temporal coherence constraint imposed by the flows does not affect the quality of the generated images. Moreover, we see the influence of using a more complex prior or enriching the variational approximation on the generative capability of the model that can achieve better FIDs. Finally, we show generated samples for the \emph{3d chairs}, \emph{starmen}, \emph{sprites}, \emph{colorMNIST} and the Radboud Faces Database consisting of 67 individuals displaying different emotions \cite{langner2010presentation}. For the latter dataset, we create sequences of 4 time steps corresponding to the emotions: \emph{anger}, \emph{happiness}, \emph{sadness} and \textit{surprise}; and down-sample the images so they are of size 64x64. We show 4 generated sequences for each dataset in Fig.~\ref{fig:generated sequences}. Thanks to the flow-based structure, the model is able to generate relevant sequences that clearly keep a temporal consistency. Additional samples can be found in Appendix \ref{app: more generations} and small movies in the supplementary materials. We also show that the proposed model does not simply \emph{memorise} the training samples by showing the closest training sequence to the generated ones in Fig.~\ref{fig:closest samples} and Appendix \ref{app: overfitting}. %These experiments show that the method is able to generate diverse and relevant sequences. 
\begin{table}[t]
\caption{FID (lower is better) computed on an independent test set with the same number of generated samples as available in the test set.}
    \centering
    \begin{sc}
    \begin{small}
    \begin{tabular}{l|ccc}
    \toprule
        Model & ColorMNIST & Sprites \\
        \midrule
        VAE         & 29.79 & 53.37 \\
        VAMP        & 33.92 & 59.85 \\
        GPVAE       & 31.93 & 56.74 \\
        Ours ($\mathcal{N}$) & 28.62 & 44.82 \\
        Ours (VAMP) & \textbf{25.07} & \textbf{40.23}\\
        Ours (IAF) & 28.14 & 41.81\\
        \bottomrule
    \end{tabular}
    \end{small}
    \end{sc}
    \label{tab:fid}
    \vskip -0.1in
\end{table}
\section{Conclusion}
In this paper, we introduced a new generative model for longitudinal data that relies on variational inference and normalizing flows. It proved able to generate relevant fully synthetic sequences and to propose plausible trajectories when conditioned on one or several seen samples in an input sequence. We also discussed and showed that our model can benefit from improvements proposed in the variational inference literature. In particular, we proposed two variants of our model using either a more complex prior or a more flexible variational posterior using flows. These independent enhancements revealed particularly useful for likelihood estimation and unconditional generations. Moreover, the proposed model demonstrated quite a good robustness to missing data and showed to be useful for missing data imputation. Nonetheless, a potential \textit{weakness} of the model is that the flow-based structure makes it discrete. Furthermore, we acknowledge that the deterministic aspect induced by the choice of normalizing flows to account for time dependency in the latent space can be seen as a limitation to the expressiveness of the model. In particular, future work may involve stochastic trajectories in a spirit similar to latent SDEs, which would add to the expressiveness of the model. Nevertheless, such determinism in trajectories combined with the proposed training scheme may also benefit the posterior variational distributions that are constrained to be sufficiently expressive to capture the stochasticity of the trajectories.

\clearpage

\bibliography{references}

\begin{thebibliography}{84}
\providecommand{\natexlab}[1]{#1}
\providecommand{\url}[1]{\texttt{#1}}
\expandafter\ifx\csname urlstyle\endcsname\relax
  \providecommand{\doi}[1]{doi: #1}\else
  \providecommand{\doi}{doi: \begingroup \urlstyle{rm}\Url}\fi

\bibitem[Aghili et~al.(2018)Aghili, Tabarestani, Adjouadi, and
  Adeli]{aghili2018predictive}
Aghili, M., Tabarestani, S., Adjouadi, M., and Adeli, E.
\newblock Predictive modeling of longitudinal data for alzheimer's disease
  diagnosis using rnns.
\newblock In \emph{PRedictive Intelligence in MEdicine}, pp.\  112--119, Cham,
  2018. Springer International Publishing.

\bibitem[Aneja et~al.(2020)Aneja, Schwing, Kautz, and
  Vahdat]{aneja_ncp-vae_2020}
Aneja, J., Schwing, A., Kautz, J., and Vahdat, A.
\newblock {NCP}-{VAE}: Variational autoencoders with noise contrastive priors.
\newblock \emph{{arXiv}:2010.02917 [cs, stat]}, 2020.

\bibitem[Arvanitidis et~al.(2018)Arvanitidis, Hansen, and
  Hauberg]{arvanitidis_latent_2018}
Arvanitidis, G., Hansen, L.~K., and Hauberg, S.
\newblock Latent space oddity: On the curvature of deep generative models.
\newblock In \emph{6th International Conference on Learning Representations,
  ICLR 2018}, 2018.

\bibitem[Aubry et~al.(2014)Aubry, Maturana, Efros, Russell, and
  Sivic]{aubry2014seeing}
Aubry, M., Maturana, D., Efros, A.~A., Russell, B.~C., and Sivic, J.
\newblock Seeing 3d chairs: exemplar part-based 2d-3d alignment using a large
  dataset of cad models.
\newblock In \emph{Proceedings of the IEEE conference on computer vision and
  pattern recognition}, pp.\  3762--3769, 2014.

\bibitem[Bernal-Rusiel et~al.(2013)Bernal-Rusiel, Greve, Reuter, Fischl,
  Sabuncu, Initiative, et~al.]{bernal2013statistical}
Bernal-Rusiel, J.~L., Greve, D.~N., Reuter, M., Fischl, B., Sabuncu, M.~R.,
  Initiative, A. D.~N., et~al.
\newblock Statistical analysis of longitudinal neuroimage data with linear
  mixed effects models.
\newblock \emph{Neuroimage}, 66:\penalty0 249--260, 2013.

\bibitem[Blackledge et~al.(2014)Blackledge, Collins, Tunariu, Orton, Padhani,
  Leach, and Koh]{blackledge2014assessment}
Blackledge, M.~D., Collins, D.~J., Tunariu, N., Orton, M.~R., Padhani, A.~R.,
  Leach, M.~O., and Koh, D.-M.
\newblock Assessment of treatment response by total tumor volume and global
  apparent diffusion coefficient using diffusion-weighted mri in patients with
  metastatic bone disease: A feasibility study.
\newblock \emph{PLOS ONE}, 9\penalty0 (4):\penalty0 e91779, 2014.

\bibitem[Blei et~al.(2017)Blei, Kucukelbir, and McAuliffe]{blei2017variational}
Blei, D.~M., Kucukelbir, A., and McAuliffe, J.~D.
\newblock Variational inference: A review for statisticians.
\newblock \emph{Journal of the American statistical Association}, 112\penalty0
  (518):\penalty0 859--877, 2017.

\bibitem[B{\^o}ne et~al.(2018)B{\^o}ne, Colliot, and
  Durrleman]{bone2018learning}
B{\^o}ne, A., Colliot, O., and Durrleman, S.
\newblock Learning distributions of shape trajectories from longitudinal
  datasets: a hierarchical model on a manifold of diffeomorphisms.
\newblock In \emph{Proceedings of the IEEE conference on computer vision and
  pattern recognition}, pp.\  9271--9280, 2018.

\bibitem[Burda et~al.(2016)Burda, Grosse, and
  Salakhutdinov]{burda_importance_2016}
Burda, Y., Grosse, R., and Salakhutdinov, R.
\newblock Importance weighted autoencoders.
\newblock \emph{{arXiv}:1509.00519 [cs, stat]}, 2016.

\bibitem[Burgess(2018)]{burgess2018vae}
Burgess, C. P. e.~a.
\newblock Understanding disentangling in $\beta$-vae.
\newblock \emph{arXiv preprint arXiv:1804.03599}, 2018.

\bibitem[Cao et~al.(2018)Cao, Wang, Li, Zhou, Li, and Li]{cao2018brits}
Cao, W., Wang, D., Li, J., Zhou, H., Li, L., and Li, Y.
\newblock Brits: Bidirectional recurrent imputation for time series.
\newblock \emph{Advances in neural information processing systems}, 31, 2018.

\bibitem[Casale et~al.(2018)Casale, Dalca, Saglietti, Listgarten, and
  Fusi]{casale2018gaussian}
Casale, F.~P., Dalca, A., Saglietti, L., Listgarten, J., and Fusi, N.
\newblock Gaussian process prior variational autoencoders.
\newblock \emph{Advances in neural information processing systems}, 31, 2018.

\bibitem[Caterini et~al.(2018)Caterini, Doucet, and
  Sejdinovic]{caterini_hamiltonian_2018}
Caterini, A.~L., Doucet, A., and Sejdinovic, D.
\newblock Hamiltonian variational auto-encoder.
\newblock In \emph{Advances in Neural Information Processing Systems}, pp.\
  8167--8177, 2018.

\bibitem[Chadebec \& Allassonni{\`e}re(2022)Chadebec and
  Allassonni{\`e}re]{chadebec2022geometric}
Chadebec, C. and Allassonni{\`e}re, S.
\newblock A geometric perspective on variational autoencoders.
\newblock \emph{Advances in Neural Information Processing Systems}, 2022.

\bibitem[Chadebec et~al.(2022{\natexlab{a}})Chadebec, Thibeau-Sutre, Burgos,
  and Allassonni{\`e}re]{chadebec_data_2021}
Chadebec, C., Thibeau-Sutre, E., Burgos, N., and Allassonni{\`e}re, S.
\newblock Data augmentation in high dimensional low sample size setting using a
  geometry-based variational autoencoder.
\newblock \emph{IEEE Transactions on Pattern Analysis and Machine
  Intelligence}, 2022{\natexlab{a}}.

\bibitem[Chadebec et~al.(2022{\natexlab{b}})Chadebec, Vincent, and
  Allassonni{\`e}re]{chadebec2022pythae}
Chadebec, C., Vincent, L.~J., and Allassonni{\`e}re, S.
\newblock Pythae: Unifying generative autoencoders in python--a benchmarking
  use case.
\newblock \emph{Proceedings of the Neural Information Processing Systems Track
  on Datasets and Benchmarks}, 2022{\natexlab{b}}.

\bibitem[Chen et~al.(2018{\natexlab{a}})Chen, Klushyn, Kurle, Jiang, Bayer, and
  Smagt]{chen_metrics_2018}
Chen, N., Klushyn, A., Kurle, R., Jiang, X., Bayer, J., and Smagt, P.
\newblock Metrics for deep generative models.
\newblock In \emph{International Conference on Artificial Intelligence and
  Statistics}, pp.\  1540--1550. PMLR, 2018{\natexlab{a}}.

\bibitem[Chen et~al.(2018{\natexlab{b}})Chen, Li, Grosse, and
  Duvenaud]{chen2019vae}
Chen, R.~T., Li, X., Grosse, R.~B., and Duvenaud, D.~K.
\newblock Isolating sources of disentanglement in variational autoencoders.
\newblock \emph{Advances in neural information processing systems}, 31,
  2018{\natexlab{b}}.

\bibitem[Chen et~al.(2018{\natexlab{c}})Chen, Rubanova, Bettencourt, and
  Duvenaud]{chen2018neural}
Chen, R.~T., Rubanova, Y., Bettencourt, J., and Duvenaud, D.~K.
\newblock Neural ordinary differential equations.
\newblock \emph{Advances in neural information processing systems}, 31,
  2018{\natexlab{c}}.

\bibitem[Chen et~al.(2016)Chen, Kingma, Salimans, Duan, Dhariwal, Schulman,
  Sutskever, and Abbeel]{chen_variational_2016}
Chen, X., Kingma, D.~P., Salimans, T., Duan, Y., Dhariwal, P., Schulman, J.,
  Sutskever, I., and Abbeel, P.
\newblock Variational lossy autoencoder.
\newblock \emph{{arXiv} preprint {arXiv}:1611.02731}, 2016.

\bibitem[Chung et~al.(2015)Chung, Kastner, Dinh, Goel, Courville, and
  Bengio]{chung2015recurrent}
Chung, J., Kastner, K., Dinh, L., Goel, K., Courville, A.~C., and Bengio, Y.
\newblock A recurrent latent variable model for sequential data.
\newblock \emph{Advances in neural information processing systems}, 28, 2015.

\bibitem[Davidson et~al.(2018)Davidson, Falorsi, De~Cao, Kipf, and
  Tomczak]{davidson_hyperspherical_2018}
Davidson, T.~R., Falorsi, L., De~Cao, N., Kipf, T., and Tomczak, J.~M.
\newblock Hyperspherical variational auto-encoders.
\newblock In \emph{34th Conference on Uncertainty in Artificial Intelligence
  2018, UAI 2018}, pp.\  856--865. Association For Uncertainty in Artificial
  Intelligence (AUAI), 2018.

\bibitem[Diggle et~al.(2002)Diggle, Diggle, Heagerty, Liang, Zeger,
  et~al.]{diggle2002analysis}
Diggle, P., Diggle, P.~J., Heagerty, P., Liang, K.-Y., Zeger, S., et~al.
\newblock \emph{Analysis of longitudinal data}.
\newblock Oxford university press, 2002.

\bibitem[Dilokthanakul et~al.(2017)Dilokthanakul, Mediano, Garnelo, Lee,
  Salimbeni, Arulkumaran, and Shanahan]{dilokthanakul_deep_2017}
Dilokthanakul, N., Mediano, P. A.~M., Garnelo, M., Lee, M. C.~H., Salimbeni,
  H., Arulkumaran, K., and Shanahan, M.
\newblock Deep unsupervised clustering with gaussian mixture variational
  autoencoders.
\newblock \emph{{arXiv}:1611.02648 [cs, stat]}, 2017.

\bibitem[Dinh et~al.(2014)Dinh, Krueger, and Bengio]{dinh2014nice}
Dinh, L., Krueger, D., and Bengio, Y.
\newblock Nice: Non-linear independent components estimation.
\newblock \emph{arXiv preprint arXiv:1410.8516}, 2014.

\bibitem[Dinh et~al.(2016)Dinh, Sohl-Dickstein, and Bengio]{dinh2016density}
Dinh, L., Sohl-Dickstein, J., and Bengio, S.
\newblock Density estimation using real nvp.
\newblock \emph{arXiv preprint arXiv:1605.08803}, 2016.

\bibitem[Dupont et~al.(2019)Dupont, Doucet, and Teh]{dupont2019augmented}
Dupont, E., Doucet, A., and Teh, Y.~W.
\newblock Augmented neural odes.
\newblock \emph{Advances in neural information processing systems}, 32, 2019.

\bibitem[Falorsi et~al.(2018)Falorsi, de~Haan, Davidson, De~Cao, Weiler,
  Forré, and Cohen]{falorsi_explorations_2018}
Falorsi, L., de~Haan, P., Davidson, T.~R., De~Cao, N., Weiler, M., Forré, P.,
  and Cohen, T.~S.
\newblock Explorations in homeomorphic variational auto-encoding.
\newblock \emph{{arXiv}:1807.04689 [cs, stat]}, 2018.

\bibitem[Fonteijn et~al.(2012)Fonteijn, Modat, Clarkson, Barnes, Lehmann,
  Hobbs, Scahill, Tabrizi, Ourselin, Fox, et~al.]{fonteijn2012event}
Fonteijn, H.~M., Modat, M., Clarkson, M.~J., Barnes, J., Lehmann, M., Hobbs,
  N.~Z., Scahill, R.~I., Tabrizi, S.~J., Ourselin, S., Fox, N.~C., et~al.
\newblock An event-based model for disease progression and its application in
  familial alzheimer's disease and huntington's disease.
\newblock \emph{NeuroImage}, 60\penalty0 (3):\penalty0 1880--1889, 2012.

\bibitem[Fortuin et~al.(2020)Fortuin, Baranchuk, R{\"a}tsch, and
  Mandt]{fortuin2020gp}
Fortuin, V., Baranchuk, D., R{\"a}tsch, G., and Mandt, S.
\newblock Gp-vae: Deep probabilistic time series imputation.
\newblock In \emph{International conference on artificial intelligence and
  statistics}, pp.\  1651--1661. PMLR, 2020.

\bibitem[Germain et~al.(2015)Germain, Gregor, Murray, and
  Larochelle]{germain2015made}
Germain, M., Gregor, K., Murray, I., and Larochelle, H.
\newblock Made: Masked autoencoder for distribution estimation.
\newblock In \emph{International Conference on Machine Learning}, pp.\
  881--889. PMLR, 2015.

\bibitem[Heusel et~al.(2017)Heusel, Ramsauer, Unterthiner, Nessler, and
  Hochreiter]{heusel_gans_2017}
Heusel, M., Ramsauer, H., Unterthiner, T., Nessler, B., and Hochreiter, S.
\newblock Gans trained by a two time-scale update rule converge to a local nash
  equilibrium.
\newblock In \emph{Advances in Neural Information Processing Systems}, 2017.

\bibitem[Higgins et~al.(2017)Higgins, Matthey, Pal, Burgess, Glorot, Botvinick,
  Mohamed, and Lerchner]{higgins_beta-vae_2017}
Higgins, I., Matthey, L., Pal, A., Burgess, C., Glorot, X., Botvinick, M.,
  Mohamed, S., and Lerchner, A.
\newblock beta-{VAE}: Learning basic visual concepts with a constrained
  variational framework.
\newblock \emph{{ICLR}}, 2\penalty0 (5):\penalty0 6, 2017.

\bibitem[Hoffman \& Johnson(2016)Hoffman and Johnson]{hoffman_elbo_2016}
Hoffman, M.~D. and Johnson, M.~J.
\newblock Elbo surgery: yet another way to carve up the variational evidence
  lower bound.
\newblock In \emph{Workshop in Advances in Approximate Bayesian Inference,
  {NIPS}}, volume~1, pp.\ ~2, 2016.

\bibitem[Hyv{\"a}rinen et~al.(2004)Hyv{\"a}rinen, Hurri, and
  V{\"a}yrynen]{hyvarinen2004unifying}
Hyv{\"a}rinen, A., Hurri, J., and V{\"a}yrynen, J.
\newblock A unifying framework for natural image statistics: spatiotemporal
  activity bubbles.
\newblock \emph{Neurocomputing}, 58:\penalty0 801--806, 2004.

\bibitem[Jedynak et~al.(2012)Jedynak, Lang, Liu, Katz, Zhang, Wyman, Raunig,
  Jedynak, Caffo, Prince, et~al.]{jedynak2012computational}
Jedynak, B.~M., Lang, A., Liu, B., Katz, E., Zhang, Y., Wyman, B.~T., Raunig,
  D., Jedynak, C.~P., Caffo, B., Prince, J.~L., et~al.
\newblock A computational neurodegenerative disease progression score: method
  and results with the alzheimer's disease neuroimaging initiative cohort.
\newblock \emph{Neuroimage}, 63\penalty0 (3):\penalty0 1478--1486, 2012.

\bibitem[Jordan et~al.(1999)Jordan, Ghahramani, Jaakkola, and
  Saul]{jordan_introduction_1999}
Jordan, M.~I., Ghahramani, Z., Jaakkola, T.~S., and Saul, L.~K.
\newblock An introduction to variational methods for graphical models.
\newblock \emph{Machine Learning}, 37\penalty0 (2):\penalty0 183--233, 1999.

\bibitem[Kalatzis et~al.(2020)Kalatzis, Eklund, Arvanitidis, and
  Hauberg]{kalatzis_variational_2020}
Kalatzis, D., Eklund, D., Arvanitidis, G., and Hauberg, S.
\newblock Variational autoencoders with riemannian brownian motion priors.
\newblock In \emph{International Conference on Machine Learning}, pp.\
  5053--5066. PMLR, 2020.

\bibitem[Kanaa et~al.(2021)Kanaa, Voleti, Kahou, and Pal]{kanaa2021simple}
Kanaa, D., Voleti, V., Kahou, S.~E., and Pal, C.
\newblock Simple video generation using neural odes.
\newblock \emph{arXiv preprint arXiv:2109.03292}, 2021.

\bibitem[Keller \& Welling(2021)Keller and Welling]{keller2021topographic}
Keller, T.~A. and Welling, M.
\newblock Topographic vaes learn equivariant capsules.
\newblock \emph{Advances in Neural Information Processing Systems},
  34:\penalty0 28585--28597, 2021.

\bibitem[Kim \& Mnih(2018)Kim and Mnih]{kim2018factorvae}
Kim, H. and Mnih, A.
\newblock Disentangling by factorising.
\newblock In \emph{International Conference on Machine Learning}, pp.\
  2649--2658. PMLR, 2018.

\bibitem[Kingma \& Ba(2014)Kingma and Ba]{kingma_adam_2014}
Kingma, D.~P. and Ba, J.
\newblock Adam: A method for stochastic optimization.
\newblock \emph{{arXiv} preprint {arXiv}:1412.6980}, 2014.

\bibitem[Kingma \& Welling(2014)Kingma and Welling]{kingma_auto-encoding_2014}
Kingma, D.~P. and Welling, M.
\newblock Auto-encoding variational bayes.
\newblock \emph{{arXiv}:1312.6114 [cs, stat]}, 2014.

\bibitem[Kingma et~al.(2016)Kingma, Salimans, Jozefowicz, Chen, Sutskever, and
  Welling]{kingma2016improved}
Kingma, D.~P., Salimans, T., Jozefowicz, R., Chen, X., Sutskever, I., and
  Welling, M.
\newblock Improved variational inference with inverse autoregressive flow.
\newblock \emph{Advances in neural information processing systems}, 29, 2016.

\bibitem[Klushyn et~al.(2021)Klushyn, Kurle, Soelch, Cseke, and van~der
  Smagt]{klushyn2021latent}
Klushyn, A., Kurle, R., Soelch, M., Cseke, B., and van~der Smagt, P.
\newblock Latent matters: Learning deep state-space models.
\newblock \emph{Advances in Neural Information Processing Systems},
  34:\penalty0 10234--10245, 2021.

\bibitem[Korkinof et~al.(2018)Korkinof, Rijken, O'Neill, Yearsley, Harvey, and
  Glocker]{korkinof_high-resolution_2018}
Korkinof, D., Rijken, T., O'Neill, M., Yearsley, J., Harvey, H., and Glocker,
  B.
\newblock High-resolution mammogram synthesis using progressive generative
  adversarial networks.
\newblock \emph{{arXiv} preprint {arXiv}:1807.03401}, 2018.

\bibitem[Koval et~al.(2017)Koval, Schiratti, Routier, Bacci, Colliot,
  Allassonni{\`e}re, Durrleman, Initiative, et~al.]{koval2017statistical}
Koval, I., Schiratti, J.-B., Routier, A., Bacci, M., Colliot, O.,
  Allassonni{\`e}re, S., Durrleman, S., Initiative, A. D.~N., et~al.
\newblock Statistical learning of spatiotemporal patterns from longitudinal
  manifold-valued networks.
\newblock In \emph{International conference on medical image computing and
  computer-assisted intervention}, pp.\  451--459. Springer, 2017.

\bibitem[Laird \& Ware(1982)Laird and Ware]{laird1982random}
Laird, N.~M. and Ware, J.~H.
\newblock Random-effects models for longitudinal data.
\newblock \emph{Biometrics}, pp.\  963--974, 1982.

\bibitem[Langner et~al.(2010)Langner, Dotsch, Bijlstra, Wigboldus, Hawk, and
  Van~Knippenberg]{langner2010presentation}
Langner, O., Dotsch, R., Bijlstra, G., Wigboldus, D.~H., Hawk, S.~T., and
  Van~Knippenberg, A.
\newblock Presentation and validation of the radboud faces database.
\newblock \emph{Cognition and emotion}, 24\penalty0 (8):\penalty0 1377--1388,
  2010.

\bibitem[{LeCun}(1998)]{lecun_mnist_1998}
{LeCun}, Y.
\newblock The {MNIST} database of handwritten digits.
\newblock 1998.

\bibitem[Li et~al.(2020)Li, Wong, Chen, and Duvenaud]{li2020scalable}
Li, X., Wong, T.-K.~L., Chen, R.~T., and Duvenaud, D.~K.
\newblock Scalable gradients and variational inference for stochastic
  differential equations.
\newblock In \emph{Symposium on Advances in Approximate Bayesian Inference},
  pp.\  1--28. PMLR, 2020.

\bibitem[Li \& Mandt(2018)Li and Mandt]{li2018disentangle}
Li, Y. and Mandt, S.
\newblock Disentangled sequential autoencoder.
\newblock In \emph{International Conference on Machine Learning}, 2018.

\bibitem[Louis et~al.(2019)Louis, Couronn{\'e}, Koval, Charlier, and
  Durrleman]{louis2019riemannian}
Louis, M., Couronn{\'e}, R., Koval, I., Charlier, B., and Durrleman, S.
\newblock Riemannian geometry learning for disease progression modelling.
\newblock In \emph{International Conference on Information Processing in
  Medical Imaging}, pp.\  542--553. Springer, 2019.

\bibitem[Luo et~al.(2018)Luo, Cai, Zhang, Xu, et~al.]{luo2018multivariate}
Luo, Y., Cai, X., Zhang, Y., Xu, J., et~al.
\newblock Multivariate time series imputation with generative adversarial
  networks.
\newblock \emph{Advances in neural information processing systems}, 31, 2018.

\bibitem[Massaroli et~al.(2020)Massaroli, Poli, Park, Yamashita, and
  Asama]{massaroli2020dissecting}
Massaroli, S., Poli, M., Park, J., Yamashita, A., and Asama, H.
\newblock Dissecting neural odes.
\newblock \emph{Advances in Neural Information Processing Systems},
  33:\penalty0 3952--3963, 2020.

\bibitem[Mathieu et~al.(2019)Mathieu, Le~Lan, Maddison, Tomioka, and
  Teh]{mathieu_continuous_2019}
Mathieu, E., Le~Lan, C., Maddison, C.~J., Tomioka, R., and Teh, Y.~W.
\newblock Continuous hierarchical representations with poincaré variational
  auto-encoders.
\newblock In \emph{Advances in neural information processing systems}, pp.\
  12565--12576, 2019.

\bibitem[Nalisnick et~al.(2016)Nalisnick, Hertel, and
  Smyth]{nalisnick_approximate_2016}
Nalisnick, E., Hertel, L., and Smyth, P.
\newblock Approximate inference for deep latent gaussian mixtures.
\newblock In \emph{NIPS Workshop on Bayesian Deep Learning}, volume~2, pp.\
  131, 2016.

\bibitem[Pang et~al.(2020)Pang, Han, Nijkamp, Zhu, and Wu]{pang_learning_2020}
Pang, B., Han, T., Nijkamp, E., Zhu, S.-C., and Wu, Y.~N.
\newblock Learning latent space energy-based prior model.
\newblock \emph{Advances in Neural Information Processing Systems}, 33, 2020.

\bibitem[Papamakarios et~al.(2017)Papamakarios, Pavlakou, and
  Murray]{papamakarios2017masked}
Papamakarios, G., Pavlakou, T., and Murray, I.
\newblock Masked autoregressive flow for density estimation.
\newblock \emph{Advances in neural information processing systems}, 30, 2017.

\bibitem[Park et~al.(2021)Park, Kim, Lee, Choo, Lee, Kim, and
  Choi]{park2021vid}
Park, S., Kim, K., Lee, J., Choo, J., Lee, J., Kim, S., and Choi, E.
\newblock Vid-ode: Continuous-time video generation with neural ordinary
  differential equation.
\newblock In \emph{Proceedings of the AAAI Conference on Artificial
  Intelligence}, volume~35, pp.\  2412--2422, 2021.

\bibitem[Paszke et~al.(2017)Paszke, Gross, Chintala, Chanan, Yang, {DeVito},
  Lin, Desmaison, Antiga, and Lerer]{paszke_automatic_2017}
Paszke, A., Gross, S., Chintala, S., Chanan, G., Yang, E., {DeVito}, Z., Lin,
  Z., Desmaison, A., Antiga, L., and Lerer, A.
\newblock Automatic differentiation in pytorch.
\newblock 2017.

\bibitem[Pearlmutter(1989)]{pearlmutter1989learning}
Pearlmutter, B.~A.
\newblock Learning state space trajectories in recurrent neural networks.
\newblock \emph{Neural Computation}, 1\penalty0 (2):\penalty0 263--269, 1989.

\bibitem[Ramchandran et~al.(2021)Ramchandran, Tikhonov, Kujanp{\"a}{\"a},
  Koskinen, and L{\"a}hdesm{\"a}ki]{ramchandran_longitudinal_2020}
Ramchandran, S., Tikhonov, G., Kujanp{\"a}{\"a}, K., Koskinen, M., and
  L{\"a}hdesm{\"a}ki, H.
\newblock Longitudinal variational autoencoder.
\newblock In \emph{International Conference on Artificial Intelligence and
  Statistics}, pp.\  3898--3906. PMLR, 2021.

\bibitem[Rangapuram et~al.(2018)Rangapuram, Seeger, Gasthaus, Stella, Wang, and
  Januschowski]{rangapuram2018deep}
Rangapuram, S.~S., Seeger, M.~W., Gasthaus, J., Stella, L., Wang, Y., and
  Januschowski, T.
\newblock Deep state space models for time series forecasting.
\newblock \emph{Advances in neural information processing systems}, 31, 2018.

\bibitem[Razavi et~al.(2020)Razavi, Oord, and Vinyals]{razavi_generating_2019}
Razavi, A., Oord, A. v.~d., and Vinyals, O.
\newblock Generating diverse high-fidelity images with vq-vae-2.
\newblock \emph{Advances in Neural Information Processing Systems}, 2020.

\bibitem[Rezende \& Mohamed(2015)Rezende and Mohamed]{rezende_variational_2015}
Rezende, D. and Mohamed, S.
\newblock Variational inference with normalizing flows.
\newblock In \emph{International Conference on Machine Learning}, pp.\
  1530--1538. PMLR, 2015.

\bibitem[Rezende et~al.(2014)Rezende, Mohamed, and
  Wierstra]{rezende_stochastic_2014}
Rezende, D.~J., Mohamed, S., and Wierstra, D.
\newblock Stochastic backpropagation and approximate inference in deep
  generative models.
\newblock In \emph{International conference on machine learning}, pp.\
  1278--1286. PMLR, 2014.

\bibitem[Roberts et~al.(2013)Roberts, Osborne, Ebden, Reece, Gibson, and
  Aigrain]{roberts2013gaussian}
Roberts, S., Osborne, M., Ebden, M., Reece, S., Gibson, N., and Aigrain, S.
\newblock Gaussian processes for time-series modelling.
\newblock \emph{Philosophical Transactions of the Royal Society A:
  Mathematical, Physical and Engineering Sciences}, 371\penalty0
  (1984):\penalty0 20110550, 2013.

\bibitem[Rubanova et~al.(2019)Rubanova, Chen, and Duvenaud]{rubanova2019latent}
Rubanova, Y., Chen, R.~T., and Duvenaud, D.~K.
\newblock Latent ordinary differential equations for irregularly-sampled time
  series.
\newblock \emph{Advances in neural information processing systems}, 32, 2019.

\bibitem[Salimans et~al.(2015)Salimans, Kingma, and
  Welling]{salimans_markov_2015}
Salimans, T., Kingma, D., and Welling, M.
\newblock Markov chain monte carlo and variational inference: Bridging the gap.
\newblock In \emph{International Conference on Machine Learning}, pp.\
  1218--1226, 2015.

\bibitem[Sauty \& Durrleman(2022)Sauty and Durrleman]{sauty2022progression}
Sauty, B. and Durrleman, S.
\newblock Progression models for imaging data with longitudinal variational
  auto encoders.
\newblock In \emph{International Conference on Medical Image Computing and
  Computer-Assisted Intervention}, pp.\  3--13. Springer, 2022.

\bibitem[Schiratti et~al.(2015)Schiratti, Allassonniere, Colliot, and
  Durrleman]{schiratti2015learning}
Schiratti, J.-B., Allassonniere, S., Colliot, O., and Durrleman, S.
\newblock Learning spatiotemporal trajectories from manifold-valued
  longitudinal data.
\newblock \emph{Advances in neural information processing systems}, 28, 2015.

\bibitem[Seeger(2004)]{seeger2004gaussian}
Seeger, M.
\newblock Gaussian processes for machine learning.
\newblock \emph{International journal of neural systems}, 14\penalty0
  (02):\penalty0 69--106, 2004.

\bibitem[Serban et~al.(2017)Serban, Sordoni, Lowe, Charlin, Pineau, Courville,
  and Bengio]{serban2017hierarchical}
Serban, I., Sordoni, A., Lowe, R., Charlin, L., Pineau, J., Courville, A., and
  Bengio, Y.
\newblock A hierarchical latent variable encoder-decoder model for generating
  dialogues.
\newblock In \emph{Proceedings of the AAAI Conference on Artificial
  Intelligence}, volume~31, 2017.

\bibitem[Shao et~al.(2018)Shao, Kumar, and Fletcher]{shao_riemannian_2018}
Shao, H., Kumar, A., and Fletcher, P.~T.
\newblock The riemannian geometry of deep generative models.
\newblock In \emph{2018 {IEEE}/{CVF} Conference on Computer Vision and Pattern
  Recognition Workshops ({CVPRW})}, pp.\  428--4288. {IEEE}, 2018.
\newblock ISBN 978-1-5386-6100-0.
\newblock \doi{10.1109/CVPRW.2018.00071}.

\bibitem[Singer et~al.(2003)Singer, Willett, Willett,
  et~al.]{singer2003applied}
Singer, J.~D., Willett, J.~B., Willett, J.~B., et~al.
\newblock \emph{Applied longitudinal data analysis: Modeling change and event
  occurrence}.
\newblock Oxford university press, 2003.

\bibitem[Singh et~al.(2016)Singh, Hinkle, Joshi, and
  Fletcher]{singh2016hierarchical}
Singh, N., Hinkle, J., Joshi, S., and Fletcher, P.~T.
\newblock Hierarchical geodesic models in diffeomorphisms.
\newblock \emph{International Journal of Computer Vision}, 117\penalty0
  (1):\penalty0 70--92, 2016.

\bibitem[Sohn et~al.(2015)Sohn, Lee, and Yan]{sohn2015learning}
Sohn, K., Lee, H., and Yan, X.
\newblock Learning structured output representation using deep conditional
  generative models.
\newblock \emph{Advances in neural information processing systems}, 28, 2015.

\bibitem[S{\o}nderby et~al.(2016)S{\o}nderby, Raiko, Maal{\o}e, S{\o}nderby,
  and Winther]{sonderby_ladder_2016}
S{\o}nderby, C.~K., Raiko, T., Maal{\o}e, L., S{\o}nderby, S.~K., and Winther,
  O.
\newblock Ladder variational autoencoder.
\newblock In \emph{29th Annual Conference on Neural Information Processing
  Systems (NIPS 2016)}, 2016.

\bibitem[Tomczak \& Welling(2018)Tomczak and Welling]{tomczak_vae_2018}
Tomczak, J. and Welling, M.
\newblock Vae with a vampprior.
\newblock In \emph{International Conference on Artificial Intelligence and
  Statistics}, pp.\  1214--1223. PMLR, 2018.

\bibitem[Tzen \& Raginsky(2019)Tzen and Raginsky]{tzen2019neural}
Tzen, B. and Raginsky, M.
\newblock Neural stochastic differential equations: Deep latent gaussian models
  in the diffusion limit.
\newblock \emph{arXiv preprint arXiv:1905.09883}, 2019.

\bibitem[Xu et~al.(2021)Xu, Fu, Liu, Wang, Li, Huang, Zhang, and
  Xue]{xu2021learning}
Xu, C., Fu, Y., Liu, C., Wang, C., Li, J., Huang, F., Zhang, L., and Xue, X.
\newblock Learning dynamic alignment via meta-filter for few-shot learning.
\newblock In \emph{Proceedings of the IEEE/CVF conference on computer vision
  and pattern recognition}, pp.\  5182--5191, 2021.

\bibitem[Yildiz et~al.(2019)Yildiz, Heinonen, and
  Lahdesmaki]{yildiz2019ode2vae}
Yildiz, C., Heinonen, M., and Lahdesmaki, H.
\newblock Ode2vae: Deep generative second order odes with bayesian neural
  networks.
\newblock \emph{Advances in Neural Information Processing Systems}, 32, 2019.

\bibitem[Zhao et~al.(2021)Zhao, Liu, Adeli, and Pohl]{zhao2021longitudinal}
Zhao, Q., Liu, Z., Adeli, E., and Pohl, K.~M.
\newblock Longitudinal self-supervised learning.
\newblock \emph{Medical Image Analysis}, 71:\penalty0 102051, 2021.

\end{thebibliography}
\bibliographystyle{icml2023}

%%%%%%%%%%%%%%%%%%%%%%%%%%%%%%%%%%%%%%%%%%%%%%%%%%%%%%%%%%%%%%%%%%%%%%%%%%%%%%%
%%%%%%%%%%%%%%%%%%%%%%%%%%%%%%%%%%%%%%%%%%%%%%%%%%%%%%%%%%%%%%%%%%%%%%%%%%%%%%%
% APPENDIX
%%%%%%%%%%%%%%%%%%%%%%%%%%%%%%%%%%%%%%%%%%%%%%%%%%%%%%%%%%%%%%%%%%%%%%%%%%%%%%%
%%%%%%%%%%%%%%%%%%%%%%%%%%%%%%%%%%%%%%%%%%%%%%%%%%%%%%%%%%%%%%%%%%%%%%%%%%%%%%%
\newpage
\appendix
\onecolumn

\section{The Data}\label{app: the data}
\begin{figure}[ht]
    \centering
    \begin{minipage}[c]{0.59\linewidth}
        \subfloat[\emph{Starmen} \label{subfig: training starmen}]{
    \includegraphics[width=0.54\linewidth]{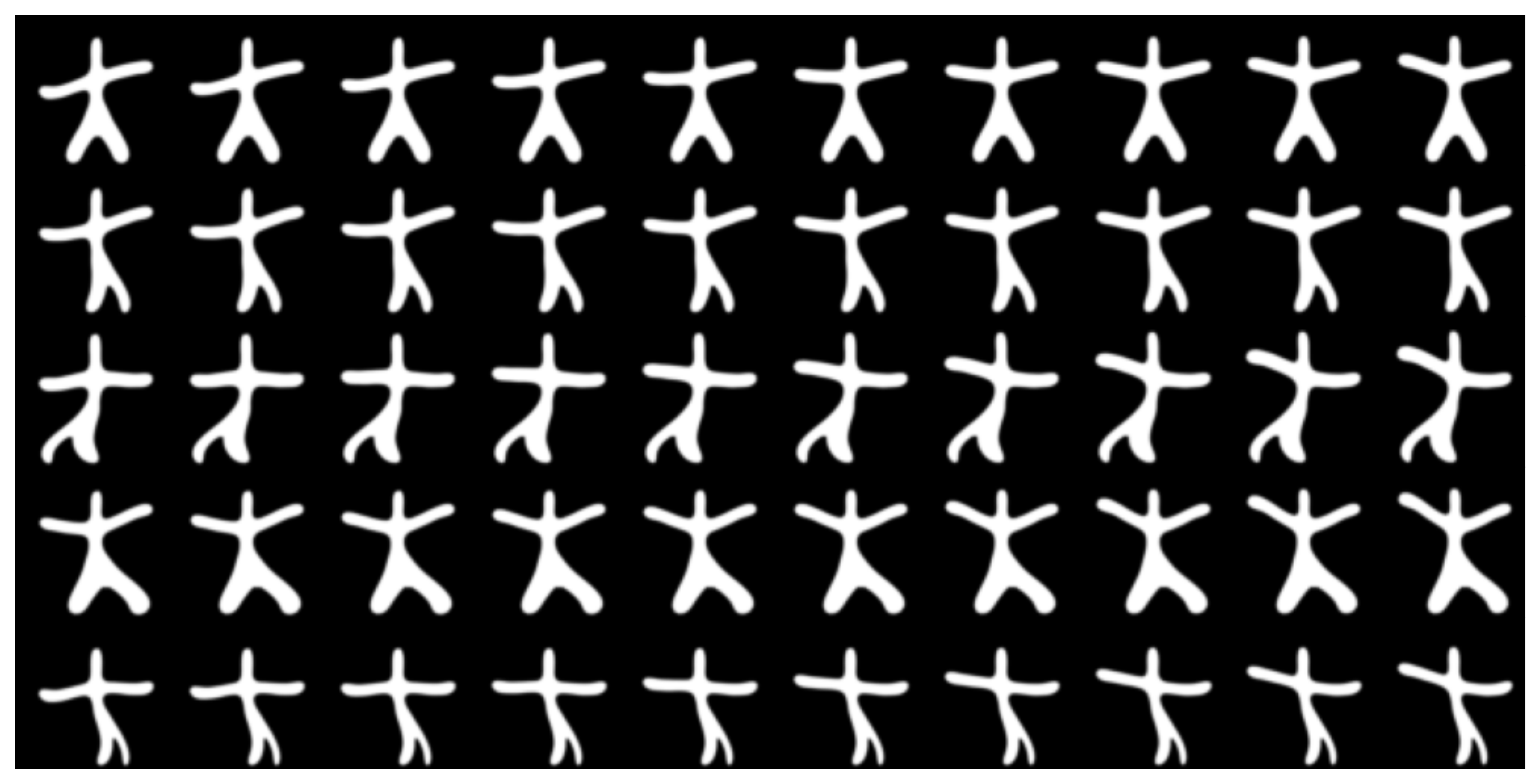}}
    \subfloat[\emph{rotMNIST} \label{subfig: training rotated}]{
    \includegraphics[width=0.435\linewidth]{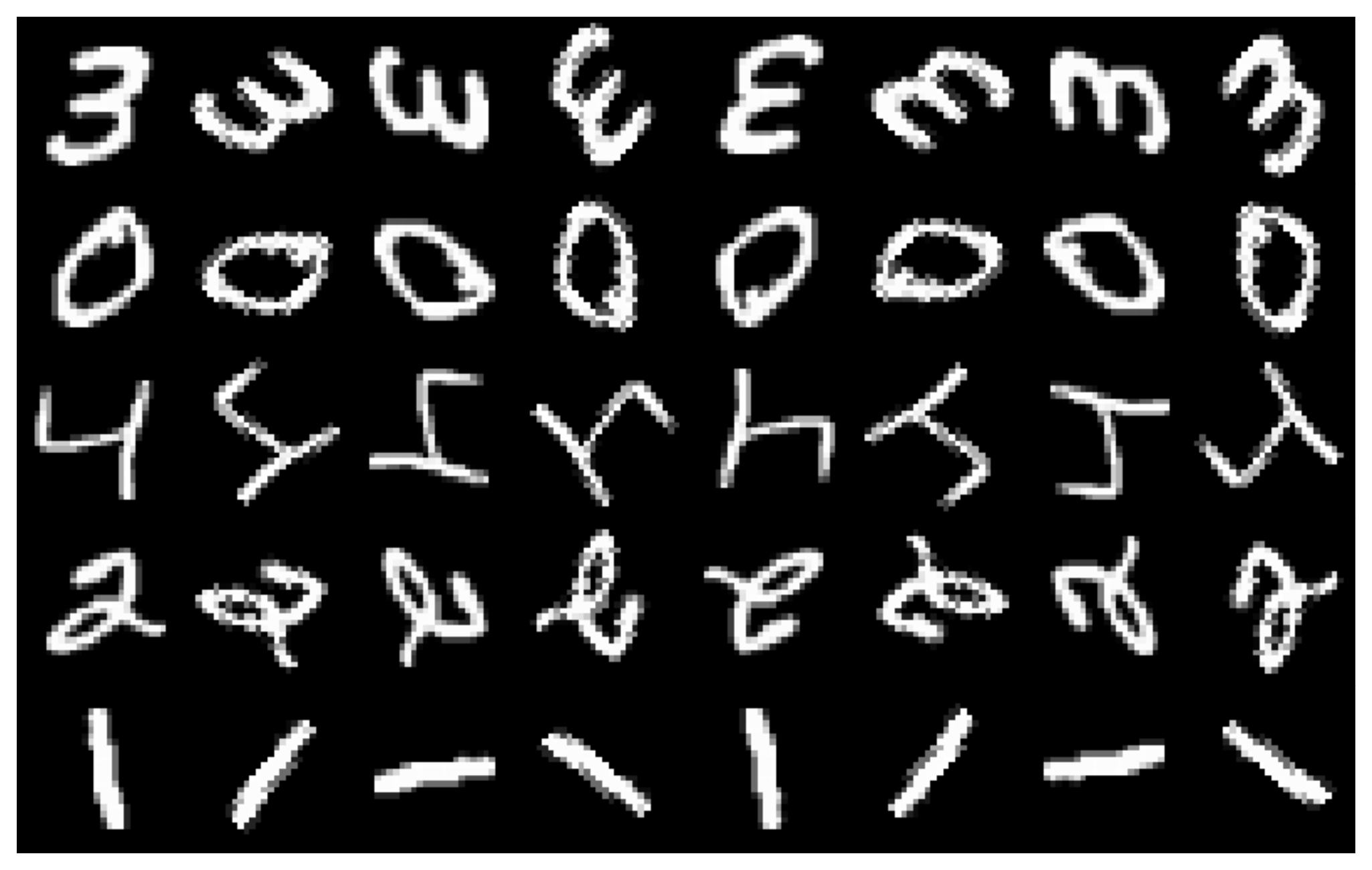}}\\
    \subfloat[\emph{colorMNIST} \label{subfig: training color}]{
    \includegraphics[width=\linewidth]{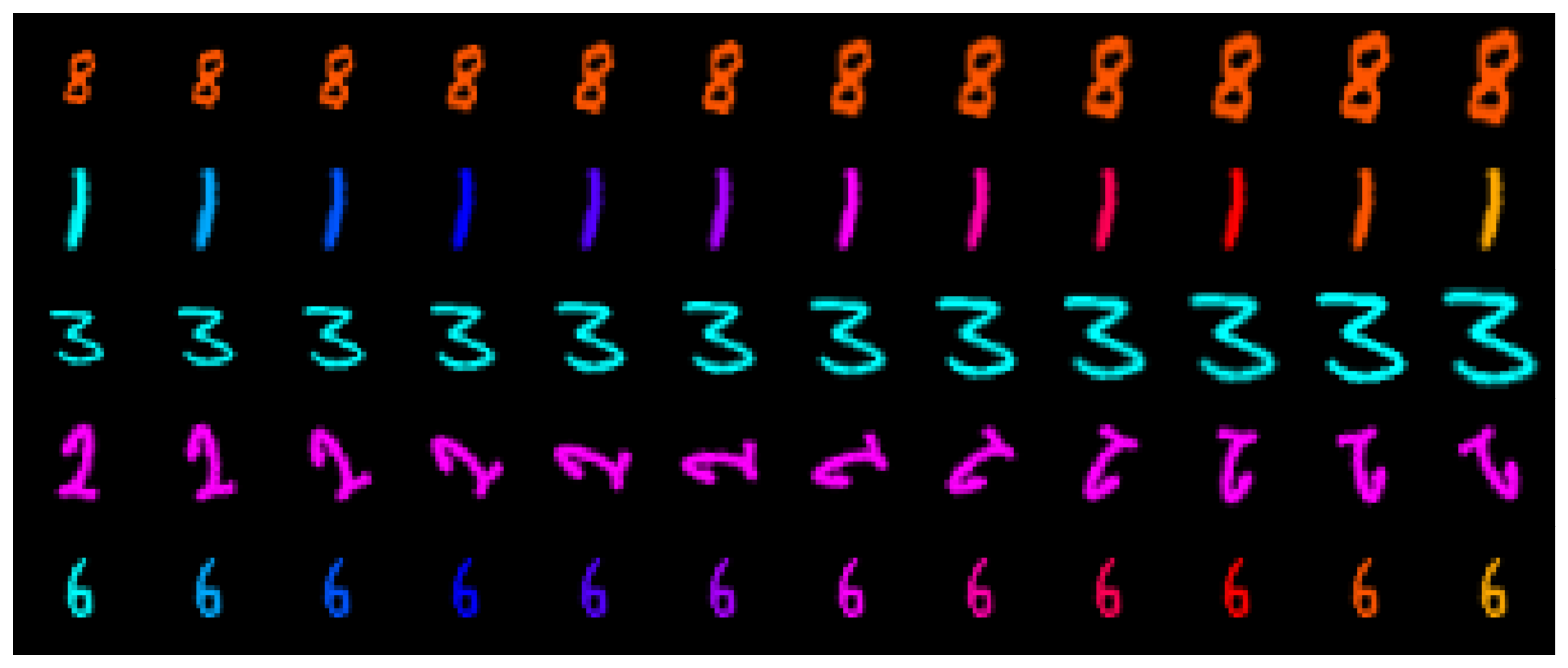}}
    \end{minipage}
    \begin{minipage}[c]{0.39\linewidth}
    \vspace{0.5em}
        \subfloat[\emph{Faces} \label{subfig: training faces}]{
    \includegraphics[width=\linewidth]{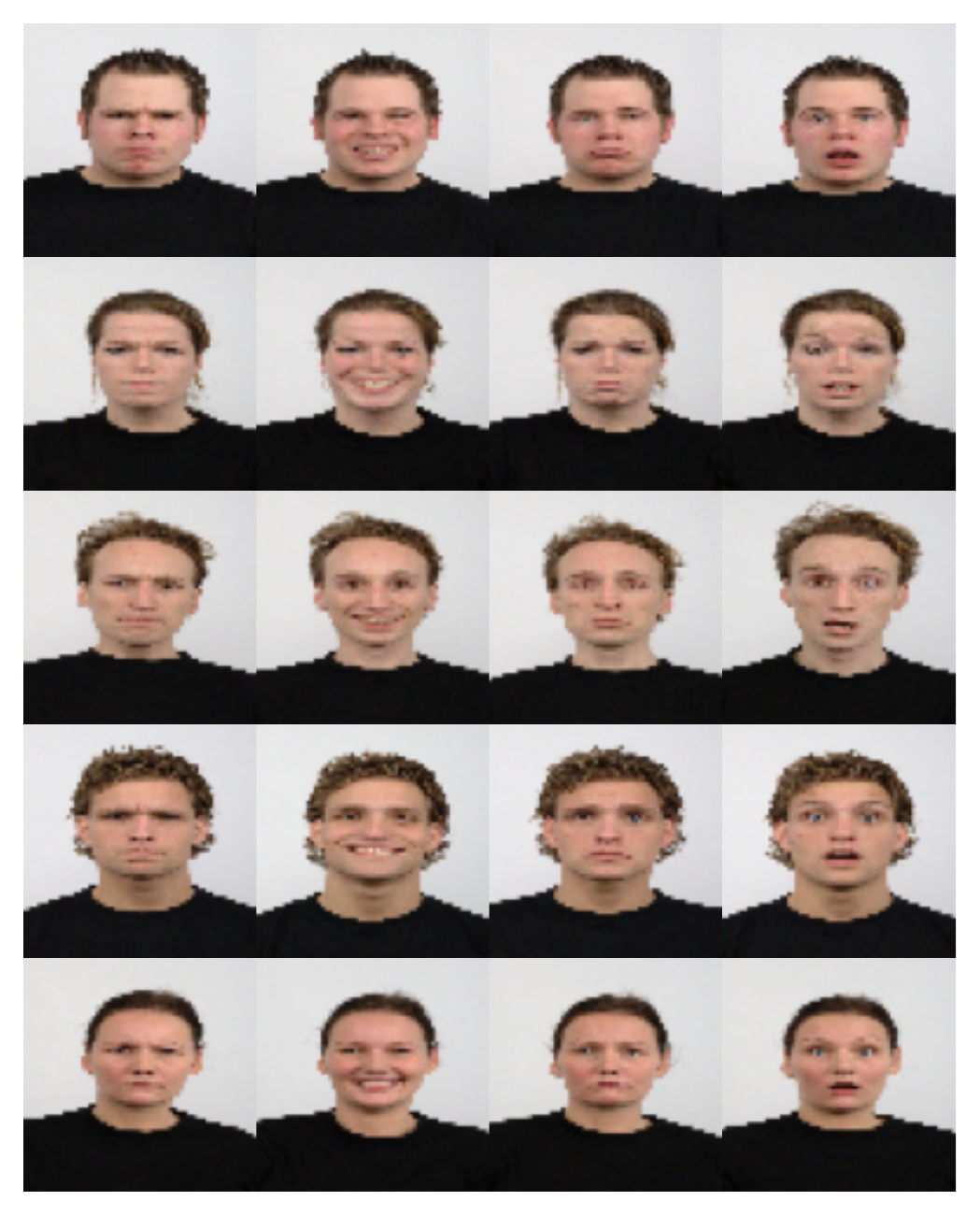}}
    \end{minipage}
    \subfloat[\emph{3d chairs} \label{subfig: training chairs}]{
    \includegraphics[width=0.56\linewidth]{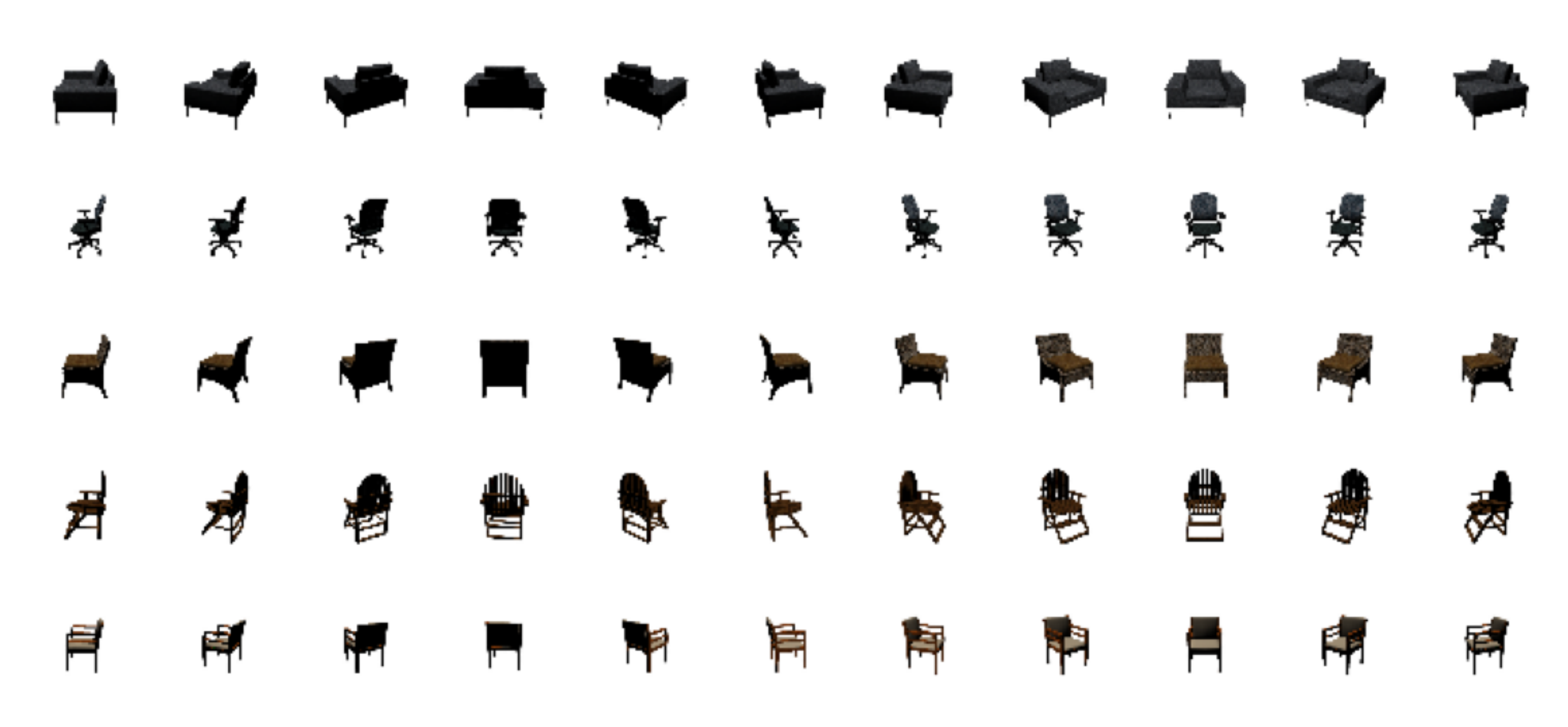}}
    \subfloat[\emph{Sprites} \label{subfig: training sprites}]{
    \includegraphics[width=0.4\linewidth]{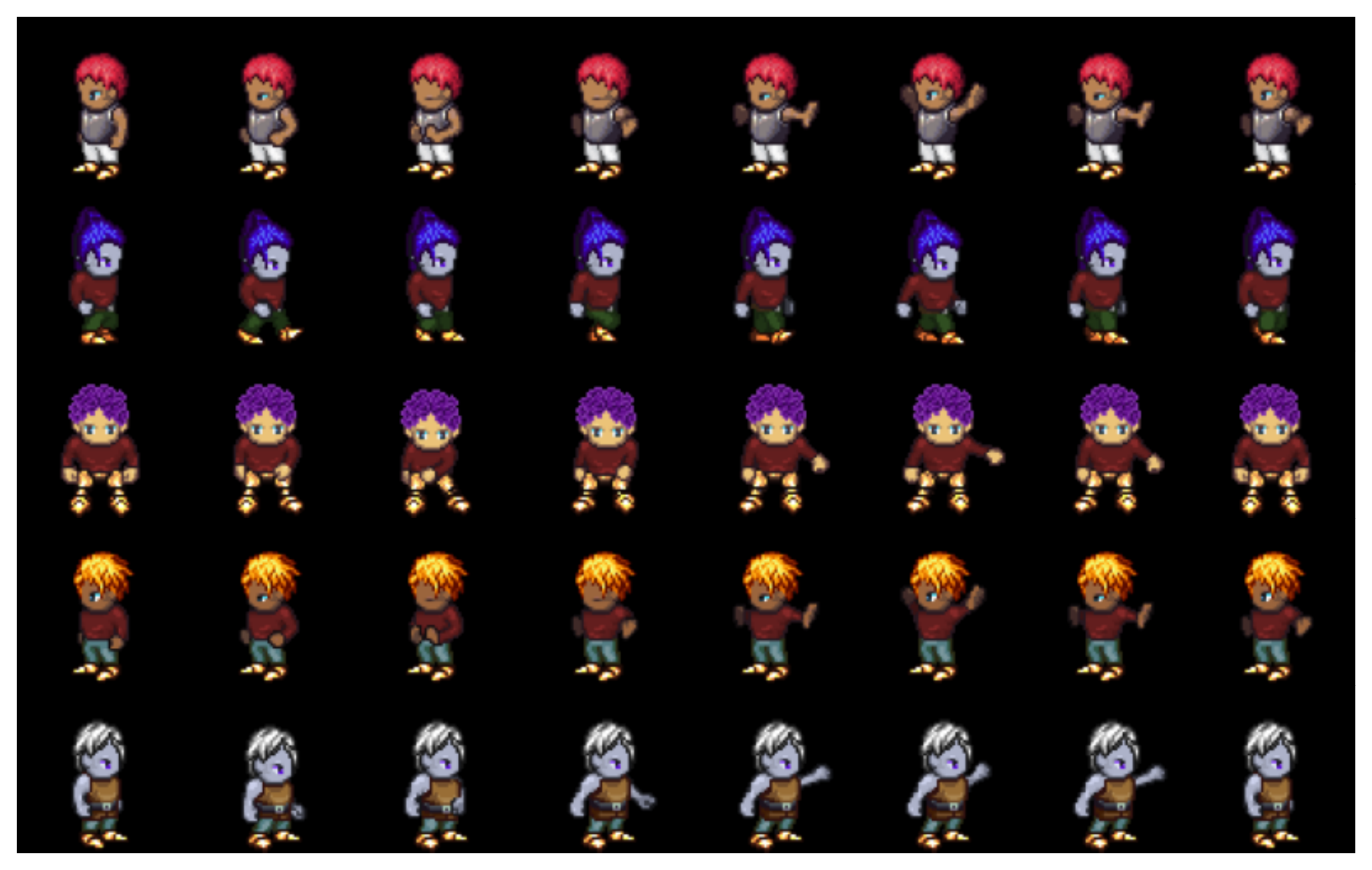}}
    \caption{5 training sequences for each dataset considered in the paper.}
\end{figure}

In this section, we display some training samples for each dataset used in the paper. We recall that the first one shown in Fig.~\ref{subfig: training starmen} is a synthetic longitudinal dataset composed of 1,000, 64x64 images of \textit{starmen} raising their left arm and generated according to the diffeomorphic model of \cite{bone2018learning}. The second one shown in Fig.~\ref{subfig: training rotated} consists of 8 evenly separated rotations applied to the MNIST database \cite{lecun_mnist_1998} from 0 to 360 degrees. The third one called \textit{colorMNIST} is created using the approach of \cite{keller2021topographic}. It consists of sequences of colored MNIST digits that can undergo three distinct types of transformations: color change (from turquoise to yellow), scale change or rotations and is presented in Fig.~\ref{subfig: training color}. It is important to note that for this database, the time dynamic cannot be fully recovered from a single image since it can correspond to different transformations. For instance, a starting turquoise \textit{6} can either change of color or undergo a change in scale as shown on line 2 and 3 of Fig.~\ref{subfig: training color}. The fourth database is created using the \textit{3d chairs} dataset \cite{aubry2014seeing} consisting of 3D CAD chair models and considering as input sequences, 11 evenly separated rotations of a chair (from 0 to 360$^{\circ}$). Some samples are displayed in Fig.~\ref{subfig: training chairs}. We also use the \textit{sprites} dataset \cite{li2018disentangle} shown in Fig.~\ref{subfig: training sprites} that consists of 64x64 RGB images of characters performing actions such as dancing or walking. Finally, we also consider the Radboud Faces Database consisting of 67 individuals expressing different emotions \cite{langner2010presentation}. For this dataset, we create sequences of 4 time steps corresponding to the emotions: \emph{anger}, \emph{happiness}, \emph{sadness} and \textit{surprise}; and down-sample the images so they are of size 64x64 as shown in Fig.~\ref{subfig: training faces}.

\section{Some More Generations}\label{app: more generations}
\begin{figure}[ht]
    \centering
    \includegraphics[scale=0.33]{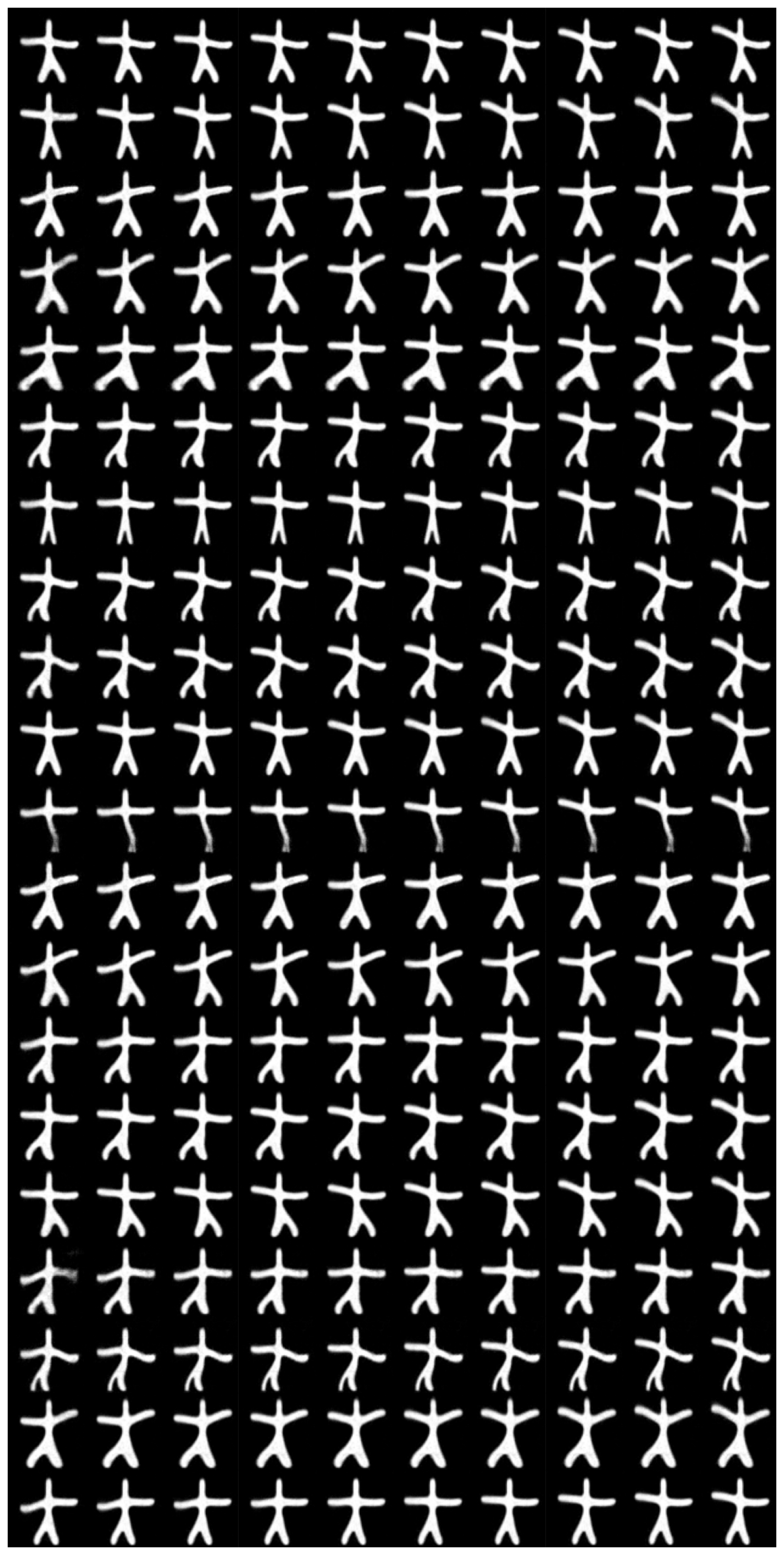}
    \caption{20 sequences generated by our model trained on the \textit{starmen} dataset.}
    \label{fig:appedix gen starmen}
\end{figure}

\begin{figure}
    \centering
    \includegraphics[scale=0.33]{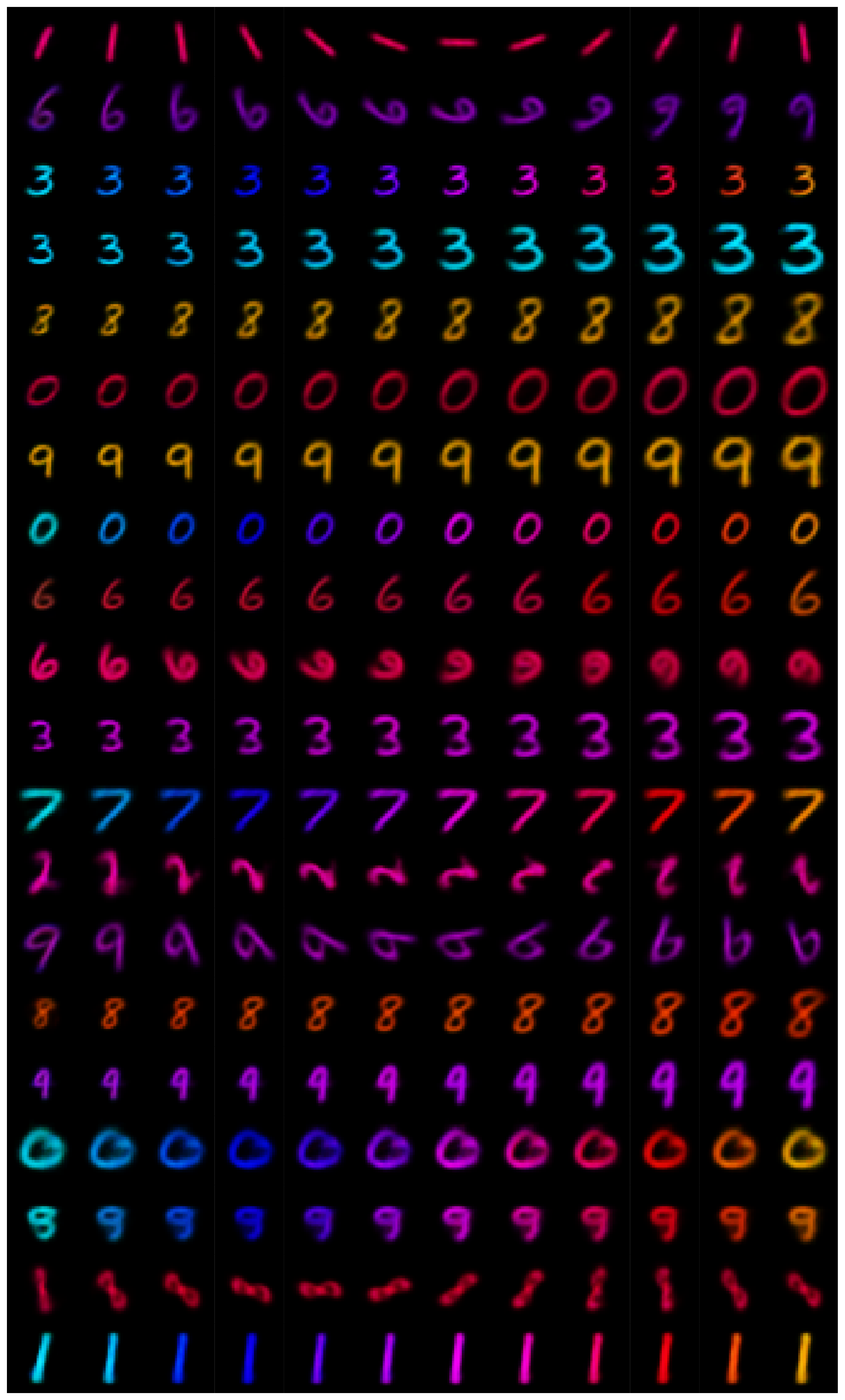}
    \caption{20 sequences generated by our model trained on the \textit{colorMNIST} dataset.}
    \label{fig:appedix gen colormnist}
\end{figure}

\begin{figure}
    \centering
    \includegraphics[scale=0.33]{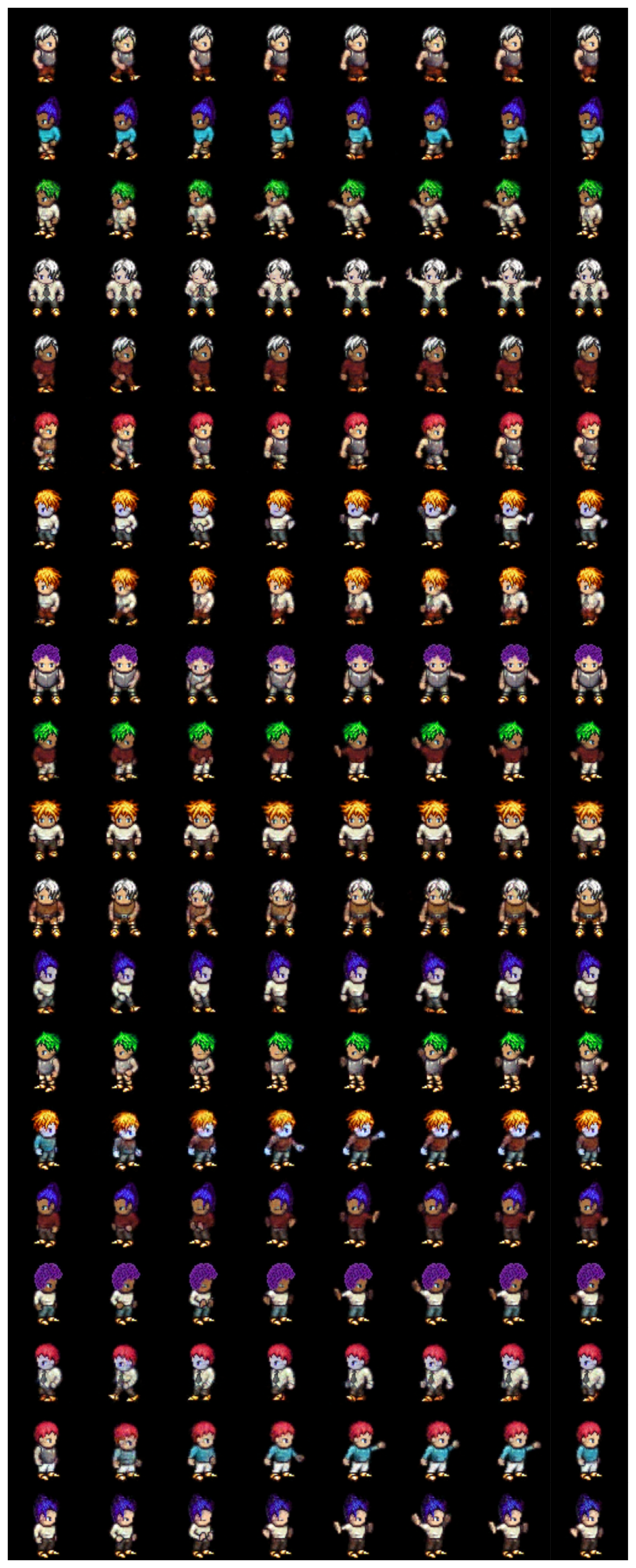}
    \caption{20 sequences generated by our model trained on the \textit{sprites} dataset.}
    \label{fig:appedix gen sprites}
\end{figure}

\begin{figure}
    \centering
    \includegraphics[scale=0.33]{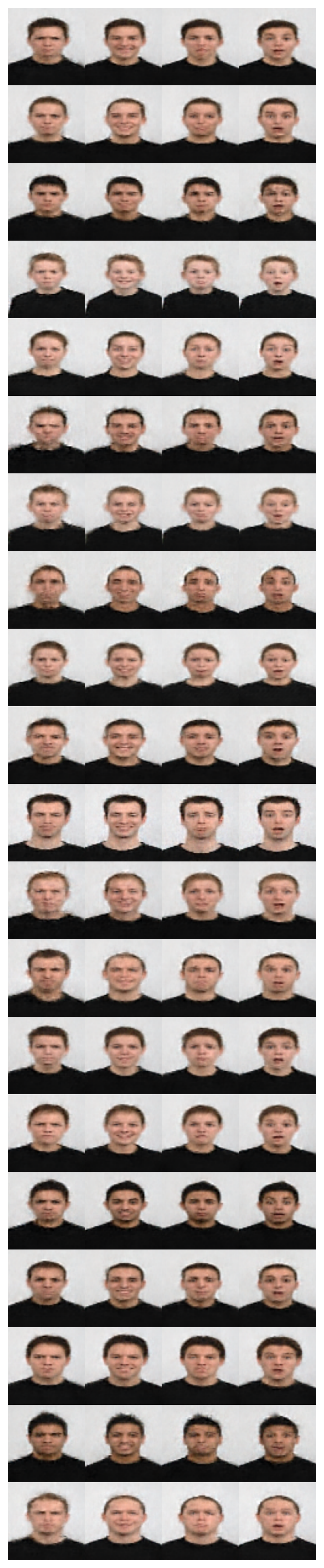}
    \caption{20 sequences generated by our model trained on the \textit{faces} dataset.}
    \label{fig:appedix gen faces}
\end{figure}

\begin{figure}
    \centering
    \includegraphics[scale=0.33]{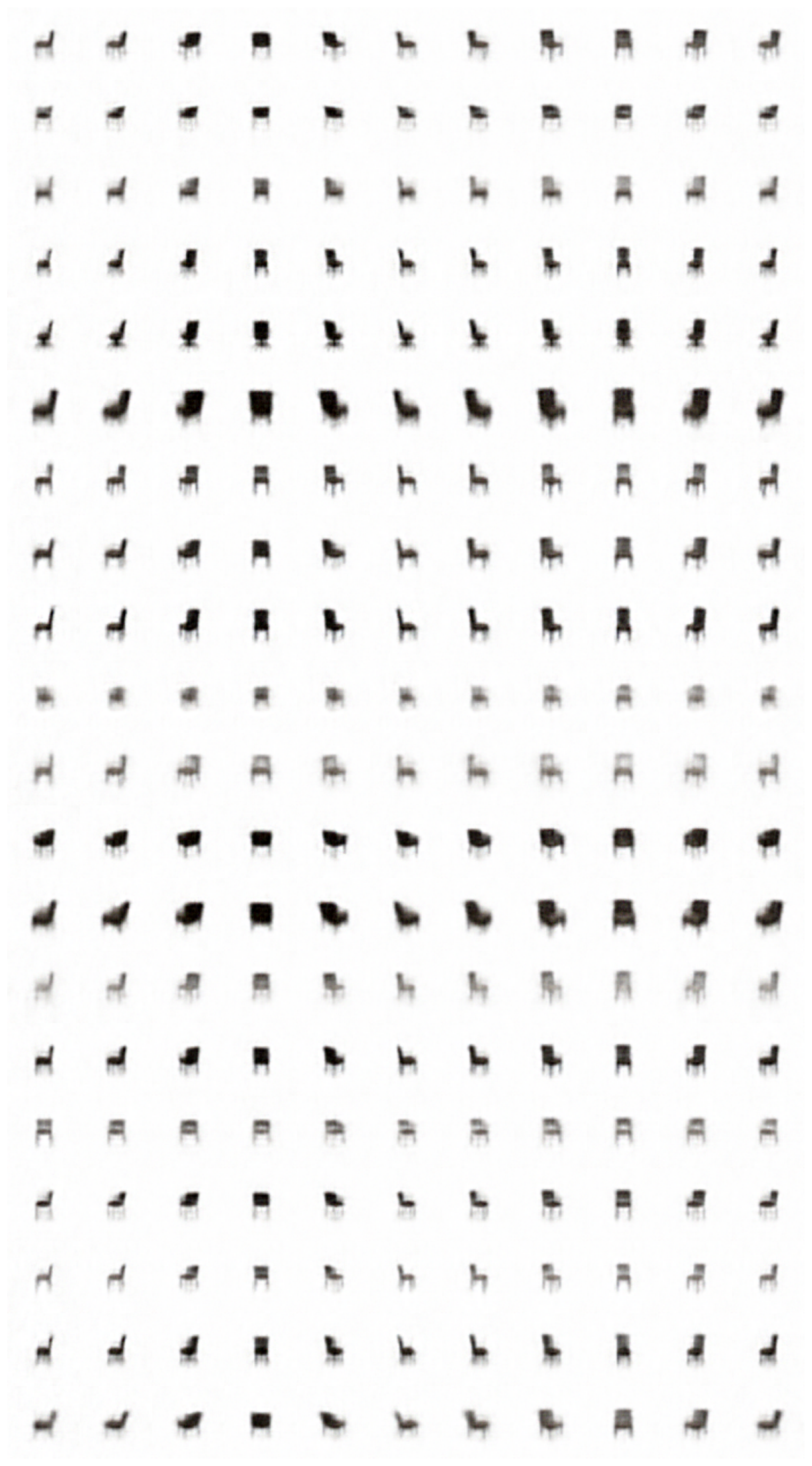}
    \caption{20 sequences generated by our model trained on the \textit{chairs} dataset.}
    \label{fig:appedix gen chairs}
\end{figure}

In this section, we show 20 additional generated sequences for the \emph{starmen} dataset in Fig.~\ref{fig:appedix gen starmen}, the \textit{colorMNIST} dataset in Fig.~\ref{fig:appedix gen colormnist}, the \textit{sprites} data in Fig.~\ref{fig:appedix gen sprites}, the \emph{faces} dataset in Fig.~\ref{fig:appedix gen faces} and the \textit{chairs} dataset in Fig.~\ref{fig:appedix gen chairs}. This experiment shows the diversity of the generated trajectories as well as their relevance.

\clearpage

\section{Exploring Overfitting}\label{app: overfitting}
In this section, we show that the proposed model generates unseen sequences by comparing 4 generated trajectories to the closest one in the train set (using L2 norm). For each dataset, we see that the generated sequence is different from the training data. For instance, for the \textit{starmen}, the individual has a different shape while for the \textit{sprites}, the individual has different pants, hair or top's color.

\begin{figure*}[ht]
    \centering
    \adjustbox{minipage=2.5em,raise=\dimexpr 1.3\height}{\small Gen\vspace{1em}
    Train}
    \includegraphics[width=0.4\linewidth]{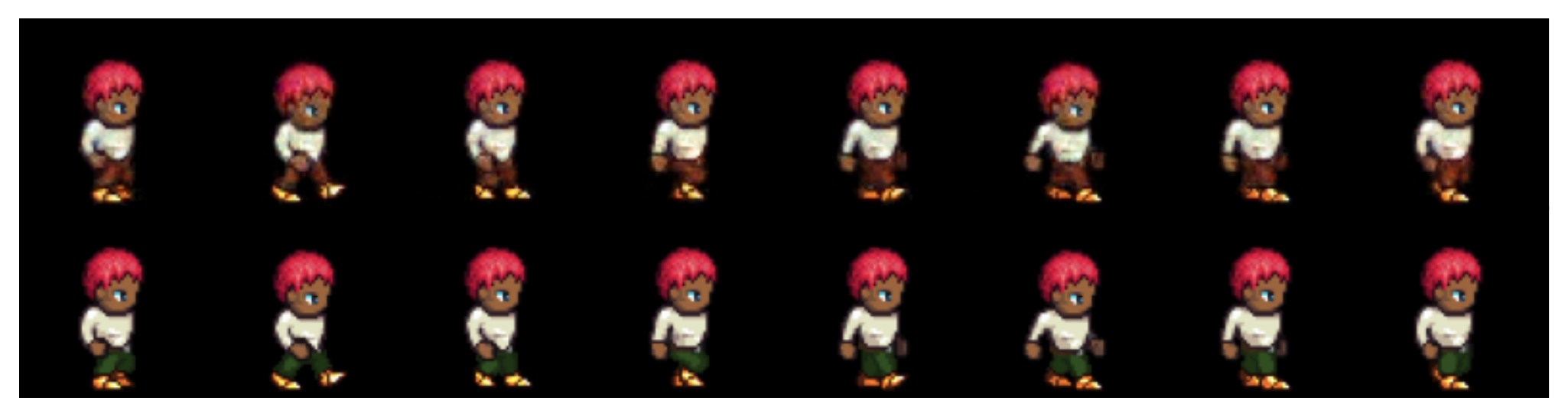}
    \includegraphics[width=0.4\linewidth]{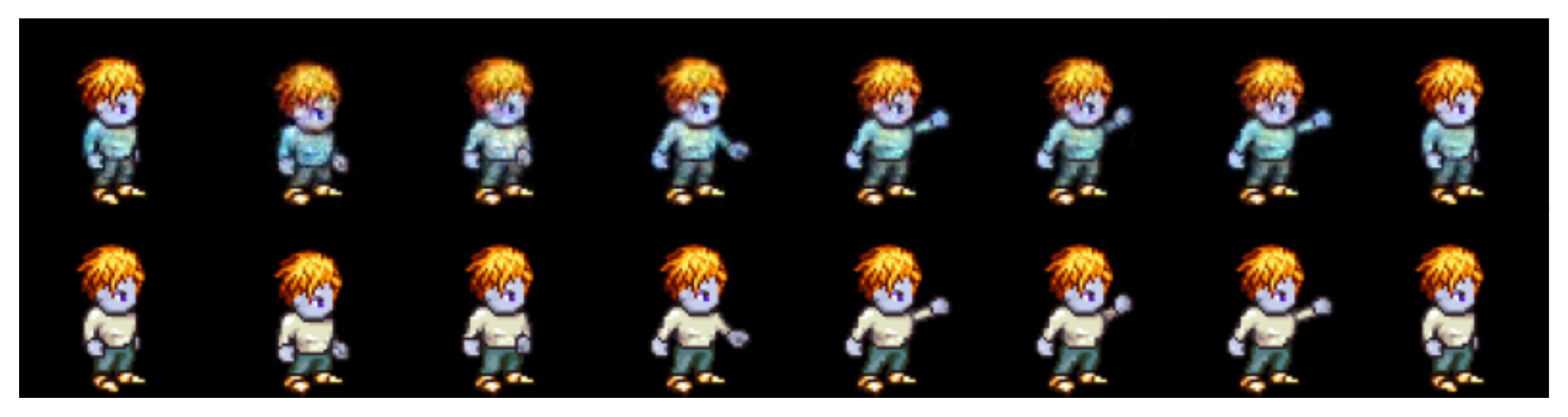}\vspace{-1.2em}\\
    \adjustbox{minipage=2.5em,raise=\dimexpr 1.4\height}{\small Gen.\vspace{1em}
    Train}
    \captionsetup[subfigure]{position=below, labelformat = empty}
    \subfloat[(a)]{\includegraphics[width=0.4\linewidth]{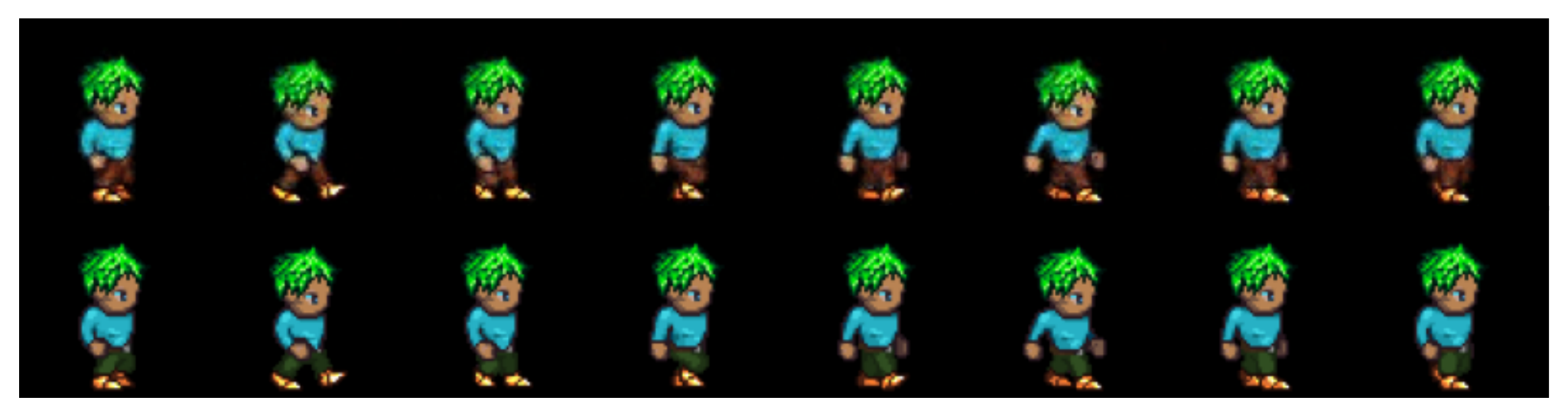}
    \includegraphics[width=0.4\linewidth]{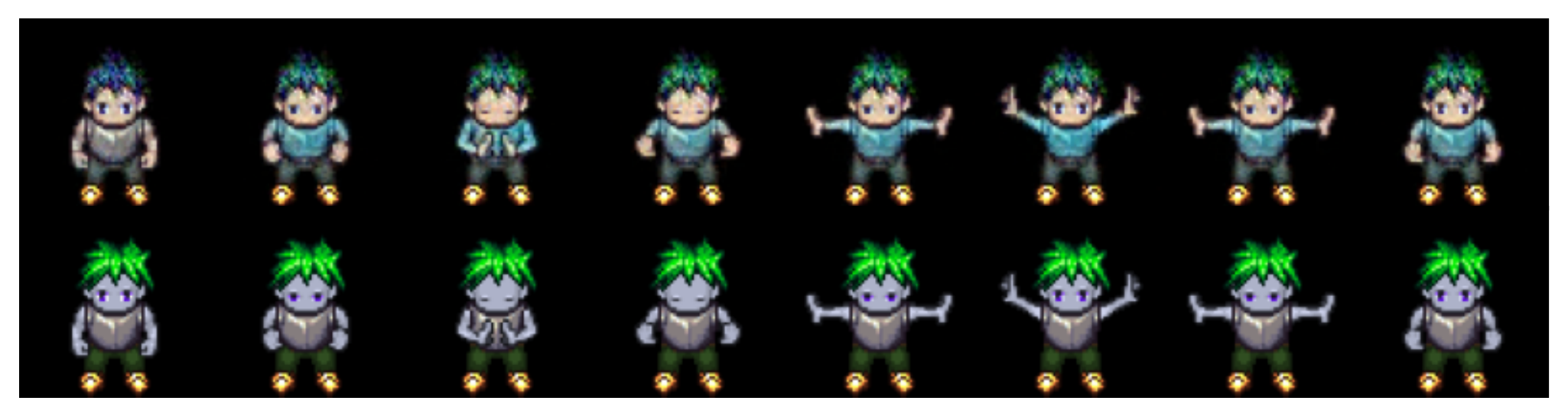}}\\
    \adjustbox{minipage=2.5em,raise=\dimexpr 1.1\height}{\small Gen.\vspace{0.8em}
    Train}
    \includegraphics[width=0.4\linewidth]{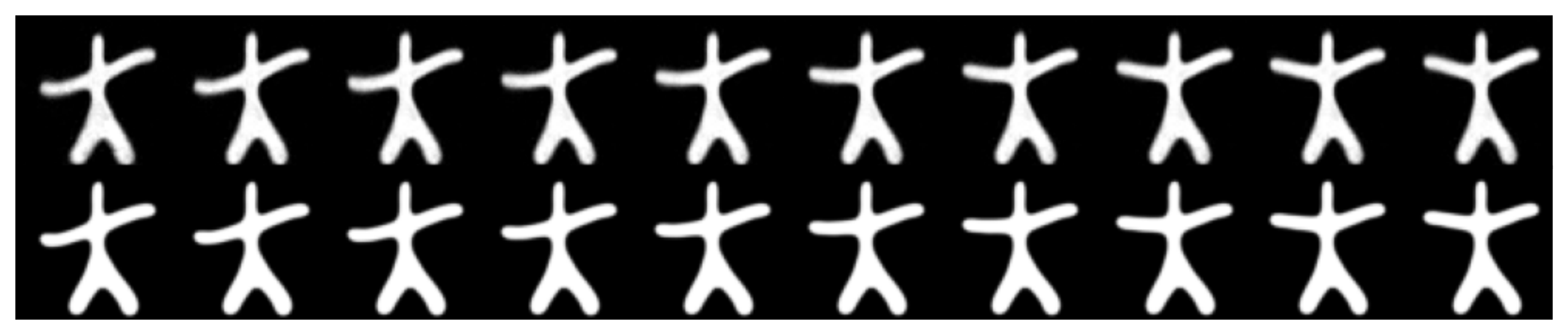}
    \includegraphics[width=0.4\linewidth]{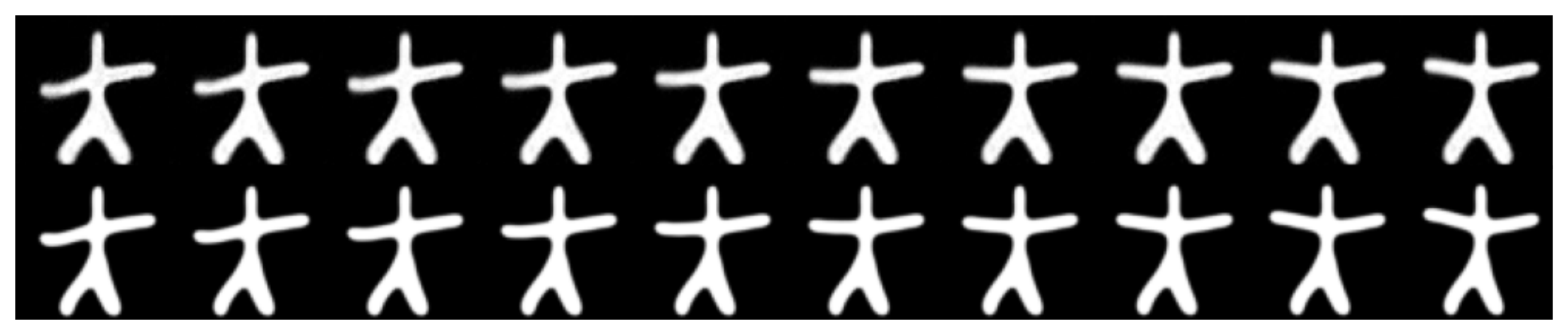}\vspace{-1.2em}\\
    \adjustbox{minipage=2.5em,raise=\dimexpr 1.2\height}{\small Gen.\vspace{0.8em}
    Train}
    \subfloat[(b)]{\includegraphics[width=0.4\linewidth]{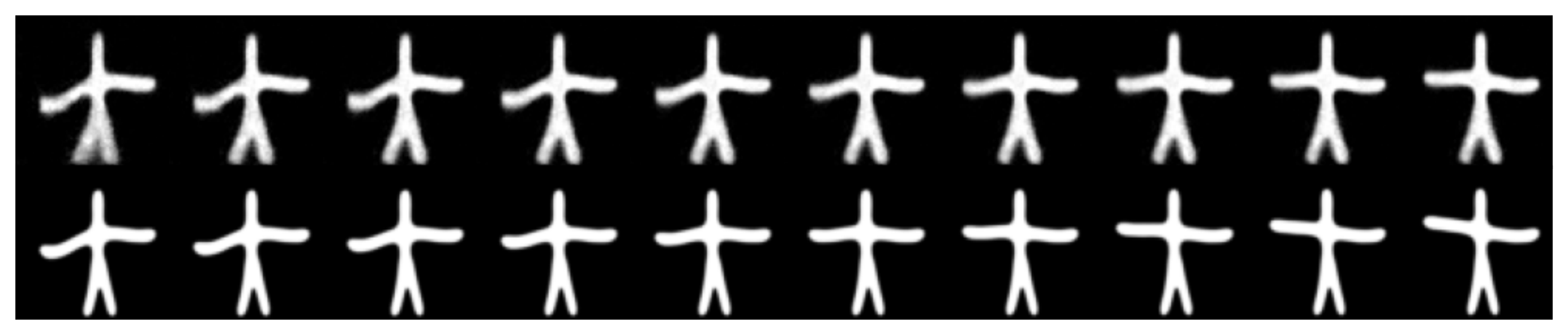}
    \includegraphics[width=0.4\linewidth]{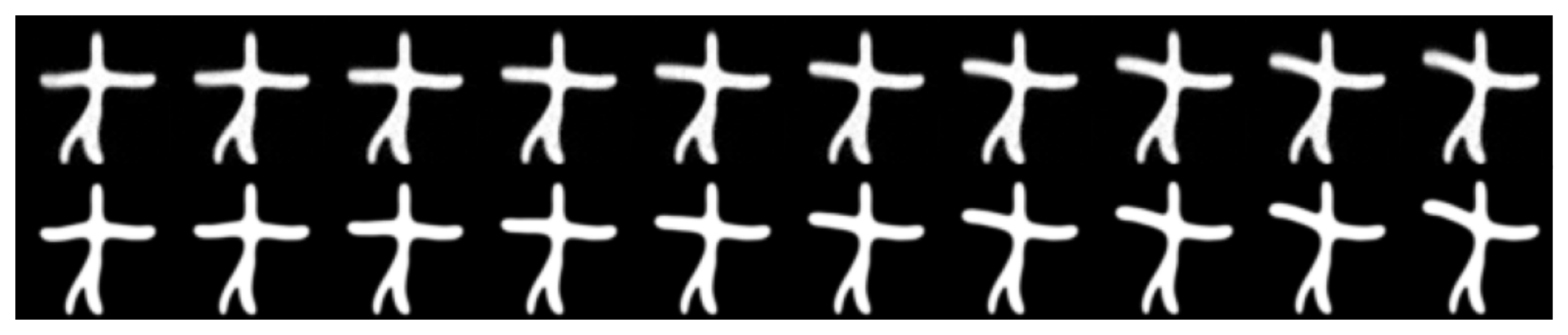}}\\
    \adjustbox{minipage=2.5em,raise=\dimexpr 1.2\height}{\small Gen.\vspace{0.4em}
    Train}
    \includegraphics[width=0.4\linewidth]{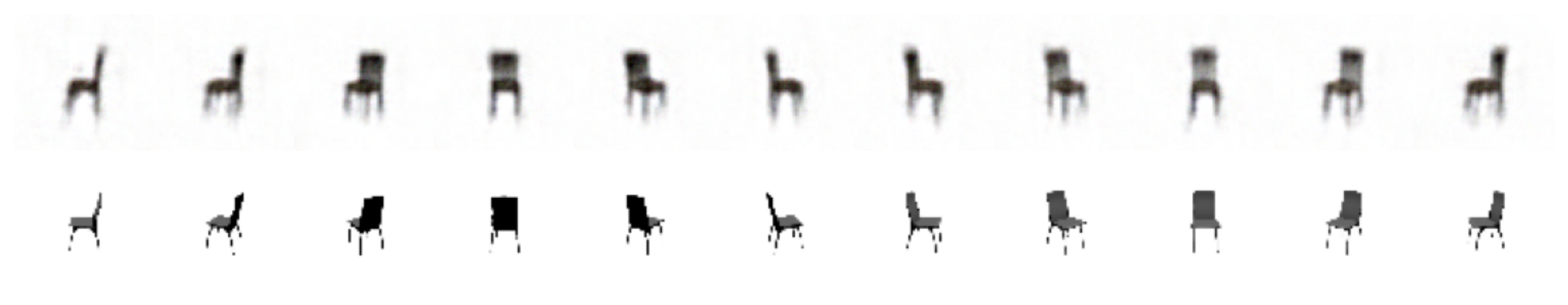}
    \includegraphics[width=0.4\linewidth]{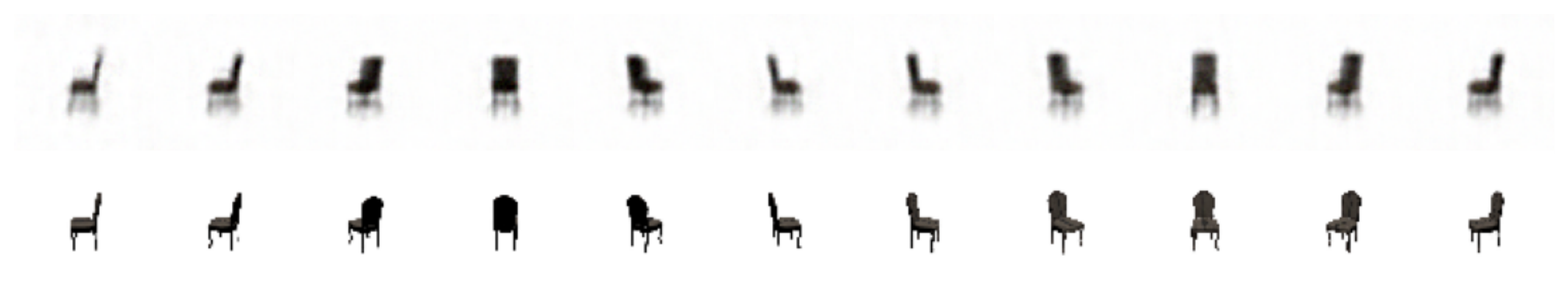}\vspace{-1.2em}\\
    \adjustbox{minipage=2.5em,raise=\dimexpr 1.2\height}{\small Gen.\vspace{0.6em}
    Train}
     \subfloat[(c)]{\includegraphics[width=0.4\linewidth]{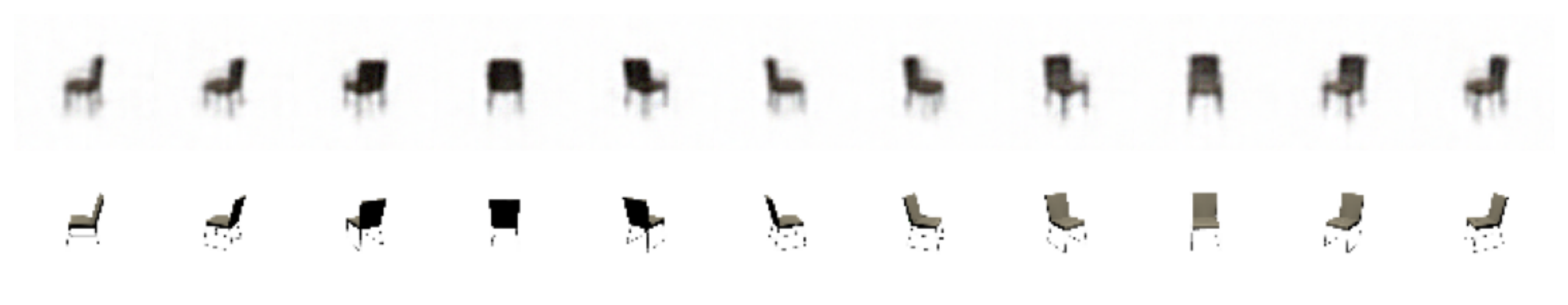}
    \includegraphics[width=0.4\linewidth]{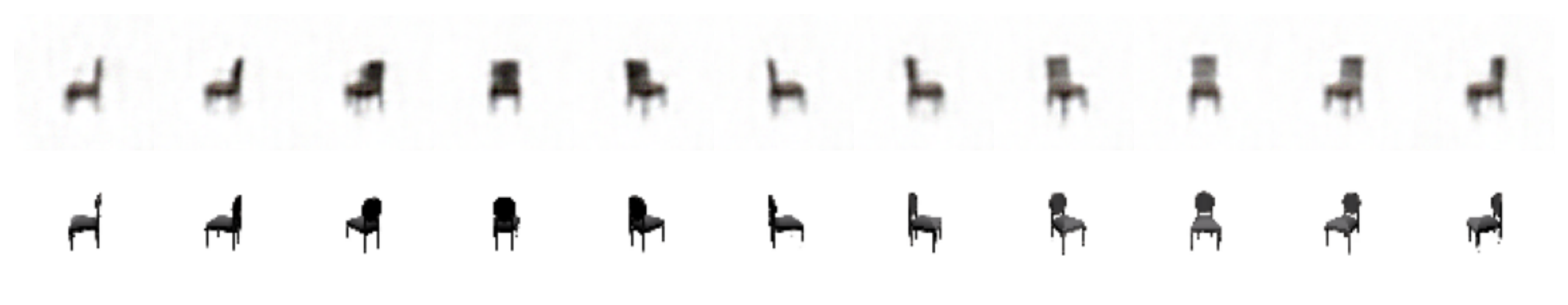}}\\
    \adjustbox{minipage=3em,raise=\dimexpr 1.5\height}{\small Gen.\vspace{1em}
    Train}
     \subfloat[(d)]{\includegraphics[width=0.193\linewidth]{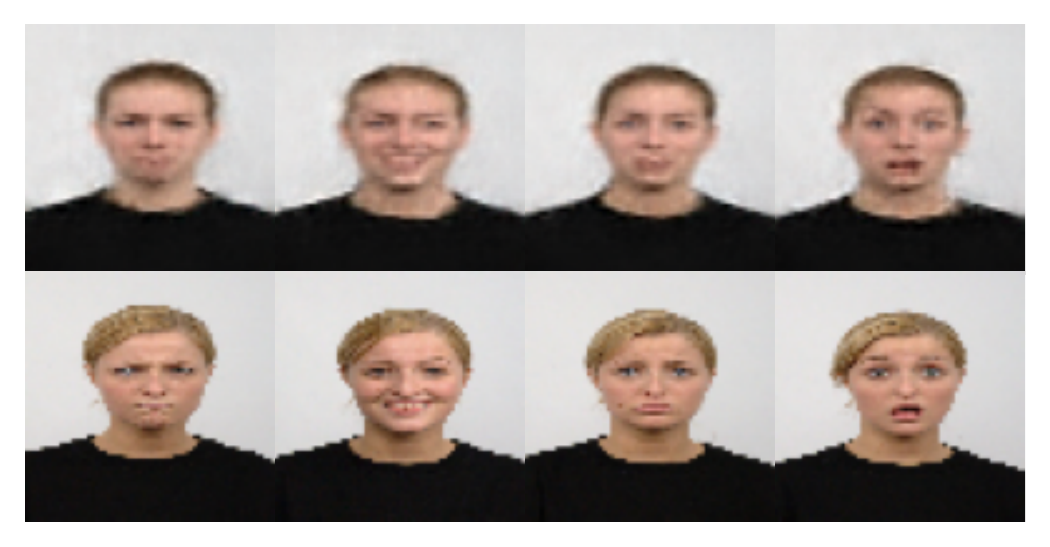}
    \includegraphics[width=0.193\linewidth]{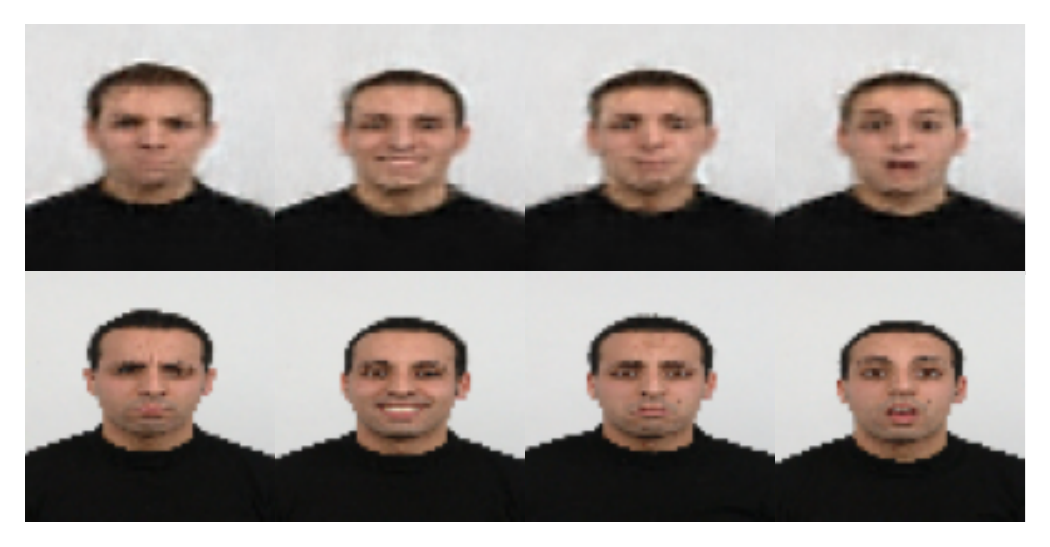}
    \includegraphics[width=0.193\linewidth]{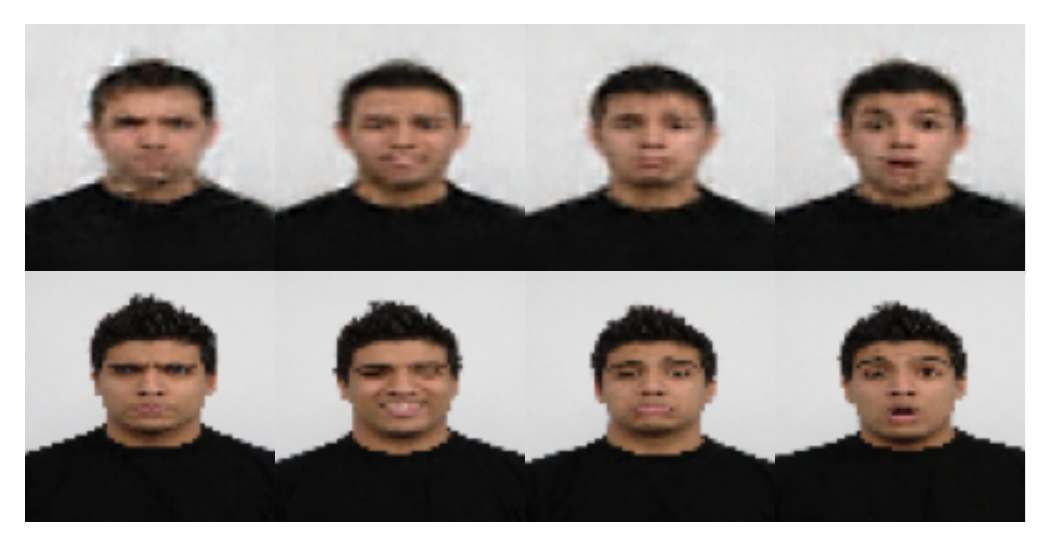}
    \includegraphics[width=0.193\linewidth]{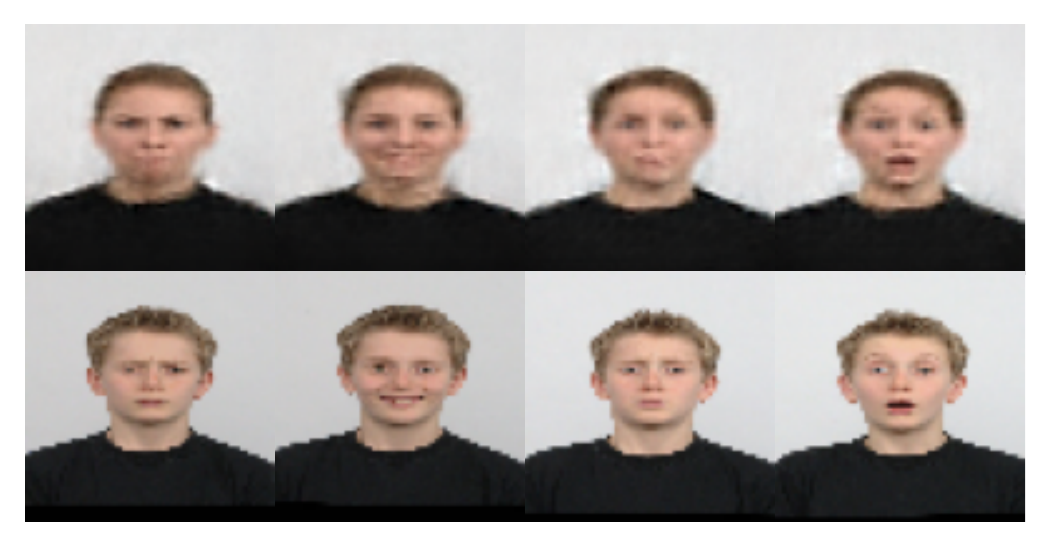}}
    \caption{Closest train sequences (train) to the generated ones (gen.) using our model trained on (a) the \emph{sprites}, (b) \textit{starmen}, (c) \emph{3d chairs} and (d) \emph{faces} datasets.}
    \label{fig:appedix overfit}
\end{figure*}

\clearpage

\section{Experimental Details}\label{app: experiement details}
In this section, we detail all the relevant parameters we used for the experiments. The datasets presented in Appendix \ref{app: the data} are first split into a train set, a validation set and a test set as shown in Table~\ref{tab:spilts}.  We train the models for 200 epochs for \emph{sprites} and \emph{rotMNIST}, 250 for \textit{colorMNIST} and 400 for \emph{starmen} and \emph{chairs} with a latent dimension set to 16 for all datasets but for the \emph{chairs} dataset where it is set to 32. We select the model achieving the lowest validation loss in each case. We use the Adam optimiser \cite{kingma_adam_2014} with a starting learning rate of $10^{-3}$ together with schedulers reducing the learning rate by a factor $0.5$ at epoch 50, 100, 125 and 150 for \emph{starmen}, by a factor $10^{-4}$ at epoch 50, 75, 100, 125 and 150 for \emph{rotMNIST}, by a factor $10^{-4}$ at epoch 50, 100, 150 and 200 for \emph{colorMNIST}, by a factor $0.5$ at epoch 150, 200, 250, 300 and 350 for \emph{3d chairs} and a factor $0.5$ at epoch 50, 100, 125 and 150 for \emph{sprites}. For the \emph{faces} dataset we use a scheduler multiplying the learning rate by $10^{-6}$ every 2,000 epochs and train the model for 10,000 epochs. We use a batch of size 128 for \emph{rotMNIST}, \emph{colorMNIST} and \emph{faces} and 64 otherwise. For the proposed model, we also use 10 \textit{warm-up} epochs where we train it like a VAE to stabilise the encoder and decoder networks and ease the learning of the flows. This hyper-parameter does not influence much the performances as shown in Appendix~\ref{app: ablation study}. The flows are implemented using \cite{chadebec2022pythae} and are composed of 2 IAF blocks using 3-layer MADE \cite{germain2015made} with 128 hidden units. For the variants of our model, we use 500 components in the VAMP prior and IAF flows are composed of 3 IAF transformations using 2-layer MADE with 128 hidden units. All models are trained on a single 32-GB V100 GPU and the FID metrics are computed using the implementation of \url{https://github.com/mseitzer/pytorch-fid}. Finally, we provide the neural networks we use in Table~\ref{tab:nn architectures}. For \emph{faces} we use the same networks as for \emph{sprites} dataset.

\begin{table}[ht]
    \centering
    \begin{sc}
    \begin{scriptsize}
    \begin{tabular}{l|ccc}
    \toprule
        Datasets & Train & Validation & Test  \\
         \midrule
        \emph{Starmen}  & 700 & 200 & 100 \\ 
        \emph{rotMNIST} & 9,000 & 1,000 & 10,000\\
        \emph{colorMNIST} & 48,000 & 12,000 & 10,000\\
        \emph{sprites} & 8,000 & 1,000 & 2,664 \\
        \emph{3d chairs} & 1,000 & 200 & 193 \\
        \bottomrule
    \end{tabular}
    \end{scriptsize}
    \end{sc}
    \caption{Number of sequences considered in the Train/Val/Test splits used in the experiments.}
    \label{tab:spilts}
\end{table}

\begin{table}[ht]
    \centering
    \begin{sc}
    \begin{scriptsize}
    \begin{tabular}{c|ccccc}
    \toprule
        Dataset & Starmen & rotMNIST & colorMNIST & 3d chairs & Sprites \\
        \midrule
        Input dimension & (1, 64, 64) & (1, 28, 28), & (3, 28, 28) & (3, 64, 64) & (3, 64, 64)  \\
        \midrule
          & Conv2D(1, 16, 4, 2)   & Linear(1024)  & Linear(1024)   & Conv2D(3, 16, 4, 2) & Conv2D(3, 16, 4, 2)  \\
            & Conv2D(16, 32, 4, 2)  & ReLU & ReLU & Conv2D(16, 32, 4, 2) & Conv2D(16, 32, 4, 2)   \\
            & LeakyReLU             & Linear(256)   & Linear(256) & LeakyReLU & LeakyReLU\\
           Inference & Conv2D(32, 64, 3, 2)  & ReLU & ReLU & Conv2D(32, 64, 3, 2) & Conv2D(32, 64, 3, 2) \\
           Network & LeakyReLU             & Linear(2x16)$^*$ & Linear(2x16)$^*$ & LeakyReLU & LeakyReLU\\
            & Conv2D(64, 128, 3, 2) & - & - & Conv2D(64, 128, 3, 2) & Conv2D(64, 128, 3, 2)\\
            & LeakyReLU             & - & - & LeakyReLU& LeakyReLU \\
            & 6 ResBlocks           & - & - & 6 ResBlocks & 6 ResBlocks\\
            & Linear (2048, 2x16)$^*$     & - & - & Linear (2048, 2x32)$^*$ & Linear (2048, 2x16)$^*$ \\
            \midrule
            Input dimension & 16 & 16 & 16 & 32 & 16 \\
            \midrule
            & Linear(2048)    &  Linear(256) & Linear(256) & Linear(2048)      & Linear(2048)     \\ 
& ConvT(128, 3, 2)  &  ReLU &  ReLU & ConvT(128, 3, 2)  & ConvT(128, 3, 2)  \\
& 6 ResBlocks       &  Linear(1024) & Linear(1024) & 6 ResBlocks       & 6 ResBlocks    \\   
& ConvT(64, 5, 2)   &  ReLU & ReLU & ConvT(64, 5, 2)   & ConvT(64, 5, 2)   \\
Generative & LeakyReLU         &  Linear(784) & Linear(2352) & LeakyReLU         & LeakyReLU       \\  
Network & ConvT(32, 5, 2)   &  Sigmoid &  Sigmoid & ConvT(32, 5, 2)   & ConvT(32, 5, 2)   \\
 & LeakyReLU         & - & - &LeakyReLU         & LeakyReLU         \\
& ConvT(16, 4, 2)   & - & - &ConvT(16, 4, 2)   & ConvT(16, 4, 2)   \\
& LeakyReLU         & - & - &LeakyReLU         & LeakyReLU         \\
& ConvT(1, 4, 2)    & - & - &ConvT(3, 4, 2)    & ConvT(3, 4, 2)\\
         \bottomrule
         \multicolumn{5}{l}{*Layer outputting the mean and covariance of the variational posterior $q_{\phi}$}
    \end{tabular}
     \end{scriptsize}
    \end{sc}
    \caption{Neural networks architectures used in the experiments and keep the same for all the models in the benchmarks. The ResBlocks use 2 convolution layers with kernel of size 3 and 1, 32 channels and stride 1.}
    \label{tab:nn architectures}
\end{table}

\clearpage

\section{Ablation Study}\label{app: ablation study}

In this section, we present an ablation study of the proposed model where we study the influence of the flow complexity, the latent space dimension, the number of \textit{warm-up} steps (when the model is trained as a VAE) and the prior complexity. We see in Table~\ref{tab:influence flow} and Table~\ref{tab:influence warmup} that neither the choice in the flows nor the number of \emph{warm-up} steps influence much the resulting likelihoods. Table~\ref{tab:influence latent dim} shows that as expected choosing a too small latent space dimension is detrimental to the model performance. Finally, Table~\ref{tab:influence prior} shows the influence of the prior complexity (number of components used in the VAMP prior). As expected, increasing the complexity of the prior allows achieving better likelihood estimates.

\begin{table}[ht]
\begin{minipage}[c]{0.48\linewidth}
    \centering
    \begin{sc}
        \begin{scriptsize}
    \begin{tabular}{cc|cc}
    \toprule
      IAF & MADE   & \multirow{2}{*}{Starmen} & \multirow{2}{*}{Sprites} \\
      Blocks & layers & &\\
      \midrule
1 & 3 & 3774.79 $\pm$ 0.19 & 11302.90 $\pm$ 0.02 \\
2 & 1 & 3773.31 $\pm$ 0.15 & 11302.25 $\pm$ 0.03 \\
2 & 2 & 3774.35 $\pm$ 0.17 & 11301.49 $\pm$ 0.04 \\
2 & 3 & 3773.23 $\pm$ 0.18 & 11301.51 $\pm$ 0.04 \\
2 & 4 & 3773.45 $\pm$ 0.12 & 11302.23 $\pm$ 0.03 \\
2 & 5 & 3773.88 $\pm$ 0.17 & 11301.47 $\pm$ 0.02 \\
3 & 3 & 3773.13 $\pm$ 0.10 & 11302.92 $\pm$ 0.03 \\
4 & 3 & 3774.12 $\pm$ 0.15 & 11301.05 $\pm$ 0.05 \\
\bottomrule
    \end{tabular}
    \caption{Influence of the flow complexity}
    \label{tab:influence flow}
     \end{scriptsize}
    \end{sc}
    \end{minipage}
\begin{minipage}[c]{0.48\linewidth}
    \centering
    \begin{sc}
        \begin{scriptsize}
    \begin{tabular}{c|cc}
    \toprule
      Warmup   & Starmen & Sprites \\
      \midrule
2 & 3773.73 $\pm$ 0.10 & 11301.59    $\pm$ 0.02 \\
5 & 3773.49 $\pm$ 0.10 & 11301.15    $\pm$ 0.03 \\
10 & 3773.23 $\pm$ 0.18 & 11301.51    $\pm$ 0.04 \\
20 & 3774.03 $\pm$ 0.10 & 11302.28    $\pm$ 0.03 \\
50 & 3773.42 $\pm$ 0.11 & 11301.32    $\pm$ 0.08 \\
100 & 3772.44 $\pm$ 0.12 & 11302.40    $\pm$ 0.06  \\
\bottomrule
    \end{tabular}
    \caption{Influence of the warmup steps}
    \label{tab:influence warmup}
     \end{scriptsize}
    \end{sc}
\end{minipage}
\end{table}
\begin{table}[ht]
\begin{minipage}[c]{0.48\linewidth}
    \centering
    \begin{sc}
        \begin{scriptsize}
    \begin{tabular}{c|cc}
    \toprule
      Warmup   & Starmen & Sprites \\
      \midrule
2  & 3817.92 $\pm$ 0.20 & 11346.92 $\pm$ 0.09 \\
8  & 3774.20 $\pm$ 0.19 & 11303.18 $\pm$ 0.02 \\
16 & 3773.23 $\pm$ 0.18 & 11301.51 $\pm$ 0.04 \\
32 & 3773.16 $\pm$ 0.16 & 11302.23 $\pm$ 0.02 \\
64 & 3773.22 $\pm$ 0.13 & 11301.79 $\pm$ 0.02 \\
\bottomrule
    \end{tabular}
    \caption{Influence of the latent dimension}
    \label{tab:influence latent dim}
     \end{scriptsize}
    \end{sc}
    \end{minipage}
    \begin{minipage}[c]{0.48\linewidth}
    \centering
    \begin{sc}
        \begin{scriptsize}
    \begin{tabular}{c|cc}
    \toprule
      VAMP    & \multirow{2}{*}{Starmen} & \multirow{2}{*}{Sprites} \\
      Components & &\\
      \midrule
10  &   3773.55 $\pm$ 0.07 & 11302.82 $\pm$ 0.04 \\
50  &   3772.71 $\pm$ 0.15 & 11301.26 $\pm$ 0.02 \\
100 &   3772.89 $\pm$ 0.16 & 11302.07 $\pm$ 0.03 \\
200 &   3772.66 $\pm$ 0.22 & 11302.03 $\pm$ 0.03 \\
500 &   3772.91 $\pm$ 0.02 & 11301.30 $\pm$ 0.02 \\
\bottomrule
    \end{tabular}
    \caption{Influence to the prior complexity}
    \label{tab:influence prior}
     \end{scriptsize}
    \end{sc}
    \end{minipage}
\end{table}

\clearpage

\section{Influence of Eq.~\eqref{eq: missing optimal} on Missing Data Imputation}\label{app:influence eq 11}

\begin{figure*}[ht]
\centering
    \captionsetup[subfigure]{position=below, labelformat = empty}
    \subfloat{\includegraphics[width=0.7\linewidth]{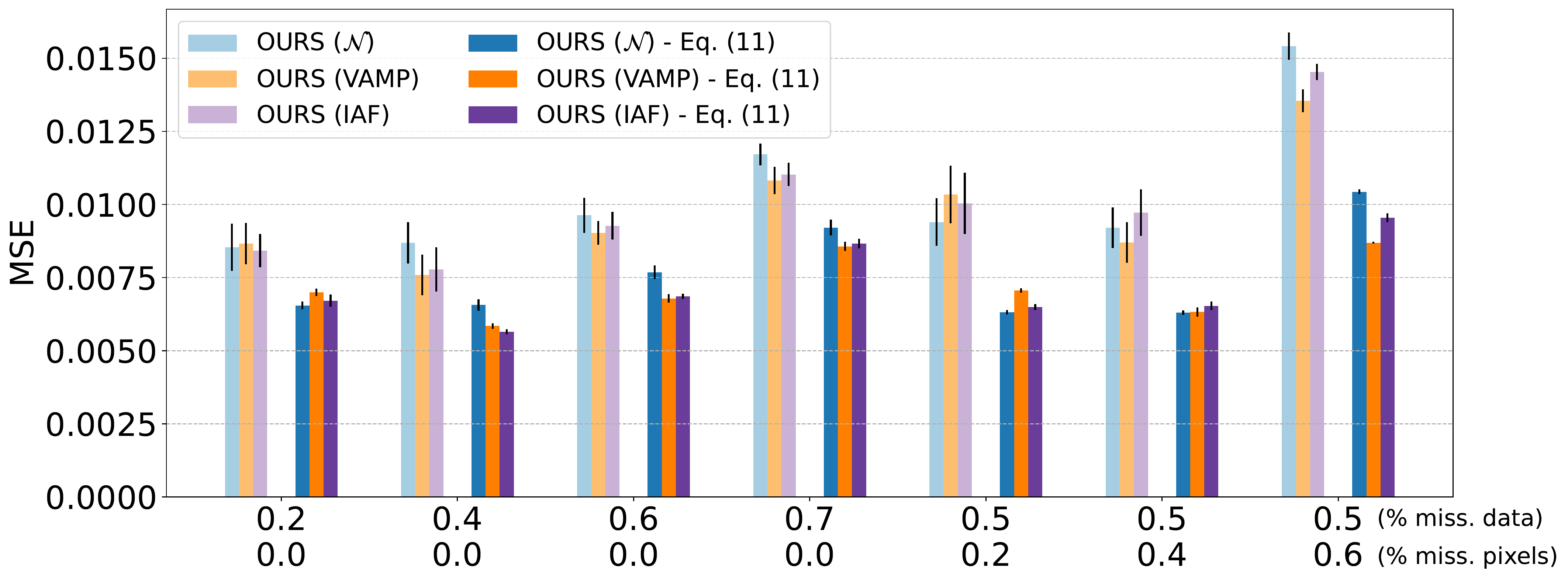}}\\\vspace{-1em}
    \subfloat{\includegraphics[width=0.7\linewidth]{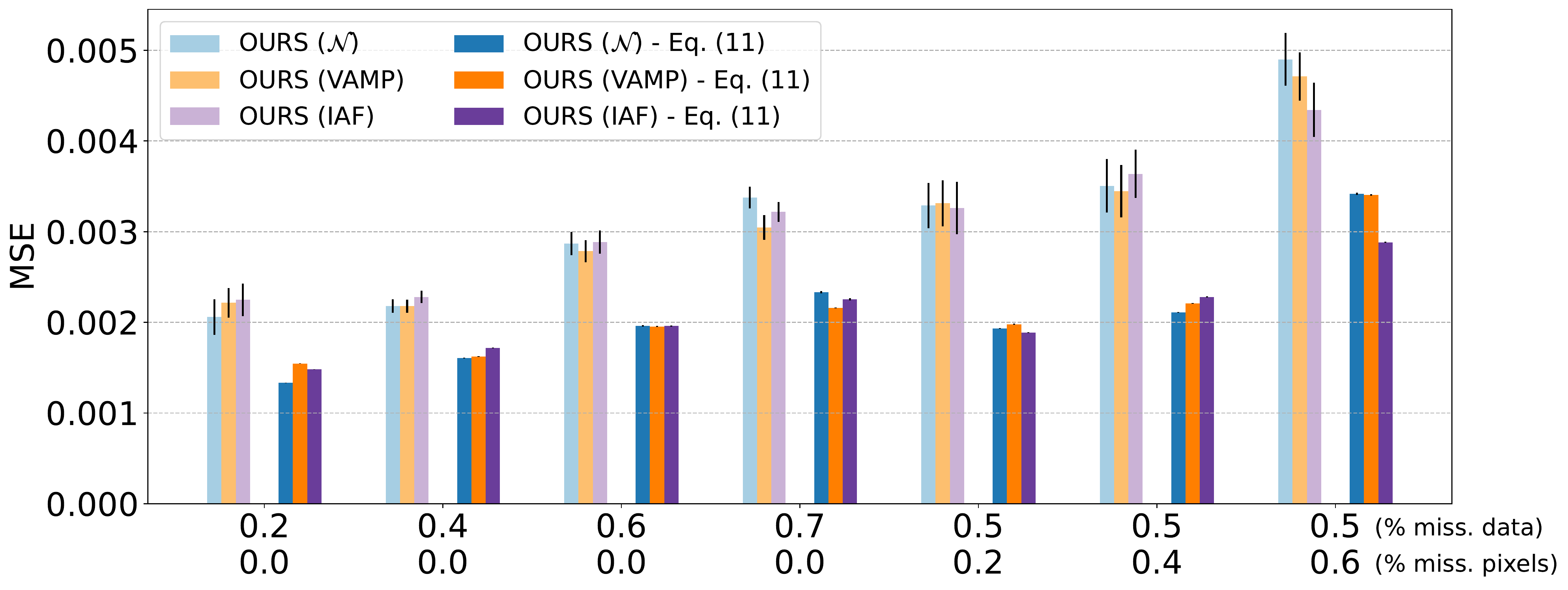}}
    \caption{Mean Square Error (MSE) on missing pixels only of the test data for different proportions of missing observations (0.2 to 0.7) and missing pixels (0.2 to 0.6) in the input train, validation and test sequences for the \emph{starmen} (top) and \emph{sprites} (bottom) datasets. Slightly transparent bars represent the naive method (consisting in using only one randomly chosen data point in the sequence to reconstruct the full sequence as done during training) while solid bars show the results obtained using the method proposed in Section \ref{sec: missing data method}.}
    \label{fig:influence eq 11}
\end{figure*}

In this appendix, we demonstrate empirically the relevance of the method proposed in Section \ref{sec: missing data method} to handle missing observations at inference time. We recall that it consists in drawing one (or several) latent variables from the posterior associated to each data point observed in an input sequence (See Alg.~\ref{alg: infer with missing}). The latent variables are then propagated though the flows and sequences are generated in the observation space using the conditional distribution $p_{\theta}(x|z)$. Using Eq.~\eqref{eq: missing optimal}, we propose to keep the trajectory achieving the highest likelihood on the observed data. This allows to benefit from all the information observed in the sequence. In Fig.~\ref{fig:influence eq 11}, we show the Mean Square Error (MSE) on the missing pixels only for the \emph{starmen} dataset (top) and \emph{sprites} dataset (bottom). We keep the same setting as presented in the paper and remove some data in the input sequences with probability 0.2, 0.4, 0.6 and 0.7 or create sequences with missing observations (randomly removed with probability 0.5) and missing pixels in the observed images (randomly removed with probability 0.2, 0.4 and 0.6). Results obtained with the naive method that consists in using only one randomly chosen data point in the sequence to reconstruct the full sequence as done during training (see Alg.~\ref{alg:training}) are presented by the slightly transparent bars while results obtained using the method proposed in Section \ref{sec: missing data method} are shown by the solid bars. In this example, we generate one trajectory per observed data point and keep the one achieving the highest likelihood according to Eq.~\eqref{eq: missing optimal}. These graphs show the relevance of this method that allows achieving lower MSE in each scenario. Moreover, it allows to decrease significantly the standard deviation (represented by the black bar) leading to a more reliable missing data imputation.

%%%%%%%%%%%%%%%%%%%%%%%%%%%%%%%%%%%%%%%%%%%%%%%%%%%%%%%%%%%%%%%%%%%%%%%%%%%%%%%
%%%%%%%%%%%%%%%%%%%%%%%%%%%%%%%%%%%%%%%%%%%%%%%%%%%%%%%%%%%%%%%%%%%%%%%%%%%%%%%

\end{document}